# Grounding Clinical AI Competency in Human Cognition Through the Clinical World Model and Skill-Mix Framework

Seyed Amir Ahmad Safavi-Naini[1,2,3*], Elahe Meftah[4†], Josh Mohess[3], Pooya Mohammadi Kazaj[1,2,5], Georgios Siontis[1], Zahra Atf[6], Peter R. Lewis[6], Mauricio Reyes[2,7,8], Girish Nadkarni[3,8], Roland Wiest[2,9,10], Stephan Windecker[1], Christoph Gräni[1], Ali Soroush[3,11], Isaac Shiri[1,2]

1- Department of Cardiology, Inselspital, Bern University Hospital, University of Bern, Bern, Switzerland

2- Department of Digital Medicine, Bern University Hospital, University of Bern, Bern, Switzerland

3- Division of Data-Driven and Digital Medicine (D3M), Icahn School of Medicine at Mount Sinai, New York, United States

4- Clinical Research Development Center, Amir Oncology Teaching Hospital, Shiraz University of Medical Sciences, Shiraz, Iran

5- Graduate School for Cellular and Biomedical Sciences, University of Bern, Bern, Switzerland

6- Faculty of Business and Information Technology, Ontario Tech University, Oshawa, Canada

7- Department of Radiation Oncology, Inselspital, Bern University Hospital and University of Bern, Bern, Switzerland

8- ARTORG Center for Biomedical Engineering Research, University of Bern, Bern, Switzerland

8- The Charles Bronfman Institute of Personalised Medicine, Icahn School of Medicine at Mount Sinai, New York, NY, USA

9- University Institute of Diagnostic and Interventional Neuroradiology, Inselspital, Bern University Hospital, University of Bern, Bern, Switzerland

10- Translational Imaging Center (TIC), Swiss Institute for Translational and Entrepreneurial Medicine, Bern, Switzerland

11- Henry D. Janowitz Division of Gastroenterology, Icahn School of Medicine at Mount Sinai, New York, NY, USA

[†] Elahe Meftah contributed equally as first authors to this work.

[*] Correspondence to Seyed Amir Ahmad Safavi-Naini (sdamirsa@ymail.com, seyed.safavinaini@students.unibe.ch)

# Abstract

The competency of any intelligent agent is bounded by its formal account of the world in which it operates. Clinical AI lacks such an account. Existing frameworks address evaluation, regulation, or system design in isolation, without a shared model of the clinical world to connect them. We introduce the Clinical World Model, a framework that formalizes care as a tripartite interaction among Patient, Provider, and Ecosystem. To formalize how any agent, whether human or artificial, transforms information into clinical action, we develop parallel decision-making architectures for providers, patients, and AI agents, grounded in validated principles of clinical cognition. The Clinical AI Skill-Mix operationalizes competency through eight dimensions. Five define the clinical competency space (condition, phase, care setting, provider role, and task) and three specify how AI engages human reasoning (assigned authority, agent facing, and anchoring layer). The combinatorial product of these dimensions yields a space of billions of distinct competency coordinates. A central structural implication is that validation within one coordinate provides minimal evidence for performance in another, rendering the competency space irreducible. The framework supplies a common grammar through which clinical AI can be specified, evaluated, and bounded across stakeholders. By making this structure explicit, the Clinical World Model reframes the field's central question from whether AI works to in which competency coordinates reliability has been demonstrated, and for whom.

# 1. Introduction

The competency of an intelligent agent and the adequacy of its design, whether human or artificial, are constrained by the formal account of the world they operate in. In clinical care, artificial intelligence (AI) has advanced without regard for this constraint, excelling on benchmarks while failing in deployment in real-world[1–4]. Analyses of radiological AI reveal that fewer than 6% of externally validated models maintain their original performance, with the area under the curve declining by an average of 8% upon external validation[5]. Similar patterns emerge across clinical domains, where systems trained to achieve a remarkable accuracy on curated datasets encounter unexpected failure modes when confronting the heterogeneity of real patients, variable imaging equipment, and evolving disease presentations[6–8]. Large language models (LLMs) offered hope, achieving passing scores on medical licensing examinations[9]. Yet examination success has not translated into clinical reliability, as models still falter under the uncertainty inherent in authentic clinical reasoning[6,10].

To overcome these limitations, the field is now building agentic architectures around LLMs, augmenting them with planning, memory, and tool use to pursue clinical goals through sequential reasoning with varying degrees of autonomy[11,12]. Yet these architectures inherit the unreliability of their underlying models while introducing cascading risk and diminishing human oversight[13]. In an agentic workflow, a single misread laboratory value can silently propagate into an incorrect differential and ultimately a harmful recommendation. The agentic turn amplifies the disconnect between the inherent structure of clinical care and the frameworks through which AI systems are designed and evaluated.

The root of this mismatch is not merely technical[14]. Each patient, provider, and care setting presents distinct characteristics that shape clinical encounters. Clinical care is fundamentally interactional, unfolding through iterative exchanges between patients and providers[15]. It is temporal, extending across care phases from risk identification through long-term management[16]. Moreover, it is cognitively complex, requiring the integration of pattern recognition with analytical deliberation, and is calibrated by metacognitive monitoring[17]. Benchmarks cannot capture these features, as they inherently abstract away the very characteristics that define clinical reasoning.

Two decades of framework development have established critical infrastructure for clinical AI, addressing complementary facets of a complex problem (**Table 1**; **Supplementary Note 1**)[16,18–] [28]. Research on healthcare work systems has clarified how organizational conditions, tools, and care environments collectively shape quality and safety[16,19,20]. Regulatory organizations operationalized risk stratification for AI-based medical devices, connecting oversight requirements to clinical function and transparency[21]. Evaluation benchmarks have facilitated the systematic measurement of capabilities across clinical tasks, languages, and populations[18,22,24,28]. Agentic architectures advanced reasoning transparency through multi-agent designs that model hypothesis generation, verification, and iterative refinement[23,25–27]. Together, these efforts have strengthened the field's ability to measure what AI systems produce.

**Table 1. Framework Traditions in Clinical AI and Their Foundational Contributions.** Four established traditions and three contributions introduced in this work are compared across foundational questions, key contributions, theoretical grounding, and dimensional topology. Each tradition addresses a subset of the thirteen dimensions defined in Section 2.1; Dimension coverage is expressed as topology using four edge types: causal (→), constraining (⊣), constitutive (↔), and temporal (over Temporality). The "Introduced in this work" rows represent the three primary contributions of this paper.

| Tradition | Foundational Question | Key Contribution | Theoretical Grounding | Dimension Coverage Topology |
|---|---|---|---|---|
| **Sociotechnical Systems (SEIPS 1.0, 2.0, 3.0)[19,20,16]** | How do work systems shape care outcomes? | Established care as an emergent system property; work system–outcome relationships | Human factors engineering; structure-process-outcome theory | Context × Actors → Outcomes and Actions over Temporality |
| **Regulatory Science (CORE-MD)[21]** | How should AI risk determine evidence requirements? | Operationalized risk stratification; linked oversight level to clinical function | Risk-benefit analysis; medical device regulation | Normativity ⊣ Authority → Outcomes and Adaptation |
| **Evaluation Benchmarks (MedHELM[18], GlobMed[24], Expert Consensus[22], MEDIC[28])** | How can we systematically measure AI capability? | Enabled systematic measurement across tasks, languages, and safety dimensions | ML evaluation methodology; psychometrics | Normativity → Outcomes and Adaptation |
| **Agentic Architectures (ArgMed-Agents[23], ClinicalLab[25], DynamiCare[26], KG4Diagnosis[27])** | How should AI reasoning be structured? | Advanced analytical reasoning, iteration, and multi-agent dynamics | Multi-agent systems; computational reasoning | Cognition → Action over Temporality under Adaptation |
| *Introduced in this work* | | | | |
| *Clinical World Model* | How can clinical reality be formally described as a unified structure for all actors? | Recovers the structure that prior traditions implicitly share; organizes dimensions and ten views | binding problem (philosophy of mind); sociotechnical systems theory | Actors × Context → Action |
| *Decision Making Architectures* | How do actions evolve from information processing through cognition? | Defines a generalizable structure for action generation across human, digital, and collaborative actors | Dual-process cognitive theory; shared decision-making; information theory | Information → Representation → Cognition → Action and Adaptation over Temporality |
| *Clinical AI Skill-Mix* | How can clinical AI competency be organized into bounded, evaluable units? | Structures AI capability specification through competency-based frameworks commensurable with clinical training | Entrustable professional activities; competency-based medical education | Codex ↔ Mandate → Outcomes |

What remains unaddressed is a unifying model of the clinical world itself, one that recognizes existing frameworks as partial projections of a shared reality rather than independent accounts. Three structural gaps explain why these traditions have not converged. Existing frameworks do not formalize how agents, information, and environments interact as a coherent system. Without such a model, they cannot specify how providers and patients actually transform information into clinical decisions, and therefore cannot articulate how any form of intelligence should integrate with that process, whether as an advisor within human reasoning, a monitor over it, or an autonomous actor replacing it. Nor do they define when validation in one clinical context provides evidence for another, leaving the mandate and boundaries of each clinical task formally unspecified. These gaps cascade and without a shared ontology of the clinical world, stakeholders engage clinical AI through incommensurable vocabularies: providers cannot locate AI outputs within their reasoning, regulators cannot anchor evidence requirements to cognitive function, and developers cannot derive implementable specifications from clinical need.

Progress may require a reorientation that, paradoxically, looks backward in order to move forward. The tools for this reorientation come from drawing on rich empirical traditions in medical cognition and human factors that predate clinical AI. Medical education research offers one lens, explaining how competency develops through structured acquisition of clinical reasoning skills, how expertise differs qualitatively from novice performance, and how assessment can capture reasoning processes rather than merely outcomes[29]. Cognitive science offers another lens, mapping how physicians reason through illness scripts and how dual-process cognition balances rapid pattern recognition against deliberate analytical evaluation[30]. These traditions are supported by decades of empirical validation and provide the grounding that current clinical AI specification lack, yet they remain largely disconnected from contemporary AI development discourse.

Recent proposals have called for grounding AI capabilities into theory-based frameworks[31–34], medical certification pathways[35], authentic evaluation methods[1], and principled deployment frameworks[3] that bind AI competency to the established structure of clinical care[2]. Such binding requires boundaries that are sufficiently precise to enable systematic evaluation while remaining grounded in clinical meaning.

Here we propose three interconnected models: the *Clinical World Model*, defining the actors, substrates, and interactions that constitute care; the *Decision-Making* architectures, specifying how clinical, patient, and AI agents decide to act; and the *Clinical AI Skill-Mix*, operationalizing competency specification. We ground clinical AI specification in empirically validated principles of cognition and of clinical practice, adopting an ontological binding approach to formalization. The central hypothesis is that the competency space of clinical AI is combinatorially vast, context-dependent, and irreducible, and that systematic specification requires formalizing the world within which care occurs, the cognitive operations that unfold within it, and the competency structures that bind them. The following sections develop the theoretical foundations and scope (Section 2), present the framework architecture and cognitive models (Sections 3 through 5), operationalize competency decomposition (Section 6), and discuss implications (Section 7).

## 2. Foundation and Scope

Multiple traditions have formalized valid but partial accounts of the same clinical reality, producing what philosophy of mind terms a binding problem: separately processed features of a shared scene that do not cohere into a unified structure[36]. Rather than proposing yet another independent account, the present work recovers the integrative structure that prior frameworks implicitly share. We identify the structural commitments common to existing frameworks (**Supplementary Note 1**), ground them in the dimensional reality of clinical world (**Supplementary Note 2**), and align them with decades of validated cognition research (**Supplementary Note 3**).

### 2.1 Dimensions within the World

Artificial and human intelligence have evolved in tandem since computing's inception[37,38]. Central to both fields is the concept of world models. The psychologist and philosopher Craik first proposed that the mind constructs "small-scale models" of reality to anticipate events[39]. Machine learning researchers adopted the term for learned compressions of environmental dynamics that support planning through simulation[40]. Across traditions, world models converge on a common function: world representations enabling an agent to simulate potential futures before acting[41,42].

AI in care operates within a world defined by distinct dimensions and interacting elements. We formalize world space through thirteen dimensions (**Figure 1**), selected for representational sensitivity, conceptual specificity, and structural parsimony (**Supplementary Note 2.1**). At the foundation of world lies an underlying *Axiom* from which all observations derive. What can be observed accumulates as *Information*, a portion of which crystallizes into decoded, clinically legible *Codex* such as disease classifications, treatment protocols, and diagnostic categories. For example, the electrical activity of the heart (Axiom) produces waveforms on an electrocardiogram (Information), from which clinicians have decoded patterns such as ST-elevation to diagnose myocardial infarction (Codex).

When a clinical need arises, a *Mandate* triggers an actor's willingness to act. *Actors* operate within a *Context* of infrastructure, resources, and norms. They reason toward action through *Cognition*, transforming data into *Representations* constrained by the decisional *Authority* allocated to them. Each action advances the world one state forward through *Temporality*, resuming the cycle. A parallel feedback loop evaluates *Outcomes* against defined criteria, governed by the *Normativity* that determines which actions are permissible. Through this process, Cognition and Codex *Adapt*.

For instance, a patient with chest pain seeks care (patient Mandate) and a physician receives the case (physician Mandate). In an equipped emergency department (physician Context), she recognizes ST-elevation on the ECG (Cognition and Representation), activates the catheterization team (Authority), and the world state advances (Temporality). Recovery (Outcome) is judged against what is defined as good care (Normativity), refining clinical cognition and the coded protocol for future care (Adaptation).

These dimensions constitute the state space of reality, but dimensions alone do not determine a model view. The same dimensions yield fundamentally different frameworks depending on the stance adopted (descriptive, normative, or generative), the query compressed, the topology imposed among dimensions, and the resolution at which the system is examined (see Supplementary Note 2.2 for different modeling frameworks). This explains why multiple traditions, each observing the same clinical world, have produced valid but structurally distinct accounts: each tradition selected overlapping dimensions but compressed a different question toward a different intention, producing divergent models, frameworks, and systems (Supplementary Table 1).

Dimensions define the state space, but it is the worldview projected onto them that dictates model architecture. Two frameworks may draw on the same dimensions yet diverge entirely because they adopt a different stance (descriptive, normative, or generative), compress a different query, impose a different topology among dimensions, or examine reality at a different resolution (see

 for different modelling traditions). This is precisely why multiple traditions, each confronting the same clinical reality, have produced structurally distinct accounts rather than converging on one ().

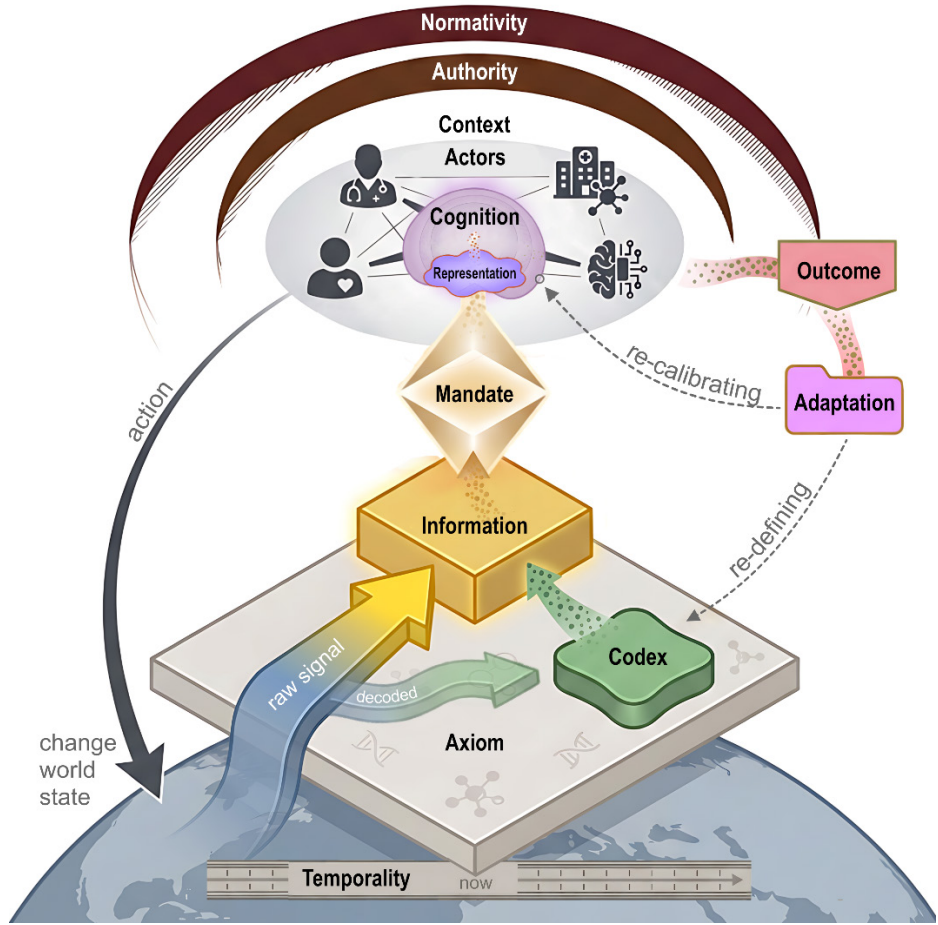

**Figure 1. Dimensions of the World.** Conceptual diagram illustrating the thirteen dimensions taxonomies that constitute the clinical world. Normativity and Authority form overarching regulatory arcs that govern all elements below. Context, Actors, Cognition, and Representation are nested within the clinical scene, where multiple actors (providers, patients, AI systems, and ecosystem components) interact through cognitive processes grounded in internal representations. Mandate mediates between the cognitive layer and the informational substrate, defining the scope of permissible action. Information and Aiophysical together form the material foundation of clinical care, with Codex representing the enacted practices through which knowledge is applied to the world. Temporality runs along the base axis, situating all dimensions within a temporal frame. An action arc (left) connects cognition to change in world state. Outcome and Adaptation (right) close the loop: outcomes feed back into the system through re-calibrating (adjusting within existing parameters) and re-defining (altering the parameters themselves), enabling the clinical world to evolve over time.

## 2.2 Scope

Before competency can be decomposed, the cognitive operations it comprises must be formalized; and before cognition can be formalized, the world within which it occurs must be defined. The present work develops one model for each link in this chain: the Reference World Model formalizes the clinical scene through the tripartite architecture of Patient, Provider, and Ecosystem (Section 3); parallel Decision-Making Models formalize how providers and patients transform information into action across time (Sections 4 and 5); and the Clinical AI Skill-Mix operationalizes competency decomposition, specifying the mandate and boundaries of each clinical task (Section 6). The scope boundary addresses operational prerequisites on which the remaining models depend: we define what exists in the clinical scene (Reference World Model), not how agents learn from it (Learning World Model); and we investigate how agents act (Decision-Making Model), not how they represent one another (Mental Model).

## 3. The Clinical World Model

Clinical care is fundamentally interactional[16], such that a provider does not diagnose in isolation, nor does a patient decide alone[43]. Rather, care emerges through interaction among people within an environment that shapes what is possible, knowable, and actionable[17,20,43,44]. The Clinical World Model formalizes this dynamic by identifying three players whose interactions constitute the scene of clinical practice. This framing from isolated agents to the relational fabric that connects them, providing the ontological foundation upon which intelligent systems must be built.

The three players are Patient, Provider, and Ecosystem[20,43]. The *Patient* is the individual experiencing illness or risk whose condition motivates the encounter and whose preferences ultimately guide care[15,43,45]. The term *Provider* refers to any healthcare professional involved in a patient's care. It is deliberately chosen to encompass physicians, nurses, pharmacists, therapists, and community health workers, rather than to narrow to a single professional role [16,20]. The *Ecosystem* comprises the environment within which care unfolds, including physical spaces, information systems, medical devices, human expertise beyond the immediate provider, and organizational routines that enable or constrain action[16,20,21,26]. These three players form a triangular relationship in which each vertex connects to the other two (**Figure 2a**).

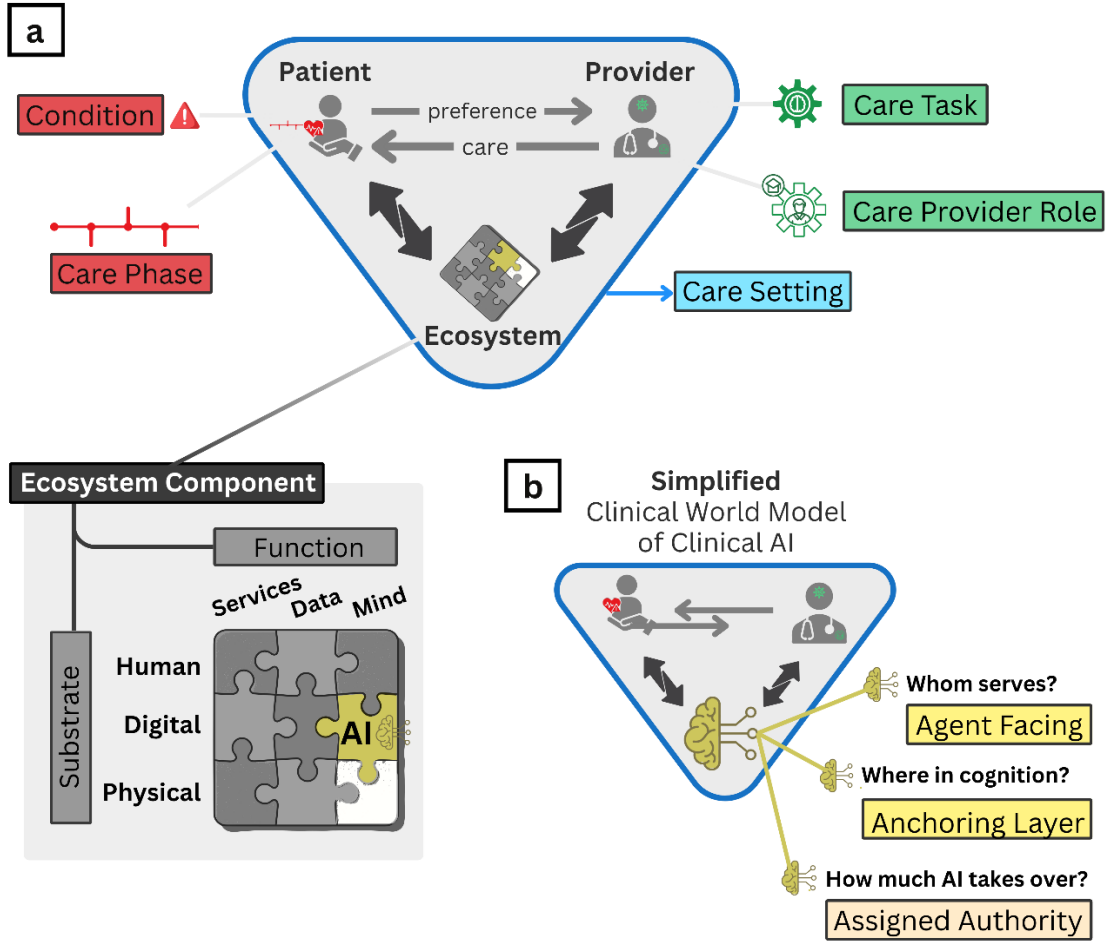

**Figure 2. The Clinical World Model framework to situate clinical AI within healthcare delivery.** (a) Tripartite architecture of the Clinical World Model, comprising the Patient, Provider, and Ecosystem. Preference and action arrows indicate the bidirectional exchange between Patient and Provider, with both drawing upon the Ecosystem. Five Skill-Mix dimensions characterize the clinical encounter: Condition, Care Phase, Care Task, Care Provider Role, and Care Setting. The Ecosystem is further structured by a Function-by-Substrate matrix (Services, Data, Mind × Human, Digital, Physical), with clinical AI occupying the Mind × Digital cell (see Supplementary Figure 1 for detail of Ecosystem components). (b) Simplified Clinical World Model for clinical AI. The AI agent is positioned within the triangular interaction between Patient and Provider. Three additional dimensions specify the AI's role within the encounter: Agent Facing (whom the AI serves), Anchoring Layer (where in the cognitive process the AI engages), and Assigned Authority (how much decisional control the AI assumes).

Information flows bidirectionally between Patient and Provider[26,44]. The patient communicates preferences, which encompass values, goals, concerns, symptoms, and choices[41,45–47]. The provider communicates actions, which include assessments, explanations, recommendations, and interventions. Importantly, both human players also interact with the Ecosystem, retrieving data, consulting expertise, and utilizing services, as well as being constrained by available resources[20].

To structure the Ecosystem, we introduce a Function-by-Substrate matrix that classifies its components along two dimensions (**Figure 2a**). The functional dimension captures what ecosystem components do: store and transmit information (Data), provide reasoning and interpretation (Mind), or execute actions and deliver care (Service). The substrate dimension captures what ecosystem components are built on: people such as clinicians and care teams (Human), software and algorithms (Digital), or devices and tangible materials (Physical). Crossing three functions with three substrates yields nine ecosystem categories, described in **Supplementary Figure 1**.

Clinical AI occupies the Mind × Digital cell of the Function-by-Substrate matrix. This logical binding identifies where AI sits within the ecosystem. As a Mind component, AI engages with reasoning processes rather than merely storing data or executing physical actions. As a Digital substrate, AI operates through computational processes distinct from human cognition. Understanding AI's position within this matrix is a prerequisite for specifying how AI should engage with human players and other components. To keep the framework tractable, we simplified the scene into its three reasoning actors, Provider, Patient, and Clinical AI, whose triangular interaction forms the core of this paper (**Figure 2b**).

## 4. Decision Making Model Within the Clinical World

Having established the structure of the clinical care scene, we now turn to the process of action and mind. The *Clinical Decision Making* (CDM) and *Patient Decision Making* (PDM) Models formalize how clinical providers and patients transform inputs into actions, treating each as a cognitive agent with shared architecture but distinct content domains and epistemic positions. The *Artificial Agent Decision Making* (ADM) Model extends this architecture to clinical AI, mapping each human cognitive component to its computational counterpart.

The decision-making skeleton draws on decades of research in human cognition. **Supplementary Table 2** details each theoretical foundation and its contribution to the architectures, spanning dual-process theory, illness script formation, metacognitive monitoring, action regulation, and phenomenological accounts of lived experience[15,16,19,20,41,43–69]. **Supplementary Note 3** provides detailed descriptions of all three models; the following subsections present the essential information for understanding their structure and interrelationships.

Our proposed decision-making architecture operates through four elements: Input, Processors, Priors, and Action. The mind receives information from distinct *Input* clusters and transforms it through functional processing units. We define these *Processors* as functionally distinct units, applying separation of concerns to make a complex system tractable for design, error analysis, and targeted integration. *Priors*, encompassing knowledge, experience, and current processing capacity, configure how each Processor operates. Processors progressively refine raw data into cues, cues into hypotheses, hypotheses into plans, and plans into actionable steps[49–53,56,63]. This refinement is not linear; it unfolds within a *Recurrent Processing Sphere* where stages interact dynamically until sufficient convergence is reached[44,52,56–58]. Action then alters the state of the world, advancing the temporality dimension one step forward. The changed world state mandates subsequent decision-making, closing the external loop[44,52]. In parallel, a feedback loop compares actual action consequences against projected ones, adopting for future processing[41].

## 4.1 Clinical Decision Making (CDM) Model

The CDM model formalizes the cognitive architecture through which providers reason during clinical encounters (**Figure 3**; detailed in **Supplementary Note 3.2**). Providers enter each decision iteration drawing from four Input clusters[16,20]. *Encounter Data* encompasses what is directly perceived during patient interaction, from verbal communication and physical findings to clinical sense, the gestalt perception that experienced practitioners struggle to fully articulate. *Patient Preference* provides the communicated context and socioeconomic circumstances that influence both disease presentation and treatment feasibility[20,46]. *Encounter Context* captures environmental parameters including available resources, team dynamics, institutional routines, and legal frameworks. *Recorded Data* extends the encounter data through documented information residing in the Ecosystem from previous encounters, including history, imaging, laboratory results, and procedural records.

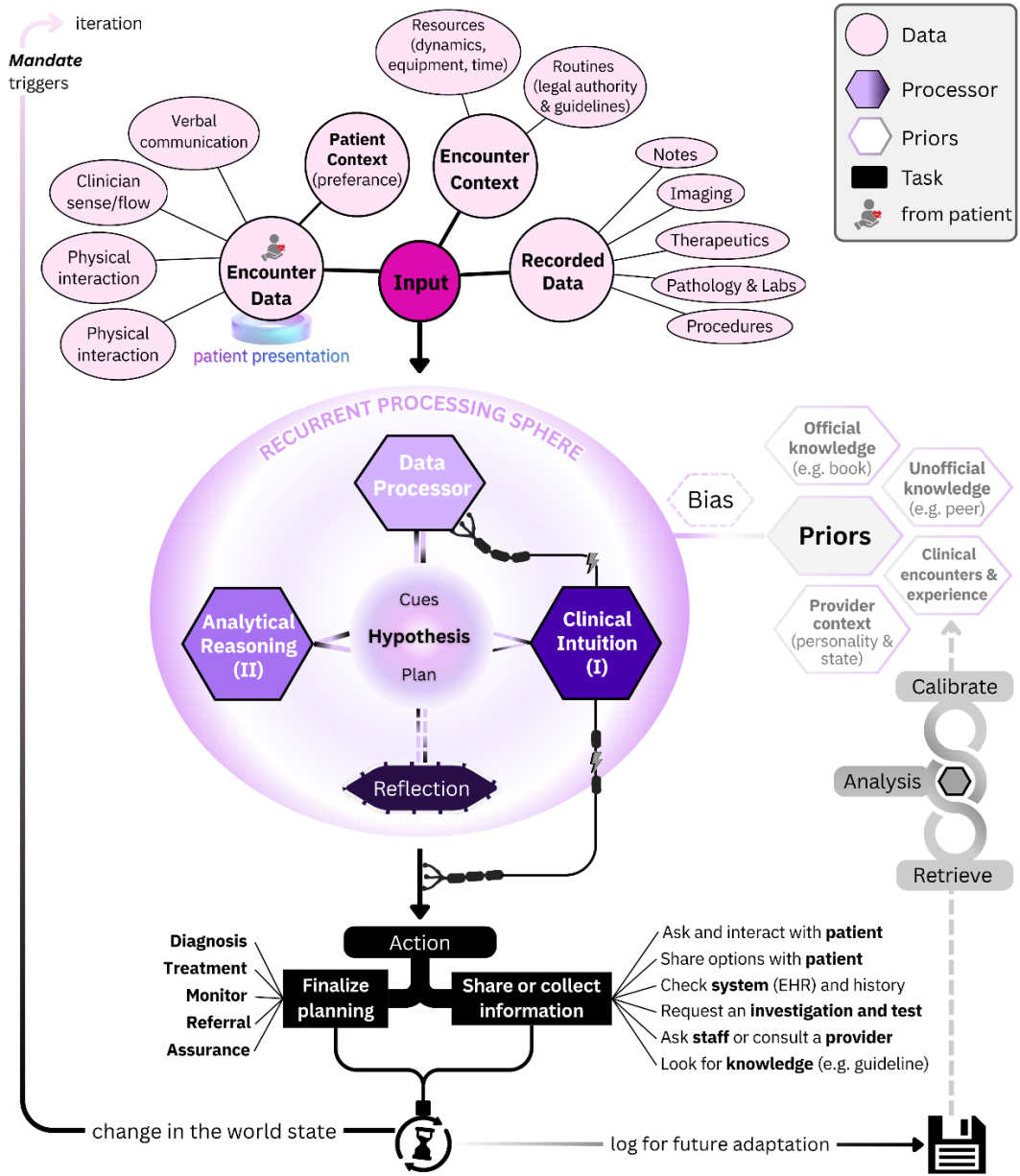

**Figure 3. The architecture of the Clinical Decision Making (CDM) Model.** Architecture of a CDM cognitive cycle. A mandate triggers each iteration (top left). Input integrates four data streams: Encounter Data drawn from direct interaction (verbal communication, clinician sense, physical interaction), Patient Context (patient preference), Encounter Context (resources, routines, and legal authority), and Recorded Data from clinical systems (notes, imaging, therapeutics, pathology and labs, procedures). Input flows into the Recurrent Processing Sphere, where the Data Processor extracts cues that generate and refine hypotheses. Two reasoning pathways operate over these hypotheses: Analytical Reasoning (System II) and Clinical Intuition (System I), with Reflection providing metacognitive oversight. Gating mechanisms (depicted as switches) regulate information flow between processing components. Priors, comprising official knowledge, unofficial knowledge, clinical experience, and provider context, feed into the processing sphere through three pathways: Retrieve, Analysis, and Calibrate, with Bias acknowledged as an influence on prior integration. Processing resolves into Action, which takes one of two forms: Finalize Planning (diagnosis, treatment, monitoring, referral, or assurance) or Share or Collect Information (interacting with the patient, consulting the system, requesting investigations, or seeking knowledge). Each action produces a change in the world state and is logged for future adaptation.

The Recurrent Processing Sphere transforms raw input into action through principled stages of maturation[53,60,62]. The *Data Processor* applies selective attention to extract clinically relevant signals (*cues*), which trigger candidate explanations (*hypotheses*), inform strategies aligned with the aim of care (*plans*), and ultimately translate into actions[52,59]. Evaluation and maturation proceed through two complementary systems in dynamic interplay[49,52]. *System I* (*Clinical Intuition*) provides rapid pattern recognition through consolidated illness scripts, operating largely outside conscious articulation and susceptible to bias[49–51]. *System II* (*Analytical Reasoning*) provides deliberate hypothetico-deductive evaluation, engaged when presentations are atypical, stakes are high, or intuitions generate uncertainty[53,54].

System I and System II operate in synergistic, bidirectional oscillation, with intuitive recognition generating hypotheses for analytical scrutiny while analytical findings trigger new patterns [52]. Expert clinicians toggle fluidly between modes, and under high recognition confidence, System I may bypass the full Recurrent Processing Sphere entirely, as when an emergency physician recognizes ST-elevation myocardial infarction and acts without deliberation[54,65]. This bypass explains both expert efficiency and susceptibility to bias; the capacity to judge when it is appropriate itself develops through accumulated experience[65,66]. Guarding against such errors, *Reflection* provides the metacognitive gate that distinguishes competent from hazardous reasoning. It evaluates whether evidence supports the emerging conclusion, projects the consequences of error, and rejects outputs when uncertainty or risk exceeds thresholds, returning processing to earlier stages[47,56,61,64,65]. This gate is not obligatory; indeed, its very bypass is what enables expert efficiency, even as it opens the door to expert error.

Throughout every stage and each processor, cognition is influenced by Priors, the accumulated configuration of the processor itself[31,63]. These encompass official knowledge (guidelines, research)[70], unofficial knowledge (clinical pearls, peer wisdom)[70], clinical experience (illness scripts, exemplar cases)[66], and provider context (cognitive style, fatigue, emotional state)[20,65,67]. Priors determine which cues attract attention, which hypotheses seem plausible, and how Reflection calibrates confidence.

The reasoning process culminates in Action, taking one of two forms. *Finalize planning* generates clinical outputs such as diagnosis, treatment, monitoring, or referral. *Share or collect information* initiates activities such as questioning the patient, ordering tests, checking records, or consulting colleagues. Neither form is terminal; each action modifies the world state and feeds back as updated input for the next iteration[41,47,67]. This external loop spans seconds (checking digital records), minutes (asking a question) to years (managing chronic disease). In parallel, a feedback architecture we characterize as *Retrieve-Analyze-Calibrate* operates across longer timescales, logging each iteration's trace, comparing projected against actual outcomes, and updating Priors to refine future processing, thereby accumulating clinical expertise[47,61,66].

## 4.2 Patient Decision Making (PDM) Model

The patient is an active cognitive agent whose reasoning shapes clinical outcomes, not a passive data source[15,43,45]. Patient-facing AI that fails to model this agency cannot meaningfully support patient deliberation. Provider-facing systems must equally account for it, as Patient Preference, itself a product of active reasoning, constitutes a primary Input cluster in the CDM. The PDM model mirrors (details the CDM architecture while incorporating inputs and mediators specific to the patient's epistemic position and lived experience (**Figure 4**; detailed in **Supplementary Note 3.3**)[48,68,69].

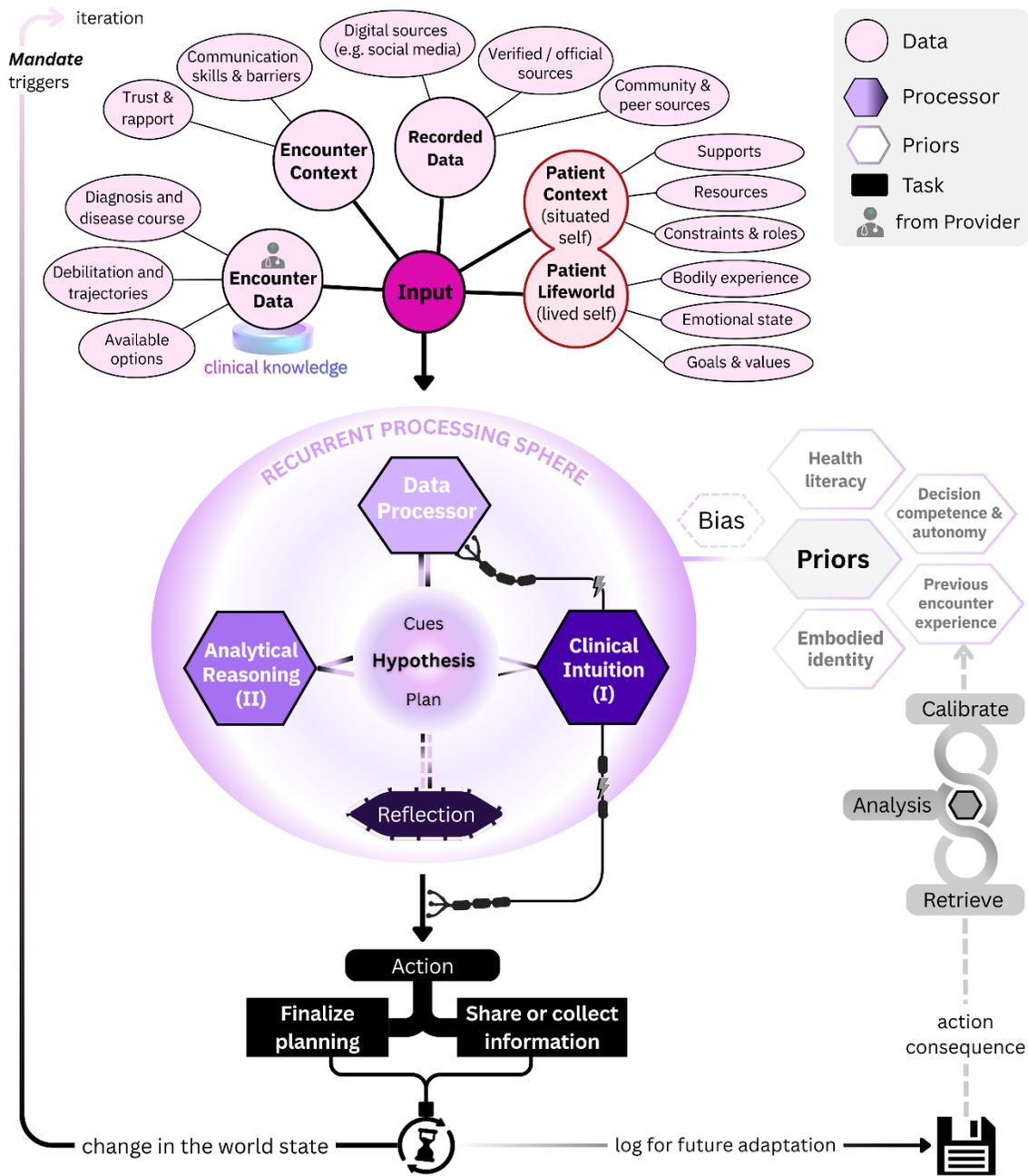

**Figure 4. The architecture of the Patient Decision Making (PDM) Model.** Architecture of a PDM cognitive cycle. A mandate triggers each iteration. Input integrates four data streams: Encounter Data (diagnosis, disease course, available options), Encounter Context (trust, communication, digital and official sources), Recorded Data (community and peer sources), and two patient-specific streams, Patient Context (situated self, including supports, resources, and constraints) and Patient Lifeworld (lived self, including bodily experience, emotional state, and goals and values). Input flows into the Recurrent Processing Sphere, where the dual-process architecture comprises Analytical Reasoning (System II), Clinical Intuition (System I), and Reflection. Priors reflect the patient's cognitive resources: health literacy, decision competence and autonomy, previous encounter experience, and embodied identity. Processing resolves into Action, either Finalize Planning or Share and Collect Information, with each action producing a change in the world state, an action consequence, and a log for future adaptation.

The critical differences emerge at the Input and Prior layers. Where providers receive encounter data through trained clinical observation, patients receive it as communicated diagnoses, prognoses, and options, filtered through encounter context factors such as trust, rapport, and communication barriers[45]. Where providers access curated recorded data through institutional systems, patients draw from a heterogeneous ecosystem spanning verified sources, community networks, and digital resources of highly variable quality[68]. Most distinctively, two Input categories have no provider equivalent and is what provider explore through encounter: *Patient's Lifeworld* captures the *lived self* (bodily experience, emotional state, goals and values), while *Patient's Context* captures the *situated self* (supports, resources, constraints and roles)[48,68,69]. The provider investigates the Lived Self through its manifestation in the encounter, while receiving Patient Preference as the selective encoding of Patient Context that the patient chooses to disclose.

Priors diverge accordingly[59,67]. Health literacy replaces official knowledge, accumulated encounter experience replaces clinical experience, and decision competence replaces provider context (i.e. personality and professional capacity)[69,71,72]. A key asymmetry governs bias. In the CDM, provider bias operates solely as a Prior to be minimized. In the PDM, patient values occupy two architectural positions: consciously as Input through Situated Self, and unconsciously as Prior through Embodied Identity[69].

The PDM model presented here reflects a conception of the patient as an empowered and autonomous agent, consistent with modern Western medical ethics and patient-centered care[71]. However, decisional authority between patients, families, and providers varies substantially across cultures and clinical contexts[73]. The framework accommodates this variability through configuration of its inputs and information flows rather than architectural modification; when family serves as the primary decision-maker, the PDM model applies to that unit, with relationship factors and the relative weights of Lived Self and Situated Self shifting according to cultural norms. We present the autonomous patient as the reference case while recognizing that clinical reality encompasses a spectrum of decisional arrangements.

### 4.3 Digital Agent Decision Making (ADM) Model

The ADM model completes the triad by extending the shared architecture to clinical AI (**Supplementary Figure 2**; detailed in **Supplementary Note 3.4**). The AI agent operates within the same clinical scene as human agents but accesses it through digitally mediated channels, unable to perceive embodied cues except insofar as they are encoded in available data[2]. Input clusters parallel those of the CDM and PDM, with the addition of Design Specification and Configuration, which encodes developer-defined instructions, logic flows, and control parameters. Priors, comprise supervised and documented knowledge (curated training data, analogous to official knowledge), observational knowledge (patterns extracted from unlabeled or labeled clinical data), experiential learning (refined through reinforcement signals and interactions), and agent capacity (computational resources and autonomy constraints), the last having no direct analog in human cognition[74,75].

The Processors are where the ADM diverges most from its human counterparts, replacing each with a machine-learned analog. *Attentive Abstraction* supplants the Data Processor, employing learned attention and dimensionality reduction to convert heterogeneous input into structured representations. The *Latent Space* replaces the Hypothesis-Cues-Plan scaffold; candidate explanations exist in continuous, high-dimensional form rather than as discrete propositions, with direct consequences for explainability[76]. *Instant Reasoning*, the System I analog, encompasses rapid feedforward inference, as in single-pass neural classification, operating over statistical regularities with limited phenomenological grounding. *Sequential Reasoning*, the System II analog, encompasses deliberate multi-step computation, as in chain-of-thought prompting or agentic systems that decompose problems iteratively[23]. *Trajectory Projection* replaces Reflection but diverges most from its human counterpart, performing uncertainty quantification, trajectory simulation, and risk estimation via programmatic metrics rather than metacognitive monitoring[14,76]. It can reject outputs and return processing to earlier stages, but cannot access the experiential dimensions that inform human judgment about when evidence is enough. The ADM also introduces a third Action type: Request Confirmation, encoding the requirement for human verification when confidence falls below threshold or actions exceed authorized scope, directly operationalizing Assigned Authority at the cognitive level[12].

The ADM is presented here as a structural counterpart to the CDM and PDM, mapping the architectural parallel at the level of functional roles rather than claiming the same empirical depth. Its value lies not in completeness but in providing a shared vocabulary through which human and artificial minds can be compared, integrated, and bounded within the same clinical scene.

# 5. Temporality in Clinical Care

Clinical cognition does not occur in isolated snapshots but unfolds through sequences of interaction over time[17]. The CDM and PDM models describe single cognitive cycles; actual clinical care involves multiple cycles that interleave between patient and provider across seconds, minutes, hours, days, months, or years. The characteristic structure is an iteration loop in which patient and provider alternate cognitive cycles while drawing upon the ecosystem. A patient shares information or expresses a preference. The provider retrieves relevant data, processes the combined input, and responds by sharing information or finalizing a plan. The patient processes this response, and either communicates a refined preference or seeks additional information. This alternation continues until both parties reach sufficient closure, though the definition of sufficient varies with the clinical context and the stakes of the decision. Each node in the sequence, whether PDM or CDM, represents a complete cognitive cycle whose action output becomes the input state for the next, extending across pre-encounter, encounter, and post-encounter phases (**Figure 5**).

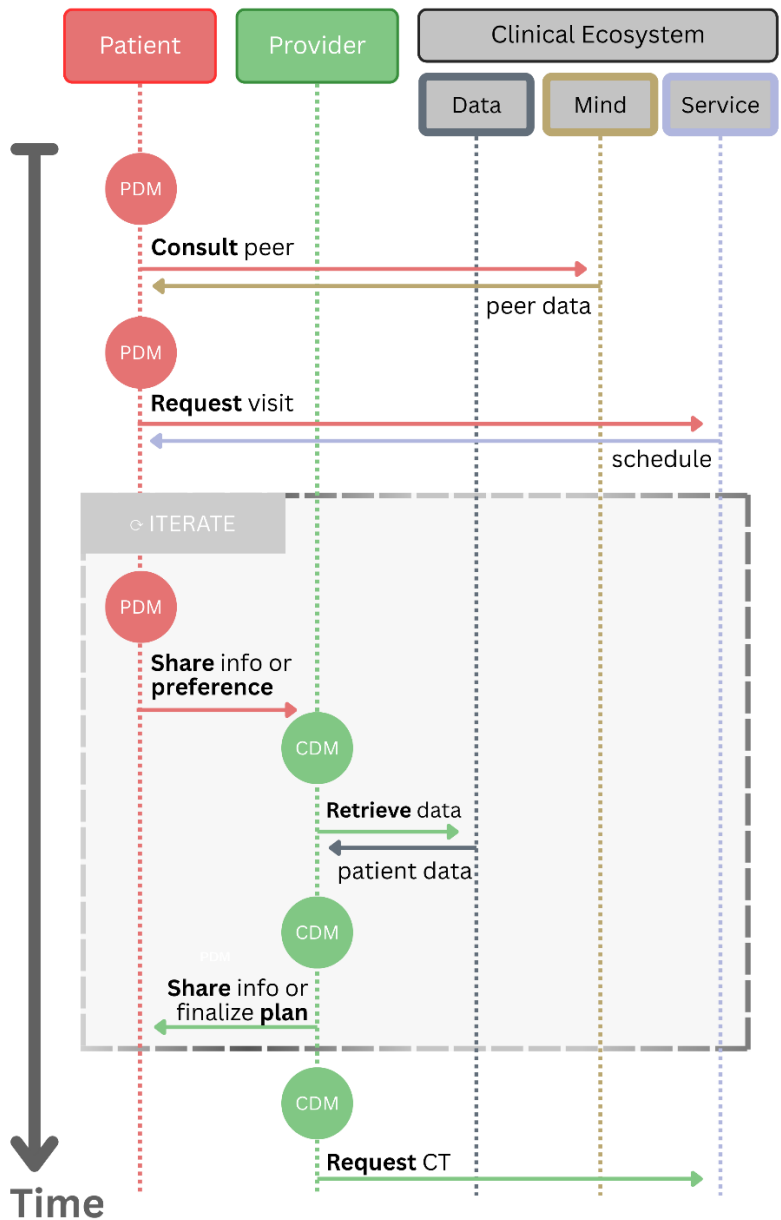

**Figure 5. Temporality in the Clinical World Model.** Sequence diagram showing alternating PDM (red) and CDM (green) cycles across a clinical episode. In the pre-encounter phase, the patient consults a peer and requests a visit, both mediated through the clinical ecosystem. Within the encounter, an ITERATE loop captures the characteristic alternation of clinical interaction: the patient shares information or expresses a preference, the provider retrieves data from the ecosystem, and the provider responds by sharing information or finalizing a plan. This cycle repeats until both parties reach sufficient closure. In the post-encounter phase, the provider requests further investigation through the ecosystem. Each node represents a complete cognitive cycle as formalized in earlier sections, with the action output of one cycle becoming the input state for the next.

Yet the iteration loop conceals a structural asymmetry. The patient experiences illness continuously; the provider samples it at discrete intervals. Between encounters, PDM cycles continue to run: symptoms evolve, adherence decisions are made and revised, and the world state drifts without any corresponding CDM cycle to observe it. This interval is where most clinical deterioration occurs and where silent complications advance beyond timely intervention[77,78]. The temporal architecture of conventional care is one of periodic observation punctuated by clinical blindness, a sampling problem that no amount of within-encounter reasoning can resolve.

The introduction of an agentic AI system into this temporal structure changes it in ways unprecedented in clinical care. Unlike the provider, whose CDM cycles activate when the encounter begins and pause when it ends, an AI agent is not inherently episodic. It can persist across real-time, continuously monitoring patient-reported data, updating its representation of the evolving clinical state, and initiating ADM cycles when its trajectory projection detects divergence from the expected course. This transforms the temporal architecture from discrete sampling to something approaching continuous observation. For the first time, a reasoning agent within the ecosystem can occupy the temporal gaps.

As a clinical illustration, consider a patient who presents with a new symptom at a scheduled routine visit. The encounter begins with a PDM cycle: the patient describes the symptom, its onset, and their concern. A CDM cycle follows as the provider processes this input, retrieves relevant history, and forms an initial assessment. An ADM cycle then enters the sequence. The AI agent receives the same clinical input, integrates it with longitudinal data accumulated between encounters, and generates a differential assessment. Rather than acting autonomously, the agent selects Request Confirmation, returning control to the provider while making its reasoning visible. This triggers a new CDM cycle in which the provider evaluates the AI's output against their own judgment, accepts or overrides it, and communicates a plan to the patient. The patient weighs the plan against their own values through a subsequent PDM cycle, and either consents or raises a concern that initiates another pass through the loop. The ADM cycles can interleave with human cycles, its outputs become inputs for subsequent human reasoning, and its authority is bounded by the same logic of closure that governs the encounter.

This temporal capability, however, is not uniform across clinical tasks. A triage decision collapses into a single cycle measured in seconds. A diagnostic workup unfolds across several cycles over days. A chronic disease management strategy must project across months or years of the anticipated trajectory. The number of cycles over which an agent must maintain coherent reasoning, and how far forward its trajectory projection must extend, varies with the clinical task it performs. Temporal horizon is therefore not a fixed property of the agent but a designable parameter that should be specified alongside disease phase, care task, care setting, and context. A framework that treats all agentic deployments as temporally equivalent fails to distinguish a real-time alert system from a longitudinal care coordinator, two applications that differ less in computational substrate than in the temporal depth of reasoning they require.

## 6. Clinical AI Skill-Mix

The Clinical World Model provides the reference world model for the architecture of clinical care. The CDM and PDM formalize the dynamics of human cognition, while the Clinical AI Skill-Mix operationalizes these constructs for the practical specification of intelligent systems within a given Mandate. We define the capability space through two complementary constructs. The *Clinical Competency Space* specifies the clinical scenario through five dimensions, and the A*I Cognitive Engagement Space* specifies how AI participates in human reasoning through three. Together, these eight dimensions capture the contextual boundaries that make each combination a distinct competency (**Figure 6**). Structured JSON specifications for each dimension and its corresponding values are provided in **Supplementary Data 1**.

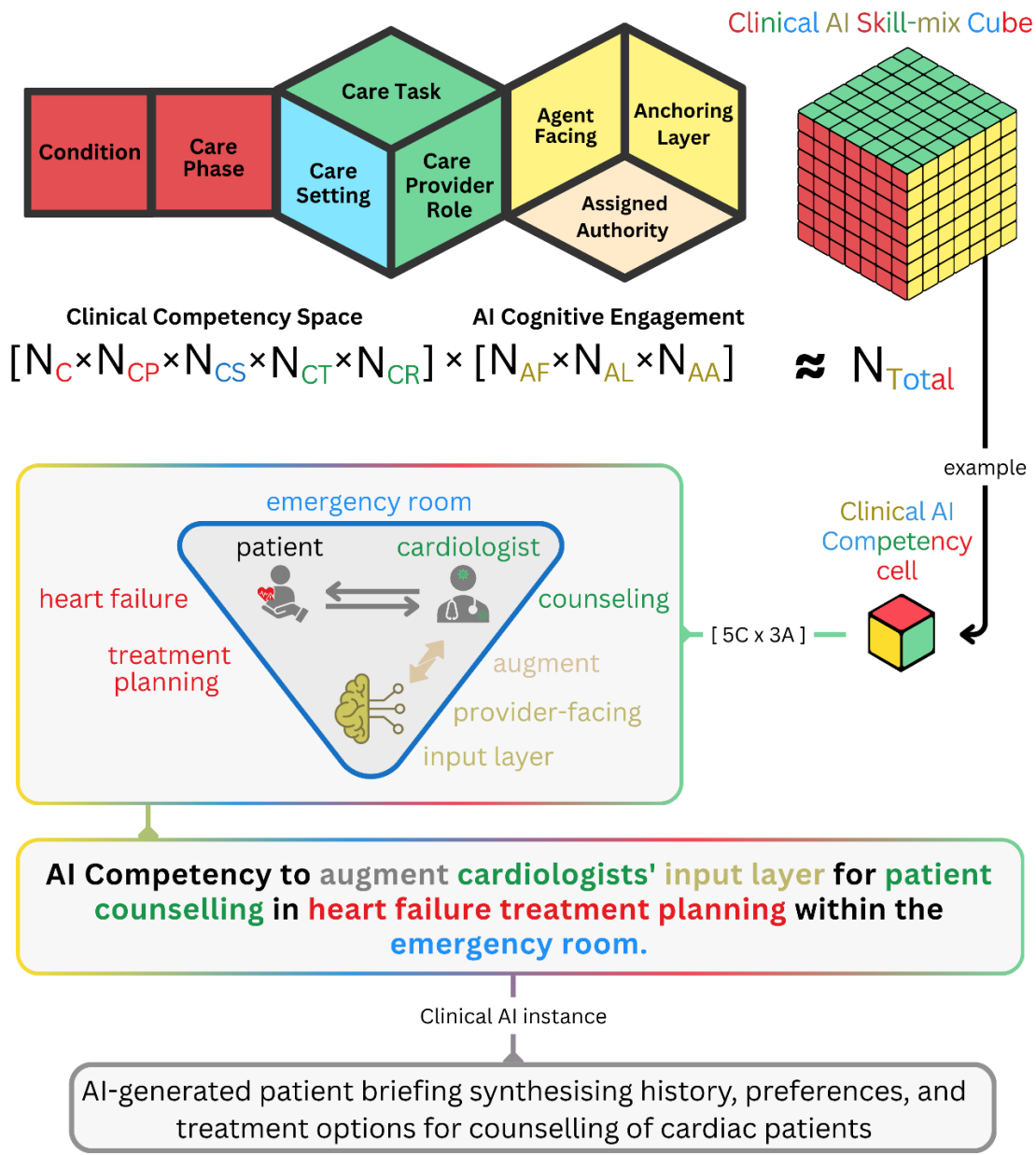

**Figure 6. The Clinical AI Skill-mix Cube for comprehensive competency specification.** The upper section shows the eight dimensions that define a clinical AI competency cell. The Clinical Competency Space comprises five dimensions (Condition, Care Phase, Care Setting, Care Task, Care Provider Role, shown in red and green). The AI Cognitive Engagement comprises three dimensions (Agent Facing, Anchoring Layer, and Assigned Authority, shown in yellow). The product of cardinalities across combinations ($N_C \times N_{CP} \times N_{CS} \times N_{CT} \times N_{CR} \times N_{AF} \times N_{AL} \times N_{AA}$) yields the total number of possible competency cells ($N_{Total}$), represented as the Clinical AI Skill-mix Cube. The lower section illustrates a concrete example: an AI competency cell [5C × 3A] specification, along with an example Clinical AI instance with this competency.

## 6.1 Clinical Competency Space

Five dimensions collectively define the clinical context in which an AI system operates. These dimensions capture the contextual variables that determine whether capabilities developed in one setting can be effectively transferred to another.

**Condition**

Condition specifies the health state being addressed, whether disease, syndrome, injury, or risk state. The term "Condition" rather than "Disease" follows recent international classification guidance[79], recognizing that clinical practice addresses states beyond classical disease categories. We operationalize Conditions using ICD-10-CM three-character codes (1,918 codes, spanning for example I21 for acute myocardial infarction and F32 for major depressive episode), whose hierarchical structure supports both granular specification and population-level prioritization; alternative taxonomies such as the Global Burden of Disease framework offer a complementary lens oriented toward epidemiological impact[80]. Because patients can present with multiple concurrent conditions and clinicians often encounter symptoms before diagnoses, a Condition may specify either a disease entity or a presenting complaint, and systems must be evaluated independently for each Condition claimed.

**Care Phase**

Care Phase represents temporal position within the illness trajectory. Building on the World Health Organization (WHO) continuum of integrated health services[81] and SEIPS framework[16], we characterize disease progression through seven milestones spanning from health through pathologic process, illness manifestation, diagnosis, treatment, follow-up, and ultimately to the final states of cure, disability, or death. These milestones correspond to seven actionable phases: at-risk identification, pre-symptomatic detection, diagnostic workup, treatment planning, post-treatment care, follow-up, and coping support. Importantly, patients may occupy multiple positions simultaneously, as when receiving treatment for a primary condition while undergoing diagnostic workup for emerging complications. **Figure 7** illustrates the integration of the Care Phase and Care Setting, demonstrating how patients may transition between settings as they progress through their illness trajectory.

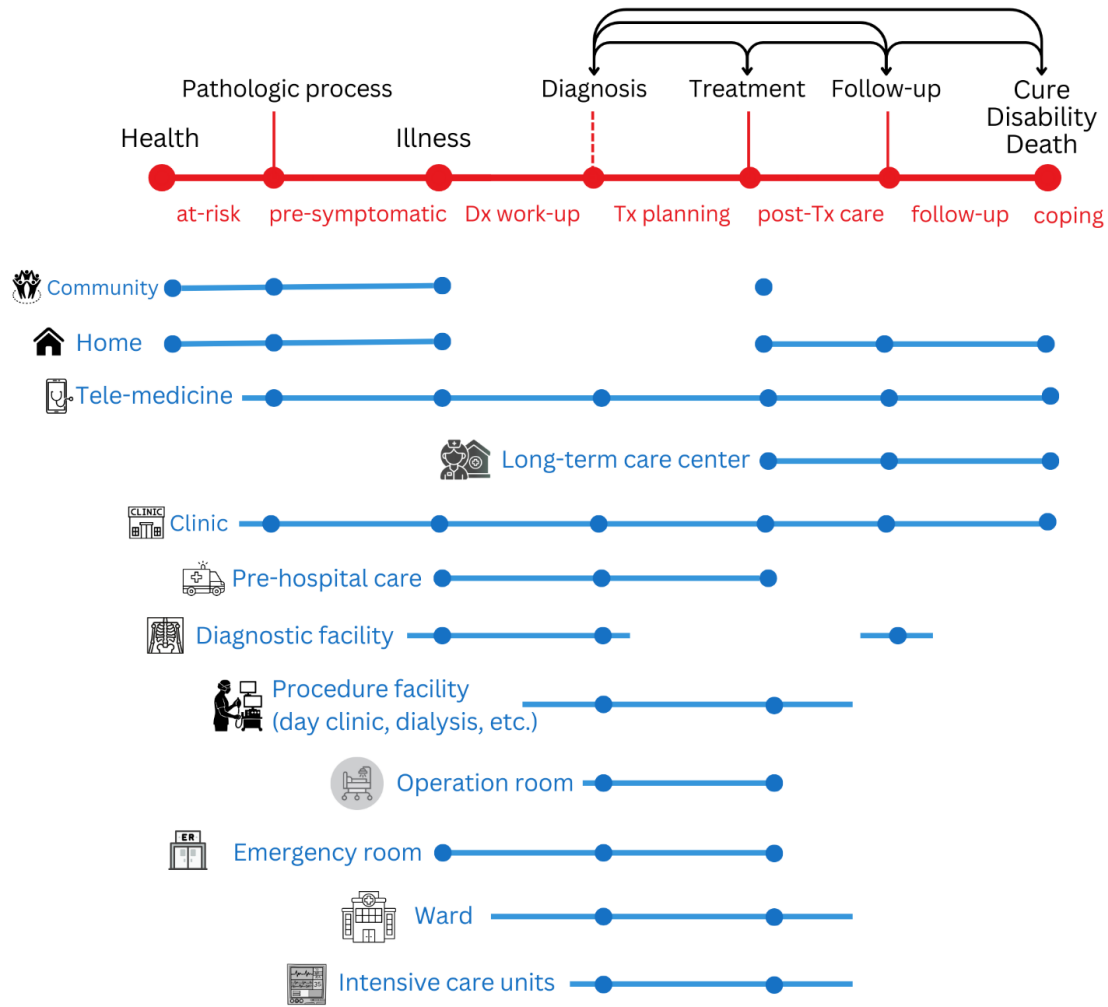

**Figure 7. Care Phase and Care Setting dimensions of the Clinical Competency Space.** The upper timeline depicts seven Care Phases spanning health to outcome: at-risk, pre-symptomatic, diagnostic work-up, treatment planning, post-treatment care, follow-up, and coping. The lower section maps 12 Care Settings to the timeline, with blue lines indicating each setting's temporal relevance across phases. Abbreviations: Dx, diagnostic; Tx, treatment.

## Care Setting

Care Setting is intrinsically linked to Care Phase, as different phases typically occur in distinct settings ranging from community screening to emergency departments, inpatient wards, intensive care units, rehabilitation facilities, and home-based care. We integrate these two dimensions into a framework (**[Figure 6](#)**) that maps the patient's journey across both time and space. Our locations were defined inspired by the SNOMED-CT activity location, which is a comprehensive clinical terminology standard that includes environments and geographical locations hierarchy, enabling standardized coding of healthcare settings for interoperability across electronic health records[82]. An alternative approach employs the SSEIPS framework, which has developed structured representations of patient journeys across care settings[16].

**Care Provider Role**

The Care Provider Role specifies the healthcare worker whose reasoning system is designed to support. We adopt the WHO International Classification of Health Workers[83], yielding 86 roles spanning health professionals, health associate professionals, personal care workers, health management personnel, and other health service providers, of which 36 represent physician specialty roles. Different roles carry different knowledge bases and cognitive demands. Competency benchmarked against specialist physicians differs fundamentally from competency benchmarked against community health workers addressing the same condition.

**Care Task**

Care Task specifies the clinical activity being addressed. To ensure alignment between AI capabilities and the authentic work of healthcare providers, we ground this dimension in established physician competency frameworks. The Physician Competency Reference Set[29], developed through synthesis of over 150 competency lists across health professions, provides a unifying taxonomy of 58 competencies organized into eight domains: Patient Care, Medical Knowledge, Practice-Based Learning and Improvement, Interpersonal and Communication Skills, Professionalism, Systems-Based Practice, Interprofessional Collaboration, and Personal and Professional Development. An alternative approach, particularly relevant for evaluating LLMs, is the MedHELM framework[18], which organizes 121 medical tasks into five categories validated by clinicians: Clinical Decision Support, Clinical Note Generation, Patient Communication and Education, Medical Research Assistance, and Administration and Workflow.

## 6.2 AI Cognitive Engagement

Three dimensions define how AI participates in human reasoning formalized by the CDM and PDM models: Assigned Authority, Agent Facing, and Anchoring Layer.

**Assigned Authority**

Assigned Authority specifies the degree to which AI assumes cognitive processing at the anchored layer, and every competency claim must declare its assigned authority explicitly. Three modes define this spectrum. In *monitoring*, AI observes reasoning processes and provides feedback for quality assessment or learning without directly influencing current decisions. In *augmentation*, AI enhances human processing while humans retain cognitive primacy, with AI outputs entering reasoning as one input among others. In *automation*, AI replaces human processing at that layer, with humans verifying retrospectively rather than performing concurrent cognitive work.

A single system may warrant automation authority for one anchored layer while remaining appropriate only for augmentation at another; authority is therefore a property of the individual competency claim, not of the system as a whole. This declaration serves a dual function: it defines the regulatory boundary of accountability for each claimed competency, and it calibrates the user's expected cognitive posture, establishing whether the human agent should reason actively alongside the system or shift to supervisory oversight. Critically, the declaration also creates a design obligation: the system must present its outputs in a manner consistent with the declared mode, since interface choices, confidence framing, and workflow integration can shift the effective authority experienced by the user regardless of the formal label. The implications of authority mode for evaluation thresholds, design decisions, and deskilling risk are developed in Section 7.1.

**Agent Facing**

Agent Facing specifies the reasoning entity whose cognition the AI engages. Four options exist: *Provider-facing*, where AI engages with provider reasoning as formalized in the CDM model; *Patient-facing*, where AI engages with patient reasoning as formalized in the PDM model; *Encounter-facing*, where AI engages with the dyadic reasoning process, facilitating shared decision-making or mediating communication between the provider and the patient; and *Ecosystem-facing*, where AI engages institutional or infrastructural endpoints that possess no cognitive model of their own.

Ecosystem-facing interactions may target any of the ecosystem's three functional components: Data, as when an agent writes to or retrieves from an electronic health record; Mind, as when an agent routes a consultation request to another agent or to a specialist panel; or Service, as when an agent dispatches an ambulance or transmits a reportable condition to a public health authority. Our framework emphasized Provider-facing applications while developing the PDM architecture as a parallel framework for Patient-facing AI and recognizing Encounter-facing and Ecosystem-facing AI as an emerging area for future work.

## Anchoring Layer

The Anchoring Layer specifies the point within the cognitive architecture at which AI engages human reasoning. Seven layers correspond to sequential stages of the CDM and PDM models: *Input*, *Data Processor*, *Hypothesis*, *System I*, *System II*, *Reflection*, and *Action* (**Table 2**). A critical distinction applies to System I engagement: while AI can perform pattern recognition functionally parallel to human intuition, its outputs necessarily surface to conscious awareness for System II evaluation rather than integrating into preconscious processing. AI "augmenting" System I thus operates as a parallel processor whose outputs clinicians must explicitly consider—fundamentally differing from how human intuition operates beneath awareness. This asymmetry means that AI-generated predictions, however rapid, add to the clinician's cognitive load rather than bypassing it.

Layer selection determines error propagation and capability requirements. Earlier layers (Input, Data Processor) shape the informational substrate for all downstream reasoning; errors here cascade forward. Central layers (Hypothesis, System I, System II) directly influence conclusions, demanding higher accuracy and interpretability. Later layers (Reflection, Action) affect implementation while preserving the integrity of antecedent reasoning. For Patient and Encounter agents, layers map to equivalent cognitive stages.

**Table 2. Anchoring Layer Mapping Across Different Agent-Facings with Examples.** Rows represent the seven anchoring layers; columns map each layer to provider-facing (interacts with CDM), patient-facing (interacts with PDM), and encounter-facing applications. Ecosystem-facing, introduced in this framework, is omitted because its layer mappings do not follow a cognitive architecture and require future elaboration.

| Layer | Function | Clinician (CDM) | Patient (PDM) | Encounter |
|---|---|---|---|---|
| Input | Data acquisition, presentation, filtering | Clinical data gathering | Health information reception | Information exchange |
| Data Processor | Abstraction, contextualization, and knowledge retrieval | Evidence-guided synthesis | Personal context integration | Shared understanding |
| Hypothesis | Candidate explanation generation | Differential formulation | Condition comprehension | Mutual problem definition |
| System I | Rapid pattern-based evaluation | Illness script activation | Experiential pattern matching | Trust recognition |
| System II | Deliberate analytical evaluation | Hypothetico-deductive analysis | Option deliberation | Shared deliberation |
| Reflection | Confidence calibration, reasoning monitoring | Metacognitive monitoring | Decision confidence | Alignment verification |
| Action | Decision implementation | Care plan execution | Preference expression | Concordant plan |

## 6.3 The Skill-Mix Cube

The intersection of Clinical Competency Space and AI Cognitive Engagement yields a multidimensional space termed the Clinical AI Skill-mix. Each cell constitutes specification of what clinical scenario AI addresses and how it participates in human reasoning. The formula $N_C \times N_{CP} \times N_{CS} \times N_{CT} \times N_{CR} \times N_{AF} \times N_{AL} \times N_{AA}$ approximates the total number of distinct Clinical AI Competencies within the combinatorial space. Applying the taxonomies defined in this framework (1,918 conditions, 7 care phases, 12 care settings, 58 care tasks, 86 provider roles), the total complexity of clinical practice may approach billions of unique scenarios. For any single provider role, such as general physician or cardiologist, this reduces to millions of clinical competencies. Crossing this clinical space with 84 AI engagement patterns produces a combinatorial total that remains vast even after excluding clinically irrelevant combinations.

**Figure 6** illustrates how the eight dimensions combine to define a precise Clinical AI Competency covered by an AI tool (heart failure, treatment planning, emergency room, counseling, augment, cardiologist, provider-facing, input layer). However, a single cell need not correspond to a single tool. A capable foundation model or a set of coordinated agents can traverse multiple cells to deliver end-to-end clinical value, and the boundaries between cells define the precise points where inter-agent hand-offs must preserve cognitive compatibility.

Consider a patient with a history of paroxysmal atrial fibrillation wearing a continuous electrocardiography patch during the long-term management phase. An agentic system monitoring the device acquires the signal continuously at the Input layer under automation authority. When the system detects an abnormal rhythm, it abstracts the trace into clinically structured features at the Data Processor layer, still operating autonomously. At this boundary, the first inter-agent hand-off occurs: the output of the signal-processing agent constitutes a typed interface whose structured abstraction must match what a downstream classification agent expects as input. Classification of the abnormality into an urgency category requires this second agent to operate at the Hypothesis layer under augmentation authority, because the resulting differential directly informs the clinician's reasoning rather than bypassing it. The system transmits a clinical summary to the patient's practitioner, a provider-facing action that may warrant automation for routine alerts but augmentation when the classification carries consequence.

In parallel, it prompts the patient to report recent symptoms and medication adherence, a patient-facing augmentation that supports history collection. Upon provider confirmation, the system prepares a structured clinical context and dispatches an ambulance, an ecosystem-level action that invokes the ADM's Request Confirmation mechanism before escalating to an irreversible intervention. This branching point constitutes another hand-off, routing from a single provider-facing reasoning chain into parallel patient-facing and ecosystem-level agents, each carrying a distinct authority declaration and accountability structure.

Overall, this single agentic system spans at least seven distinct Skill-Mix cells across one condition, two care phases (long-term management transitioning to acute intervention), two care settings (ambulatory and emergency), six care tasks, and two provider roles, with AI engagement patterns covering three agent-facing modes (provider, patient, and ecosystem), four anchoring layers (Input, Data Processor, Hypothesis, and Action), and three authority modes (monitoring, automation and augmentation). The Skill-Mix Cube makes this internal heterogeneity visible, exposing where authority shifts and where cognitive compatibility must be maintained across hand-offs.

# 7. Implications for Clinical AI

## 7.1 Implications for Evaluation

Evaluating intelligent systems requires distinguishing three related but distinct constructs: intelligence, competency, and performance[84]. Intelligence refers to raw cognitive capacity, encompassing the ability to learn, reason, and adapt to novel situations. Competency describes what an agent can do in a specific domain given adequate conditions, representing acquired capability that exists whether or not it is currently being demonstrated. Performance captures what actually manifests under real-world constraints, reflecting competency filtered through context, available resources, and environmental factors[85,86]. Performance and competency can diverge in either direction: a highly competent system may show low performance under resource constraints or distribution shift, just as a low-performance system may appear adequate when operating within narrow competency that match its training.

These distinctions have significant implications for the evaluation of clinical AI. **Supplementary Table 3** reviews existing frameworks for classifying AI system intelligence, addressing the underlying capacity for reasoning and adaptation through dimensions such as how models act autonomously[87], learn patterns[88], and solve complexity[89].

Recent systematic work has synthesized these requirements into seven complementary clusters named the IMPACTS Framework[90]: integration and workflow compatibility, monitoring and governance structures, performance quality metrics, acceptability and trust factors, cost and economic sustainability, technological safety and transparency, and scalability with clinical impact. Detailed specification of metrics and thresholds within each cluster exceeds the present scope but represents essential work for framework operationalization.

An additional dimension, recognizing AI as an ecosystem component that interacts with human cognition, is autonomy preservation[91]. This evaluates whether Clinical AI maintains meaningful human agency through appropriate override design, mitigates deskilling risk from prolonged reliance, and protects clinical skill development[91]. The dimension becomes increasingly critical as systems assume greater authority.

The Clinical World Model provides for evaluation specification by establishing that requirements vary systematically across Skill-Mix coordinates. Automation demands higher performance thresholds than augmentation; central reasoning layers require greater transparency than peripheral layers; evidence sufficient for one care setting may require extension before generalization to another. These interactions mean evaluation must be defined for each Clinical AI Competency cell rather than at the system level. Development then follows a natural progression from narrow, well-characterized applications toward broader deployment as evidence accumulates across an expanding scope.

This cell-specific approach reframes evaluation from whether a system performs adequately to in which competency coordinates adequacy has been demonstrated. Validity evidence must progress through internal validation on development data, external validation within the intended deployment setting, and prospective validation under actual clinical conditions[92]. Threshold determination for each cell requires stakeholder consensus integrating clinical expertise, patient perspectives, regulatory requirements, and institutional risk tolerance. Finally, system readiness emerges as a property of competency coverage, enabling calibrated trust through an explicit mapping of demonstrated capabilities to a bounded deployment scope.

## 7.2 Implications for Design

The framework provides principled criteria for a fundamental design decision: whether AI should augment human reasoning or automate clinical functions. The cognitive architecture formalized in the CDM and PDM models directly informs this design choice. System I tasks, characterized by pattern recognition within established illness scripts, suit automation when relevant training data can be feasibly collected. System II tasks, requiring deliberation over novel presentations, value tradeoffs, or patient preferences, require augmentation because such decisions involve moral judgment and accountability that must remain with human agents[50]. Tasks dependent on tacit clinical intuition may remain unsuitable for automation regardless of performance metrics, because such knowledge resists encoding entirely[50].

Three considerations guide design decisions within specific Clinical Skill-Mix cells. Encodability: explicitly representable knowledge, such as guidelines, protocols, and diagnostic criteria, supports automation, while tacit expertise favors augmentation[70]. Reversibility: irreversible or high-stakes decisions warrant greater human oversight, aligning with phased certification pathways that grant autonomy incrementally. Relational requirements: tasks in which therapeutic goals depend on human connection remain appropriate for augmentation even when technically automatable.

The PDM model introduces distinct constraints for patient-facing applications. Patient decisions depend on trust, communication quality, and the translation of clinical information into personally meaningful representations[93]. Systems designed for patient-facing roles must support rather than supplant the clinician-patient dialogue, enhancing information exchange without disrupting relational dynamics essential to preference formation[93]. Administrative and documentation functions impose fewer relational demands than diagnostic conversations or treatment discussions, creating natural boundaries for automation scope.

Beyond driving design, the framework offers institutional utility for standardizing competency disclosure. At the point of AI-assisted reporting in radiology, pathology, or clinical documentation, the Clinical Skill-Mix provides vocabulary for communicating system competency alongside outputs, paralleling the model card concept for transparent AI documentation[94]. Institutions adapting standardized disclosure can systematically track performance across cells, identifying where AI adds value, where gaps persist, and where targeted development would yield the greatest benefit.

## 7.3 Implications for Agentic Systems

As clinical AI evolves toward multi-agent architectures, the Skill-Mix cell becomes the natural unit of agent decomposition[74]. Each cell functions as a typed interface specification grounded in clinical cognition: an agent anchored at the Data Processor layer outputs abstractions that an agent at the Hypothesis layer expects as input. This semantic typing distinguishes clinically principled orchestration from ad hoc task routing, because inter-agent handoffs must preserve cognitive compatibility across layers rather than merely pass data[95,96]. The cell specification further constrains agent configuration, defining permissible tool access, behavioral instructions, and output format so that each agent's operational envelope is derived from its demonstrated clinical scope.

The anchoring layer determines what tools an agent requires and what authority it may exercise. An input-layer agent retrieves laboratory values and clinical notes, operating appropriately in monitoring mode where errors remain correctable downstream. A hypothesis-layer agent accesses knowledge bases to generate differentials, requiring augmentation authority because its outputs directly enter the clinician's deliberative process. An action-layer agent interacting with order entry systems may warrant automation for structured tasks but augmentation when actions carry therapeutic consequence.

When agents are chained, authority does not compose predictably[97]. A clinician who engages with an augmentation-level agent at the input layer may be unaware that a downstream agent operates in automation mode at the hypothesis layer, producing a composite chain whose effective authority exceeds what any individual agent was certified for[98,99]. The framework makes this escalation structurally visible by requiring per-cell authority declarations, exposing the precise point where the cognitive posture expected of the human shifts. This visibility can suggest a principled guardrail: no composite chain should present outputs at an authority level exceeding the most restrictive cell in the sequence without explicit governance that accounts for the emergent behavior of the whole. Operationally, the ADM's Request Confirmation (Section 4) extends from a human-in-the-loop mechanism to an orchestration primitive, routing at competency boundaries to a human or to another agent with demonstrated competency in the adjacent cell[98].

By mapping each agent to its validated Skill-Mix cells, institutions can generate structured disclosures of what a multi-agent system can and cannot do, extending the competency disclosure introduced in Section 7.2 from individual models to composite workflows[94]. As agent configurations evolve and evidence accumulates across new cells, these disclosures update accordingly, providing a living record of demonstrated capability bounded by the framework's dimensional structure.

## 8. Discussion

The Clinical World Model and Skill-Mix framework help explain why clinical AI development, deployment, and regulation have remained challenging[8]. Medicine encompasses myriads of conditions across multiple care phases, settings, provider roles, and tasks. Combined with dimensions specifying how AI engages human cognition, the combination yields a billion-scale competency space, in which validation in one cell provides minimal evidence for another. No prior framework has captured this scale; most have condensed it through necessary simplification, though at the cost of obscuring the space. By rendering this structure, the framework makes legible what clinicians have long intuited, that even a highly accurate single-task AI, covering only a fraction of a single competency cell, remains far from the range of competencies required for meaningful clinical action.

Current AI development emphasizes architectural complexity and training scale, yet this approach may not resolve the fundamental problem. Without alignment-by-design that specifies where AI engages cognition and with what authority, providers can only fall back on whether AI outputs agree with their own assessment. This strategy breaks down when AI detects patterns the clinician has no prior experience to evaluate, and over time risks eroding clinical confidence and skill through uncritical reliance on AI outputs[100]. The Anchoring Layer and Assigned Authority dimensions address this gap directly, providing vocabulary for both deployment specification and responsibility bounding[101]. Regulators can use the Assigned Authority to distinguish the decision owner accountable for misjudgment, while providers can use the Anchoring Layer to understand how AI outputs should enter their cognitive process. Recent FDA guidance on AI-enabled devices, for instance, mandates lifecycle oversight and change control plans that could be operationalized and communicated through this shared language[102].

The framework accommodates both augmentation and automation paradigms. While grounded in clinical augmentation, we recognize that some competency cells may benefit from fully automated pipelines while others require human oversight. The Clinical World Model provides structure for either scenario and, importantly, for emerging multi-agent systems where Human-AI and AI-AI interactions require principled orchestration[103]. The paramount consideration remains community health and patient care rather than ideological commitment to either paradigm.

Several limitations warrant acknowledgment. Framework development require more diverse inclusion for maturation, and future iterations should incorporate broader stakeholder input. Contextual adaptation at the national level remains essential to align the model with local biopsychosocial norms and regulatory structures. Additionally, while we have articulated the theoretical architecture of the Clinical World Model, empirical validation through prospective application to specific clinical AI systems remains necessary to assess its practical utility. Theoretical extensions should address multimodal foundation models that blur traditional task boundaries, the role of AI in reshaping clinical education, and the ethical implications of systems that may exceed human performance while remaining epistemically opaque.

The PDM model presented here reflects a conception of the patient as an empowered and autonomous agent, consistent with modern Western medical ethics and patient-centered care[71]. However, decisional authority between patients, families, and providers varies substantially across cultures and clinical contexts[73]. The framework accommodates this variability through configuration of its inputs and information flows rather than architectural modification, with the autonomous patient serving as the reference case while recognizing that clinical reality encompasses a spectrum of arrangements.

The central contribution of this work lies in reframing the field's longstanding question of whether clinical AI works. The more constructive question is in which competency coordinates a system has demonstrated reliability, and for whom. A framework's value will ultimately be determined not by theoretical completeness but by whether it enables a common language across stakeholders. Clinical World Model and Clinical AI Skill-Mix offer such grammar. Clinical AI will continue to evolve, and the relevant question is whether that evolution remains sustainable and aligned with patient welfare.

## Acknowledgements

We thank Drs Ali Safavi-Naini, Nariman Naderi, Aref Mahjob, Mohammad Mahdi Mollahasani, and Sayeh Jalali for their helpful feedback on earlier draft. Large language model (Claude Opus 4.6) was used to assist with language editing, exploration, and prose refinement. All content was critically reviewed, verified, and revised by the authors, who take full responsibility for the final manuscript.

## Authors' Contribution

SAASN conceptualized the study, developed the theoretical framework, wrote the original draft, and created the visualizations and code. EM contributed substantially to conceptualization, framework development, and manuscript writing. AS and IS provided pivotal intellectual input during conceptualization and methodology development. JM, PMK, GS, and SW contributed to the investigation and critically reviewed the manuscript. ZA, PRL, MR, GN, RW, and CG validated the framework and critically reviewed and edited the manuscript. AS supervised the study and reviewed and edited the manuscript. IS supervised the project and reviewed and edited the manuscript. All authors read and approved the final manuscript.

## Competing Interests

All authors declare no financial or non-financial conflict of interest related to this work.

## Data Availability

Machine-readable JSON specifications for all eight Clinical AI Skill-Mix dimensions are provided in **Supplementary Data 1** and are publicly available at https://github.com/Sdamirsa/Clinical-World-Model/tree/main/clinical-skill-mix. The five Clinical Competency dimensions (Condition, Care Phase, Care Setting, Clinical Task, and Care Provider Role) and three AI Cognitive Engagement dimensions (Agent Facing, Anchoring Layer, and Assigned Authority) are each provided as individual JSON files within the repository.

This study used three publicly available reference datasets. Medical conditions were derived from the International Classification of Diseases, 10th Revision, Clinical Modification (ICD-10-CM) (https://www.cms.gov/medicare/coding-billing/icd-10-codes). Care provider roles were derived from the WHO International Standard Classification of Health Workers (https://www.who.int/publications/m/item/classifying-health-workers). Cognitive care tasks were derived from the Physician Competency Reference Set (Englander et al., 2013, doi: 10.1097/ACM.0b013e31829a3b2b).

## Code Availability

Code for generating outputs is available at https://github.com/Sdamirsa/Clinical-World-Model/. An interactive interface for exploring the Clinical World Model, Skill-Mix dimensions, CDM, PDM, and individual competency cell combinations is available at https://sdamirsa.github.io/Clinical-World-Model/.

# 9. References

1. Gallifant, J. & Bitterman, D. S. Humanity's next medical exam: preparing to evaluate superhuman systems. *NEJM AI* **2**, AIe2501008 (2025).

2. Tikhomirov, L. *et al.* Medical artificial intelligence for clinicians: the lost cognitive perspective. *Lancet Digit. Health* **6**, e589–e594 (2024).

3. Azad, T. D., Krumholz, H. M. & Saria, S. Principles to guide clinical AI readiness and move from benchmarks to real-world evaluation. *Nat. Med.* 1–3 (2026) doi:10.1038/s41591-025-04198-1.

4. Kunst, M. *et al.* Real-world performance of large vessel occlusion artificial intelligence–based computer-aided triage and notification algorithms—what the stroke team needs to know. *J. Am. Coll. Radiol.* **21**, 329–340 (2024).

5. Yu, A. C., Mohajer, B. & Eng, J. External validation of deep learning algorithms for radiologic diagnosis: a systematic review. *Radiol. Artif. Intell.* **4**, e210064 (2022).

6. Blagec, K., Kraiger, J., Frühwirt, W. & Samwald, M. Benchmark datasets driving artificial intelligence development fail to capture the needs of medical professionals. *J. Biomed. Inform.* **137**, 104274 (2023).

7. Wong, A. *et al.* External validation of a widely implemented proprietary sepsis prediction model in hospitalized patients. *JAMA Intern. Med.* **181**, 1065–1070 (2021).

8. Arab, R. A. E. *et al.* Bridging the gap: from AI success in clinical trials to real-world healthcare implementation—a narrative review. *Healthcare* **13**, (2025).

9. Safavi-Naini, S. A. A. *et al.* Benchmarking proprietary and open-source language and vision-language models for gastroenterology clinical reasoning. *Npj Digit. Med.* **8**, 797 (2025).

10. Singh, A., Ehtesham, A., Kumar, S. & Khoei, T. T. A survey on large concept models: advancing applications in healthcare, finance, and education. in *2025 IEEE World AI IoT Congress (AIIoT)* 0971–0975 (2025). doi:10.1109/AIIoT65859.2025.11105333.

11. Vatsal, S., Dubey, H. & Singh, A. Agentic AI in healthcare and medicine: a seven-dimensional taxonomy for empirical evaluation of LLM-based agents. *IEEE Access* **14**, 4840–4863 (2026).

12. Moritz, M., Topol, E. & Rajpurkar, P. Coordinated AI agents for advancing healthcare. *Nat. Biomed. Eng.* **9**, 432–438 (2025).

13. Zhu, K. *et al.* Where LLM agents fail and how they can learn from failures. Preprint at https://doi.org/10.48550/arXiv.2509.25370 (2025).

14. Atf, Z. *et al.* The challenge of uncertainty quantification of large language models in medicine. Preprint at https://doi.org/10.48550/arXiv.2504.05278 (2025).

15. Stiggelbout, A. M., Pieterse, A. H. & De Haes, J. C. J. M. Shared decision making: concepts, evidence, and practice. *Patient Educ. Couns.* **98**, 1172–1179 (2015).

16. Carayon, P., Wooldridge, A., Hoonakker, P., Hundt, A. S. & Kelly, M. M. SEIPS 3.0: human-centered design of the patient journey for patient safety. *Appl. Ergon.* **84**, 103033 (2020).

17. Yazdani, S. & Hoseini Abardeh, M. Five decades of research and theorization on clinical reasoning: a critical review. *Adv. Med. Educ. Pract.* **10**, 703–716 (2019).

18. Bedi, S. *et al.* MedHELM: holistic evaluation of large language models for medical tasks. Preprint at https://doi.org/10.48550/arXiv.2505.23802 (2025).

19. Carayon, P. *et al.* Work system design for patient safety: the SEIPS model. *Qual. Health Care* **15**, i50–i58 (2006).

20. Holden, R. J. *et al.* SEIPS 2.0: a human factors framework for studying and improving the work of healthcare professionals and patients. *Ergonomics* **56**, 1669–1686 (2013).

21. Rademakers, F. E. *et al.* CORE-MD clinical risk score for regulatory evaluation of artificial intelligence-based medical device software. *Npj Digit. Med.* **8**, 90 (2025).

22. Chang, Q. *et al.* 2025 expert consensus on retrospective evaluation of large language model applications in clinical scenarios. *Intell. Med.* **5**, 318–330 (2025).

23. Hong, S., Xiao, L., Zhang, X. & Chen, J. ArgMed-agents: explainable clinical decision reasoning with LLM disscusion via argumentation schemes. in *2024 IEEE International Conference on Bioinformatics and Biomedicine (BIBM)* 5486–5493 (2024). doi:10.1109/BIBM62325.2024.10822109.

24. Yang, R. *et al.* Toward global large language models in medicine. Preprint at https://doi.org/10.48550/arXiv.2601.02186 (2026).

25. Yan, W. *et al.* ClinicalLab: aligning agents for multi-departmental clinical diagnostics in the real world. Preprint at https://doi.org/10.48550/arXiv.2406.13890 (2024).

26. Shang, T. *et al.* DynamiCare: a dynamic multi-agent framework for interactive and open-ended medical decision-making. Preprint at https://doi.org/10.48550/arXiv.2507.02616 (2025).

27. Zuo, K., Jiang, Y., Mo, F. & Lio, P. KG4Diagnosis: a hierarchical multi-agent LLM framework with knowledge graph enhancement for medical diagnosis. Preprint at https://doi.org/10.48550/arXiv.2412.16833 (2025).

28. Kanithi, P. *et al.* MEDIC: comprehensive evaluation of leading indicators for LLM safety and utility in clinical applications. Preprint at https://doi.org/10.48550/arXiv.2409.07314 (2026).

29. Englander, R. *et al.* Toward a common taxonomy of competency domains for the health professions and competencies for physicians. *Acad. Med.* **88**, 1088 (2013).

30. Schmidt, H. G. & Rikers, R. M. J. P. How expertise develops in medicine: knowledge encapsulation and illness script formation. *Med. Educ.* **41**, 1133–1139 (2007).

31. Salaudeen, O. *et al.* Measurement to meaning: a validity-centered framework for AI evaluation. Preprint at https://doi.org/10.48550/arXiv.2505.10573 (2025).

32. Jo, N. & Wilson, A. What does your benchmark really measure? A framework for robust inference of AI capabilities. Preprint at https://doi.org/10.48550/arXiv.2509.19590 (2025).

33. Raji, D., Denton, E., Bender, E. M., Hanna, A. & Paullada, A. AI and the everything in the whole wide world benchmark. in *Proceedings of the Neural Information Processing Systems Track on Datasets and Benchmarks* (eds Vanschoren, J. & Yeung, S.) vol. 1 (2021).

34. Harding, J. & Sharadin, N. What is it for a machine learning model to have a capability? *Br. J. Philos. Sci.* https://doi.org/10.1086/732153 doi:10.1086/732153.

35. Rajpurkar, P. & Topol, E. J. A clinical certification pathway for generalist medical AI systems. *The Lancet* **405**, 20 (2025).

36. Roskies, A. L. The binding problem. *Neuron* **24**, 7–9 (1999).

37. Newell, A. & Simon, H. The logic theory machine–a complex information processing system. *IRE Trans. Inf. Theory* **2**, 61–79 (1956).

38. Fox, S. Human–artificial intelligence systems: how human survival first principles influence machine learning world models. *Systems* **10**, (2022).

39. Craik, K. J. W. *The Nature of Explanation*. vol. 445 (CUP Archive, 1967).

40. Qazi, M. A., Nadeem, M. & Yaqub, M. Beyond generative AI: world models for clinical prediction, counterfactuals, and planning. Preprint at https://doi.org/10.48550/arXiv.2511.16333 (2025).

41. Friston, K. The free-energy principle: a unified brain theory? *Nat. Rev. Neurosci.* **11**, 127–138 (2010).

42. Sakagami, R. *et al.* Robotic world models—conceptualization, review, and engineering best practices. *Front. Robot. AI* **10**, (2023).

43. Coulter, A. & Collins, A. *Making Shared Decision-Making a Reality: No Decision about Me, without Me.* https://www.kingsfund.org.uk/publications/making-shared-decision-making-reality (2011).

44. Ilgen, D. R., Hollenbeck, J. R., Johnson, M. & Jundt, D. Teams in Organizations: From Input-Process-Output Models to IMOI Models. *Annu. Rev. Psychol.* **56**, 517–543 (2005).

45. Elwyn, G. *et al.* Shared decision making: a model for clinical practice. *J. Gen. Intern. Med.* **27**, 1361–1367 (2012).

46. Engel, G. L. The need for a new medical model: a challenge for biomedicine. *Science* **196**, 129–136 (1977).

47. Zacher, H. Action regulation theory. in *Oxford research encyclopedia of psychology*.

48. Dahlberg, K., Todres, L. & Galvin, K. Lifeworld-led healthcare is more than patient-led care: an existential view of well-being. *Med. Health Care Philos.* **12**, 265–271 (2009).

49. Evans, J. S. B. T. & Stanovich, K. E. Dual-process theories of higher cognition: advancing the debate. *Perspect. Psychol. Sci. J. Assoc. Psychol. Sci.* **8**, 223–241 (2013).

50. Pelaccia, T., Tardif, J., Triby, E. & Charlin, B. An analysis of clinical reasoning through a recent and comprehensive approach: the dual-process theory. *Med. Educ. Online* **16**, (2011).

51. Thinking, fast and slow. *Macmillan Publishers* https://us.macmillan.com/books/9780374533557/thinkingfastandslow/.

52. Croskerry, P. A universal model of diagnostic reasoning. *Acad. Med.* **84**, 1022 (2009).

53. Elstein, A. S., Shulman, L. S. & Sprafka, S. A. Medical problem solving: a ten-year retrospective. *Eval. Health Prof.* **13**, 5–36 (1990).

54. Schmidt, H. & {de Volder}, M. *Tutorials in Problem-Based Learning*. (Koninklijke Van Gorcum BV, Netherlands, 1984).

55. McGrath, J. E. *Social Psychology : A Brief Introduction*. (Holt, Rinehart and Winston, 1970).

56. Lamme, V. A. F. & Roelfsema, P. R. The distinct modes of vision offered by feedforward and recurrent processing. *Trends Neurosci.* **23**, 571–579 (2000).

57. Mashour, G. A., Roelfsema, P., Changeux, J.-P. & Dehaene, S. Conscious processing and the global neuronal workspace hypothesis. *Neuron* **105**, 776–798 (2020).

58. Dehaene, S., Kerszberg, M. & Changeux, J. P. A neuronal model of a global workspace in effortful cognitive tasks. *Proc. Natl. Acad. Sci. U. S. A.* **95**, 14529–14534 (1998).

59. Posner, M. I. Orienting of attention*. *Q. J. Exp. Psychol.* **32**, 3–25 (1980).

60. Ackoff, R. L. From data to wisdom. *J. Appl. Syst. Anal.* **16**, 3–9 (1989).

61. Flavell, J. H. Metacognition and cognitive monitoring: a new area of cognitive–developmental inquiry. *Am. Psychol.* **34**, 906–911 (1979).

62. Endsley, M. R. Toward a theory of situation awareness in dynamic systems. *Hum. Factors* **37**, 32–64 (1995).

63. Patton, L. Helmholtz on unconscious inference in experience. in *The routledge handbook of philosophy and implicit cognition* (Routledge, 2022).

64. Miller, G. A., Galanter, E. & Pribram, K. H. *Plans and the Structure of Behavior*. (Henry Holt and Co., New York, 1960). doi:10.1037/10039-000.

65. Croskerry, P., Singhal, G. & Mamede, S. Cognitive debiasing 1: origins of bias and theory of debiasing. *BMJ Qual. Saf.* **22**, ii58–ii64 (2013).

66. The process of experiential learning. in *Strategic learning in a knowledge economy* 313–331 (Butterworth-Heinemann, 2000). doi:10.1016/B978-0-7506-7223-8.50017-4.

67. Lerner, J. S., Li, Y., Valdesolo, P. & Kassam, K. S. Emotion and decision making. *Annu. Rev. Psychol.* **66**, 799–823 (2015).

68. Schutz, A. & Luckmann, T. *The Structures of the Life-World*. vol. 1 (northwestern university press, 1973).

69. Merleau-Ponty, M., Landes, D., Carman, T. & Lefort, C. *Phenomenology of Perception*. (Routledge, 2013).

70. Henry, S. G. Recognizing tacit knowledge in medical epistemology. *Theor. Med. Bioeth.* **27**, 187–213 (2006).

71. Fan, R. Self-determination vs. family-determination: two incommensurable principles of autonomy: a report from east Asia. *Bioethics* **11**, 309–322 (1997).

72. Berkman, N. D., Sheridan, S. L., Donahue, K. E., Halpern, D. J. & Crotty, K. Low health literacy and health outcomes: an updated systematic review. *Ann. Intern. Med.* **155**, 97–107 (2011).

73. Alden, D. L. *et al.* Who decides: me or we? Family involvement in medical decision making in eastern and western countries. *Med. Decis. Mak. Int. J. Soc. Med. Decis. Mak.* **38**, 14–25 (2018).

74. Liu, F. *et al.* A foundational architecture for AI agents in healthcare. *Cell Rep. Med.* **6**, 102374 (2025).

75. Greengrass, C. J. Transforming clinical reasoning-the role of AI in supporting human cognitive limitations. *Front. Digit. Health* **7**, 1715440 (2025).

76. Dawid, A. & LeCun, Y. Introduction to latent variable energy-based models: a path toward autonomous machine intelligence. *J. Stat. Mech. Theory Exp.* **2024**, 104011 (2024).

77. Ghaferi, A. A., Birkmeyer, J. D. & Dimick, J. B. Variation in hospital mortality associated with inpatient surgery. *N. Engl. J. Med.* **361**, 1368–1375 (2009).

78. Curry, L. A. *et al.* What distinguishes top-performing hospitals in acute myocardial infarction mortality rates? A qualitative study. *Ann. Intern. Med.* **154**, 384–390 (2011).

79. Harrison, J. E., Weber, S., Jakob, R. & Chute, C. G. ICD-11: an international classification of diseases for the twenty-first century. *BMC Med. Inform. Decis. Mak.* **21**, 206 (2021).

80. Vos, T. *et al.* Global burden of 369 diseases and injuries in 204 countries and territories, 1990–2019: a systematic analysis for the global burden of disease study 2019. *The Lancet* **396**, 1204–1222 (2020).

81. World Health Organization. *Framework on Integrated People-Centred Health Services.* https://apps.who.int/gb/ebwha/pdf_files/wha69/a69_39-en.pdf (2016).

82. National Library of Medicine (NLM). Overview of SNOMED CT. *US National Institute of Health (NIH)* https://www.nlm.nih.gov/healthit/snomedct/snomed_overview.html (2016).

83. Health Workforce (HWF). Classifying health workers. https://www.who.int/publications/m/item/classifying-health-workers.

84. Carro, M. V. *et al.* A conceptual framework for AI capability evaluations. Preprint at https://doi.org/10.48550/arXiv.2506.18213 (2025).

85. Blömeke, S., Gustafsson, J.-E. & Shavelson, R. J. Beyond dichotomies. *Z. Für Psychol.* **223**, 3–13 (2015).

86. Firestone, C. Performance vs. competence in human–machine comparisons. *Proc. Natl. Acad. Sci. U. S. A.* **117**, 26562–26571 (2020).

87. Huang, Y. Levels of AI agents: from rules to large language models. Preprint at https://doi.org/10.48550/arXiv.2405.06643 (2024).

88. Li, Y. *et al.* Personal LLM agents: insights and survey about the capability, efficiency and security. Preprint at https://doi.org/10.48550/arXiv.2401.05459 (2024).

89. Morris, M. R. *et al.* Levels of AGI for operationalizing progress on the path to AGI. Preprint at https://doi.org/10.48550/arXiv.2311.02462 (2025).

90. Jacob, C. *et al.* AI for IMPACTS framework for evaluating the long-term real-world impacts of AI-powered clinician tools: systematic review and narrative synthesis. *J. Med. Internet Res.* **27**, e67485 (2025).

91. Natali, C., Marconi, L., Dias Duran, L. D. & Cabitza, F. AI-induced deskilling in medicine: a mixed-method review and research agenda for healthcare and beyond. *Artif. Intell. Rev.* **58**, 356 (2025).

92. You, J. G., Hernandez-Boussard, T., Pfeffer, M. A., Landman, A. & Mishuris, R. G. Clinical trials informed framework for real world clinical implementation and deployment of artificial intelligence applications. *Npj Digit. Med.* **8**, 107 (2025).

93. Lorenzini, G., Arbelaez Ossa, L., Shaw, D. M. & Elger, B. S. Artificial intelligence and the doctor–patient relationship expanding the paradigm of shared decision making. *Bioethics* **37**, 424–429 (2023).

94. Mitchell, M. *et al.* Model cards for model reporting. in *Proceedings of the Conference on Fairness, Accountability, and Transparency* 220–229 (Association for Computing Machinery, New York, NY, USA, 2019). doi:10.1145/3287560.3287596.

95. Chen, Y.-J., Albarqawi, A. & Chen, C.-S. Enhancing clinical decision-making: integrating multi-agent systems with ethical AI governance. in *2025 IEEE Conference on Computational Intelligence in Bioinformatics and Computational Biology (CIBCB)* 1–7 (2025). doi:10.1109/CIBCB66090.2025.11177136.

96. Miao, Y., Wen, J., Luo, Y. & Li, J. MedARC: adaptive multi-agent refinement and collaboration for enhanced medical reasoning in large language models. *Int. J. Med. Inf.* **206**, 106136 (2026).

97. Klang, E. *et al.* Orchestrated multi agents sustain accuracy under clinical-scale workloads compared to a single agent. *Npj Health Syst.* **3**, 23 (2026).

98. Xu, H. *et al.* TAMA: a human-AI collaborative thematic analysis framework using multi-agent LLMs for clinical interviews. Preprint at https://doi.org/10.48550/arXiv.2503.20666 (2025).

99. Kim, Y. *et al.* Tiered agentic oversight: a hierarchical multi-agent system for healthcare safety. Preprint at https://doi.org/10.48550/arXiv.2506.12482 (2025).

100. Smith, H. Clinical AI: opacity, accountability, responsibility and liability. *AI Soc.* **36**, 535–545 (2021).

101. Bleher, H. & Braun, M. Diffused responsibility: attributions of responsibility in the use of AI-driven clinical decision support systems. *Ai Ethics* **2**, 747–761 (2022).

102. U.S. Food And Drug Administration. *Artificial Intelligence-Enabled Device Software Functions: Lifecycle Management and Marketing Submission Recommendations.* https://www.fda.gov/regulatory-information/search-fda-guidance-documents/artificial-intelligence-enabled-device-software-functions-lifecycle-management-and-marketing (2025).

103. Hinostroza Fuentes, V. G., Karim, H. A., Tan, M. J. T. & AlDahoul, N. AI with agency: a vision for adaptive, efficient, and ethical healthcare. *Front. Digit. Health* **7**, 1600216 (2025).

104. Pearl, J. *Causality*. (Cambridge University Press, Cambridge, 2009). doi:10.1017/CBO9780511803161.

105. Sheiner, L. B. & Steimer, J.-L. Pharmacokinetic/pharmacodynamic modeling in drug development. *Annu. Rev. Pharmacol. Toxicol.* **40**, 67–95 (2000).

106. Corral-Acero, J. *et al.* The 'digital twin' to enable the vision of precision cardiology. *Eur. Heart J.* **41**, 4556–4564 (2020).

107. Chandak, P., Huang, K. & Zitnik, M. Building a knowledge graph to enable precision medicine. *Sci. Data* **10**, 67 (2023).

108. Murphy, S. N. *et al.* Serving the enterprise and beyond with informatics for integrating biology and the bedside (i2b2). *J. Am. Med. Inform. Assoc. JAMIA* **17**, 124–130 (2010).

109. Overhage, J. M., Ryan, P. B., Reich, C. G., Hartzema, A. G. & Stang, P. E. Validation of a common data model for active safety surveillance research. *J. Am. Med. Inform. Assoc. JAMIA* **19**, 54–60 (2012).

110. Mandel, J. C., Kreda, D. A., Mandl, K. D., Kohane, I. S. & Ramoni, R. B. SMART on FHIR: a standards-based, interoperable apps platform for electronic health records. *J. Am. Med. Inform. Assoc. JAMIA* **23**, 899–908 (2016).

111. Lindberg, D. A. B., Humphreys, B. L. & McCray, A. T. The unified medical language system. *Methods Inf. Med.* **32**, 281–291 (1993).

112. Haig, K. M., Sutton, S. & Whittington, J. SBAR: a shared mental model for improving communication between clinicians. *Jt. Comm. J. Qual. Patient Saf.* **32**, 167–175 (2006).

113. Andrews, R. W., Lilly, J. M., Srivastava, D. & Feigh, K. M. The role of shared mental models in human-AI teams: a theoretical review. *Theor. Issues Ergon. Sci.* **24**, 129–175 (2023).

114. Premack, D. & Woodruff, G. Does the chimpanzee have a theory of mind? *Behav. Brain Sci.* **1**, 515–526 (1978).

115. Schmutz, J. B., Outland, N., Kerstan, S., Georganta, E. & Ulfert, A.-S. AI-teaming: redefining collaboration in the digital era. *Curr. Opin. Psychol.* **58**, 101837 (2024).

116. Rosen, M. A. *et al.* Teamwork in healthcare: key discoveries enabling safer, high-quality care. *Am. Psychol.* **73**, 433 (2018).

117. Ericsson, K. A. Deliberate practice and the acquisition and maintenance of expert performance in medicine and related domains. *Acad. Med.* **79**, S70 (2004).

118. Guyatt, G. H. *et al.* GRADE: an emerging consensus on rating quality of evidence and strength of recommendations. https://doi.org/10.1136/bmj.39489.470347.AD (2008) doi:10.1136/bmj.39489.470347.AD.

119. Wu, E. *et al.* How medical AI devices are evaluated: limitations and recommendations from an analysis of FDA approvals. *Nat. Med.* **27**, 582–584 (2021).

120. Lekadir, K. *et al.* FUTURE-AI: international consensus guideline for trustworthy and deployable artificial intelligence in healthcare. https://doi.org/10.1136/bmj-2024-081554 (2025) doi:10.1136/bmj-2024-081554.

121. Krathwohl, D. R. A revision of bloom's taxonomy: an overview: theory into practice: vol 41, No 4. *Theory Pract.* **41**, 212–218 (2002).

122. Miller, G. E. The assessment of clinical skills/competence/performance. *Acad. Med. J. Assoc. Am. Med. Coll.* **65**, S63-67 (1990).

123. Skivington, K. *et al.* A new framework for developing and evaluating complex interventions: update of medical research council guidance. *BMJ* **374**, n2061 (2021).

124. McCulloch, P. *et al.* No surgical innovation without evaluation: the IDEAL recommendations. *Lancet* **374**, 1105–1112 (2009).

125. van der Vegt, A. H. *et al.* Implementation frameworks for end-to-end clinical AI: derivation of the SALIENT framework. *J. Am. Med. Inform. Assoc. JAMIA* **30**, 1503–1515 (2023).

126. Adnan, H. S., Shidani, A., Clifton, L., Bankhead, C. R. & Perera-Salazar, R. Implementation framework for AI deployment at scale in healthcare systems. *iScience* **28**, 112406 (2025).

127. Beauchamp, T. L., Childress, J. F., Beauchamp, T. L. & Childress, J. F. *Principles of Biomedical Ethics*. (Oxford University Press, Oxford, New York, 2019).

128. Gruppen, L. D. & Frohna, A. Z. Clinical reasoning. in *International handbook of research in medical education* (eds Norman, G. R. et al.) 205–230 (Springer Netherlands, Dordrecht, 2002). doi:10.1007/978-94-010-0462-6_8.

129. Patel, V. L., Kaufman, D. R. & Kannampallil, T. G. Diagnostic reasoning and decision making in the context of health information technology. *Rev. Hum. Factors Ergon.* **8**, 149–190 (2013).

130. Cader, R., Campbell, S. & Watson, D. Cognitive continuum theory in nursing decision-making. *J. Adv. Nurs.* **49**, 397–405 (2005).

131. Graham, E. A working theory of a learned model in a partially observable environment for cognitive decision-making. in *2022 Systems and Information Engineering Design Symposium (SIEDS)* 381–386 (2022). doi:10.1109/SIEDS55548.2022.9799386.

# Supplementary Information

## Grounding Clinical AI Competency in Human Cognition Through the Clinical World Model and Skill-Mix Framework

Seyed Amir Ahmad Safavi-Naini[1,2,3*], Elahe Meftah[4†], Josh Mohess[3], Pooya Mohammadi Kazaj[1,2,5], Georgios Siontis[1], Zahra Atf[6], Peter R. Lewis[6], Mauricio Reyes[2,7,8], Girish Nadkarni[3,8], Roland Wiest[2,9,10], Stephan Windecker[1], Christoph Gräni[1], Ali Soroush[3,11], Isaac Shiri[1,2]

1- Department of Cardiology, Inselspital, Bern University Hospital, University of Bern, Bern, Switzerland

2- Department of Digital Medicine, Bern University Hospital, University of Bern, Bern, Switzerland

3- Division of Data-Driven and Digital Medicine (D3M), Icahn School of Medicine at Mount Sinai, New York, United States

4- Clinical Research Development Center, Amir Oncology Teaching Hospital, Shiraz University of Medical Sciences, Shiraz, Iran

5- Graduate School for Cellular and Biomedical Sciences, University of Bern, Bern, Switzerland

6- Faculty of Business and Information Technology, Ontario Tech University, Oshawa, Canada

7- Department of Radiation Oncology, Inselspital, Bern University Hospital and University of Bern, Bern, Switzerland

8- ARTORG Center for Biomedical Engineering Research, University of Bern, Bern, Switzerland

8- The Charles Bronfman Institute of Personalised Medicine, Icahn School of Medicine at Mount Sinai, New York, NY, USA

9- University Institute of Diagnostic and Interventional Neuroradiology, Inselspital, Bern University Hospital, University of Bern, Bern, Switzerland

10- Translational Imaging Center (TIC), Swiss Institute for Translational and Entrepreneurial Medicine, Bern, Switzerland

11- Henry D. Janowitz Division of Gastroenterology, Icahn School of Medicine at Mount Sinai, New York, NY, USA

# List of Supplementary Material

**Supplementary Figures**

**Supplementary Figure 1.** Function-by-Substrate taxonomy of ecosystem components in the Clinical World Model.

**Supplementary Figure 2.** Architecture of the Artificial Agent Decision Making (ADM) Model.

**Supplementary Tables**

**Supplementary Table 1.** Complementary Modeling Frameworks constituting the Clinical World.

**Supplementary Table 2.** Theoretical Underpinnings of Clinical Decision Making (CDM) and Patient Decision Making (PDM) in the Clinical World Model.

**Supplementary Table 3.** Classification schemes for intelligent systems based on development method, task complexity, and autonomy level.

**Supplementary Notes**

**Supplementary Note 1.** Review of Existing Frameworks for Intelligent System Design or Evaluation.

**Supplementary Note 2.** Dissection of World Models and its Dimension Taxonomies.

**Supplementary Note 3.** Detail of Cognitive Architectures for Decision Making Within the Clinical World.

**Supplementary Data**

**Supplementary Data 1.** Structured JSON specifications for all eight Skill-Mix dimensions, including 1,918 Conditions, 7 Care Phases, 12 Care Settings, 58 Clinical Tasks, 86 Care Provider Roles, 4 Agent Facing patterns, 7 Anchoring Layer patterns, and 3 Assigned Authority patterns. Also available at: https://github.com/Sdamirsa/Clinical-World-Model/tree/main/clinical-skill-mix

# Supplementary Figures

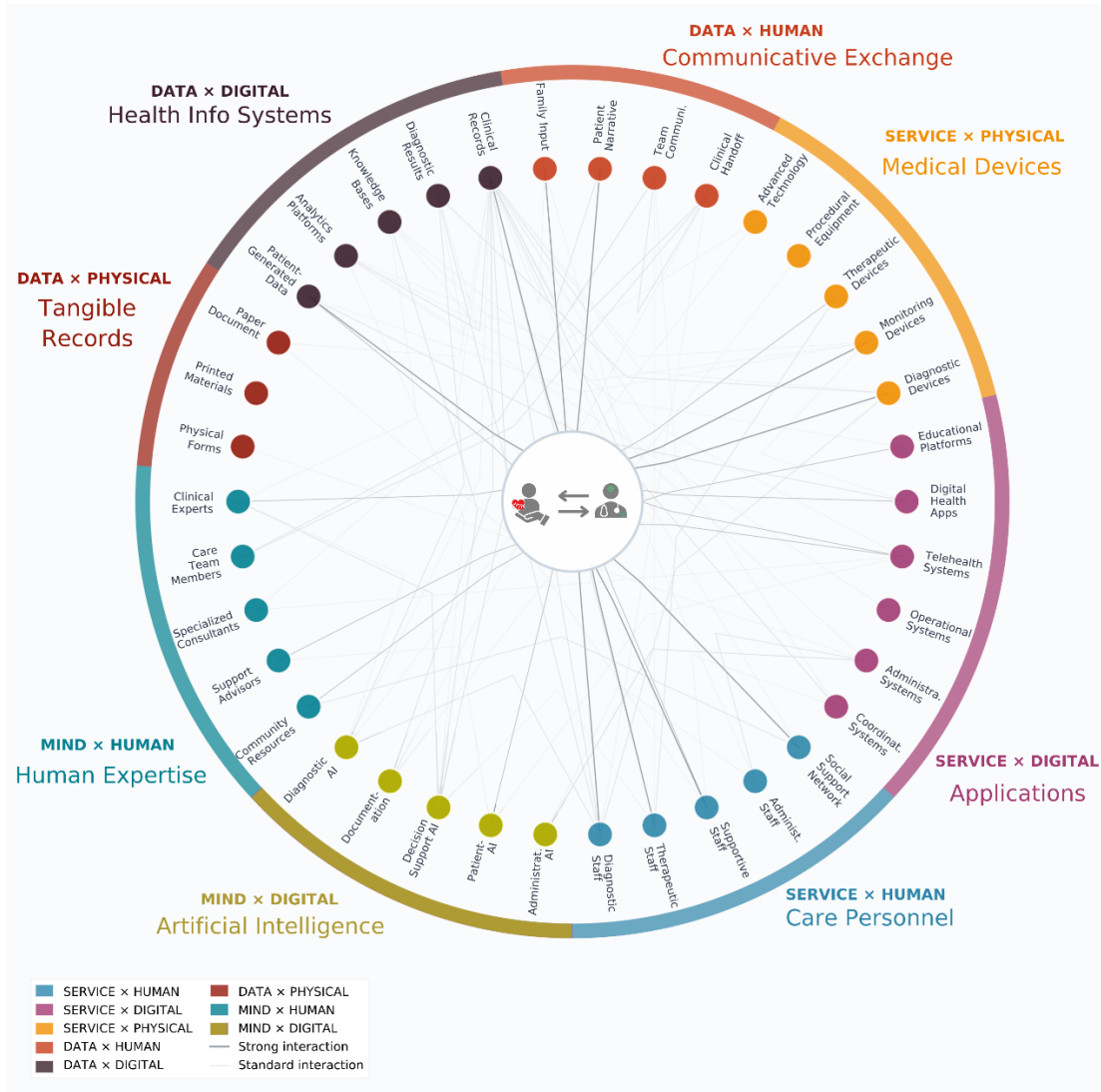

**Supplementary Figure 1. Function-by-Substrate taxonomy of ecosystem components in the Clinical World Model.**

Circular diagram displaying ecosystem components organized by substrate (Human, Digital, Physical) and function (Data, Mind, Service). Crossing three functions with three substrates yields nine theoretical categories, eight of which are instantiated here; the Mind × Physical cell has no clear real-world counterpart and is intentionally left empty. Data in the Human substrate takes the form of communicative exchanges; in the Digital substrate, data appears in health information systems such as electronic health records, laboratory databases, and imaging archives; and in the Physical substrate, data exists as tangible records such as paper documents. Mind in the Human substrate takes the form of human expertise, exemplified by the clinical judgment of consultants, specialists, and experienced colleagues, and in the Digital substrate as artificial intelligence, the algorithmic reasoning systems central to this framework. Service in the Human substrate is delivered by care personnel executing clinical tasks; in the Digital substrate by applications that perform clinical functions without direct human operation; and in the Physical substrate by medical devices that deliver diagnostic or therapeutic interventions. Individual components within each category appear as colored dots along the outer ring. Clinical AI occupies the Mind × Digital cell, and this position within the broader ecosystem constrains how AI engages with human players who span other cells of the matrix.

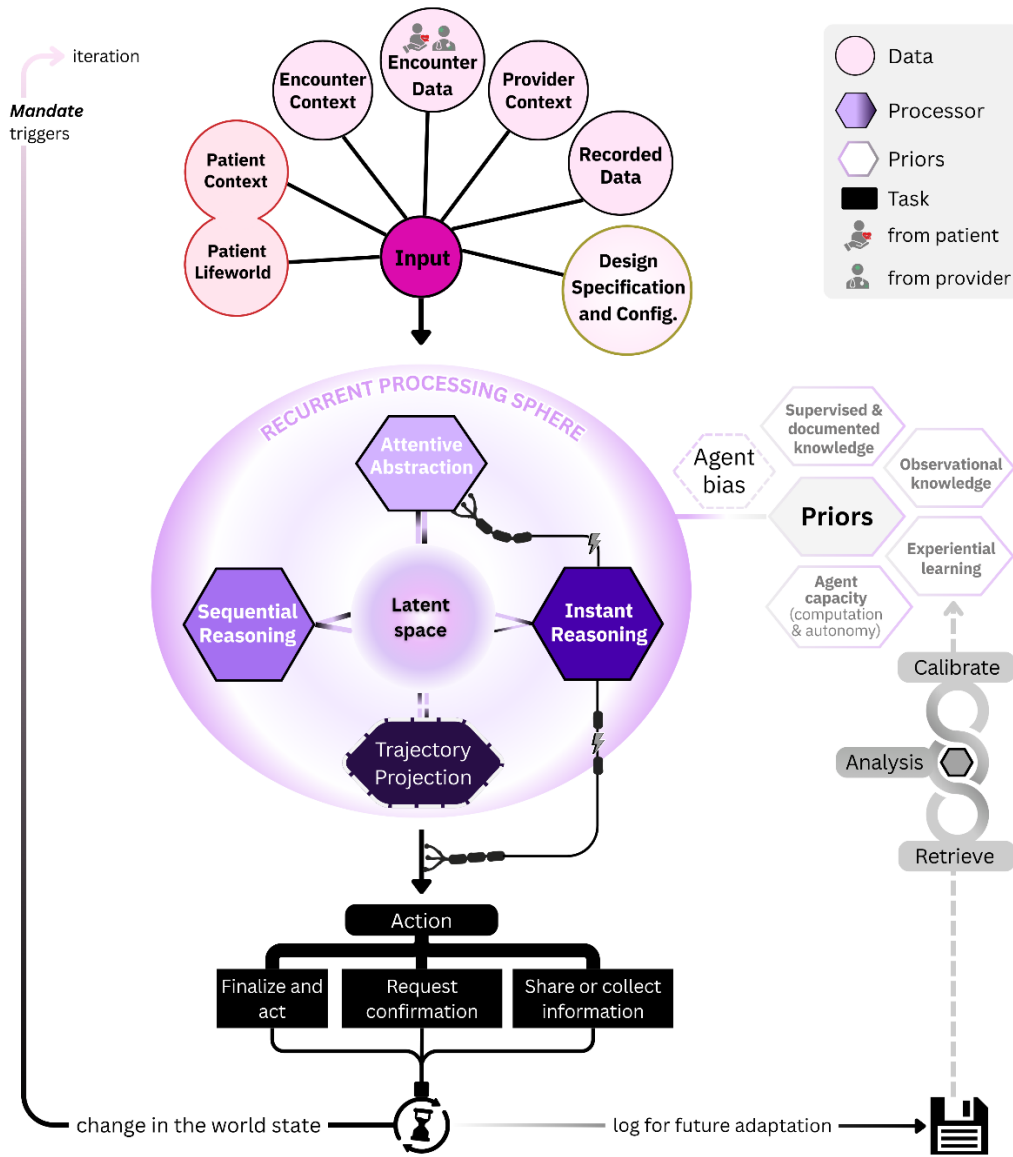

**Supplementary Figure 2. The architecture of Artificial Agent Decision Making (ADM) Model.**

Architecture of an ADM cognitive cycle, structured as a parallel to the CDM and PDM models with AI-native components. A mandate triggers each iteration. Input integrates the same clinical scene accessible to human agents (Encounter Data, Encounter Context, Patient Context, Patient Lifeworld, Provider Context, and Recorded Data) alongside a unique stream, Design Specification and Configuration, which encodes the agent's engineered constraints. Input flows into the Recurrent Processing Sphere, where human cognitive components are systematically mapped to AI-native counterparts: Attentive Abstraction replaces the Data Processor, Latent Space replaces Hypothesis, and dual reasoning pathways are recast as Instant Reasoning and Sequential Reasoning. Trajectory Projection replaces Reflection, providing forward-looking evaluation of action consequences. Priors comprise supervised and documented knowledge, observational knowledge, experiential learning, and agent capacity (computation and autonomy), with Agent Bias acknowledged as an influence on prior integration. Processing resolves into Action, which includes a third type absent from both human models: Request Confirmation, alongside Finalize and Act and Share or Collect

Information. This third action type encodes human-in-the-loop oversight at the architectural level. Each action produces a change in the world state and is logged for future adaptation.

# Supplementary Tables

**Supplementary Table 1. Complementary Modeling Frameworks constituting the Clinical World.** Ten framework modelling traditions organized by type: world model (W) represents the state of some aspect of clinical reality; process model (P) describes how agents transform that state through cognition and action; interaction model (I) specifies how multiple agents jointly act upon shared reality across boundaries; specification framework (F) provides normative or structural criteria against which the other models are designed, evaluated, or governed. "Absorbed Traditions" lists established frameworks subsumed within each model. Dimensional coverage is classified as Defines (primary scope), Engages (necessarily uses), or Touches (peripheral contact).

| Model | Type | Central Question | Absorbed Traditions | Dimensions - Defines | Dimensions - Engages | Dimensions: Touches |
|---|---|---|---|---|---|---|
| Reference World Model | Describe - World | What shapes the clinical world? | SEIPS work system models; ecological/environmental models[16,19,20] | Actors, Context, Temporality, Axiom (ontological), Codex | — | Information, Mandate, Authority |
| Causal World Model | Generate - World | What causes what, and what would happen if? | Structural causal models; Bayesian clinical decision support; counterfactual reasoning; pathophysiological models; PK/PD models; digital twins; biological knowledge bases; epistemic uncertainty models[104–106]. | Axiom (mechanistic) | Information, Codex, Temporality, Outcomes | Representation, Cognition |
| Data and Knowledge Representation Model | Normative - Specification | How should clinical reality be formally encoded for computational processing? | Data exchange standards; research common data models; clinical knowledge graphs; cross-terminology mapping; medical ontology engineering; EHR data models[107–111] | Information (formal encoding), Representation (computable structures), Codex (standardized clinical vocabulary) | Context, Actors | Axiom, Temporality |
| Decision-Making Model | Describe - Process | How do agents transform information into action? | Dual-process theory; bounded rationality; naturalistic decision-making; cognitive load theory; clinical reasoning models; confidence calibration; forward simulation[52,52,112] | Cognition | Mandate, Information, Representation, Temporality | Actors, Codex, Outcomes |
| Mental Model | Describe - World | How do agents represent one another internally? | Theory of mind; shared mental models; trust models[112–114] | Representation (recursive inter-agent) | Cognition, Actors, Temporality, Adaptation, Information | Axiom, Mandate, Normativity, Outcomes |

| Model | Type | Central Question | Absorbed Traditions | Dimensions - Defines | Dimensions - Engages | Dimensions: Touches |
|---|---|---|---|---|---|---|
| Collaborative World Model | Describe - Interaction | How do agents coordinate across boundaries? | Governance and institutional models; multi-agent orchestration; team training frameworks; structured [115,116] | Actors and Action (inter-agent governance), Authority | Mandate, Temporality | Context, Codex, Information, Normativity |
| (Machine) Learning World Model | Generate - World | How are representations acquired and updated from experience? | Experiential learning; deliberate practice; model-based RL [76,117] | Adaptation, Information (acquisition from experience), Representation (learned encodings) | Axiom, Temporality, Cognition, Outcomes | Codex |
| Normative World Model | Normative - Specification | What ought to be done, and what is permissible? | Bioethics principles; clinical ethics; value theory; research ethics declarations; regulatory frameworks [90,118–120] | Normativity, Outcomes | Authority, Mandate | Adaptation |
| Skill-Mix | Normative - Specification | How is competency organized into bounded, evaluable units? | Competency-based medical education; physician competency standards; evaluation frameworks [35,121,122]. | Mandate | Actors, Context, Codex, Authority, Outcomes | Temporality |
| Deployment Model | Describe - Interaction | Under what conditions does a capability hold in practice, and how is that validity sustained? | model cards; MLOps and monitoring; FDA phased deployment guidance; AI readiness frameworks [92,123–126] | Outcomes (validation evidence), Adaptation (lifecycle governance) | Mandate, Authority, Context, Codex | Actors, Temporality, Normativity |

**Supplementary Table 2. Theoretical Underpinnings of Clinical Decision Making (CDM) and Patient Decision Making (PDM) in the Clinical World Model.** The theories and frameworks include: Dual Process Theory[49–51], Universal Model of Diagnostic Reasoning[52], Hypothetico-Deductive Model[53], Illness Script Theory[54], Input-Process-Output (IPO) Model[55], Input-Mediator-Output-Input (IMOI) Model[44], Recurrent Processing Theory[56], Global Workspace Theory[57,58], Global Neuronal Workspace[57,58], Attentional Orienting[59], Data-Information-Knowledge-Wisdom (DIKW) Hierarchy[60], A Model of Cognitive Monitoring[61], Situational Awareness[62], Unconscious Inference[63], Free Energy Principle[41], Test-Operate-Test-Exit (TOTE) Model[64], Action Regulation Theory[47], Cognitive Debiasing[65], Experiential Learning Cycle[66], Emotion-Imbued Choice Model[67], Biopsychosocial Model[46], Shared Decision-Making Framework[15,43,45], Systems Engineering Initiative for Patient Safety (SEIPS) Models[19,20,16], Structures of the Life-World[68], Phenomenology of Perception[69], Lifeworld-Led Healthcare[48].

| First Author, Theory, Year | Theory Aim & Core Elements | Framework Components & Implications |
|---|---|---|
| Evans, Pelaccia, Kahneman, **Dual Process Theory**, 2008[49–51] | Distinguishes System 1 (fast, automatic, intuitive) from System 2 (slow, deliberate, analytical). Systems operate in parallel with asymmetric resource demands and error profiles. | **System I (Clinical Intuition) & System II (Analytical Processing)**: Adopted dual-system architecture. Oscillation between systems within Recurrent Workspace enables dynamic calibration of intuitive pattern recognition versus deliberate analysis during hypothesis evaluation. |
| Croskerry, **Universal Model of Diagnostic Reasoning**, 2009[52] | Operationalizes dual-process theory for diagnosis. Introduces bidirectional switching between Type 1 and Type 2 processing. Identifies calibration mechanisms and cognitive/affective override pathways. (Toggle function terminology formalized in Croskerry 2013.) | **System I–II Bidirectional Oscillation**: Toggle function implemented as dynamic switching during hypothesis evaluation. Override capacity preserved—either system can reject outputs when incongruence detected, triggering re-processing. |
| Elstein et al., **Hypothetico-Deductive Model**, 1978[53] | Clinical reasoning proceeds cyclically: cue acquisition → hypothesis generation → cue interpretation → hypothesis evaluation. Early hypotheses constrain subsequent inquiry. | **Internal processing sequence**: Directly informed: Data→Cues→Hypothesis→Evaluation→Plan→Action. Hypothesis-driven attention (evaluating hypothesis triggers return to Data Processor for targeted cue extraction) reflects Elstein's iterative refinement. |
| Schmidt & Boshuizen, **Illness Script Theory**, 1993[54] | Expertise develops via encapsulation: detailed biomedical knowledge compressed into illness scripts (enabling conditions, fault, consequences). Experts pattern-match; novices reason causally. | **Priors and System I efficiency**: Accumulated experience shapes Prior states and processor parameters. Mature illness scripts enable System I to bypass extended deliberation, shortcutting directly to Plan when pattern confidence exceeds threshold. |
| McGrath, Input-Process-Output **(IPO)** Model, 1964[55] | Systems framework from Social Psychology: A Brief Introduction. Inputs transformed through processes into outputs. Establishes linear temporal causality as foundational model for performance analysis. | **External Iteration of Input-Action-Input**: Adopted for outer iteration loop. Action (output) generates environmental change producing new Input, initiating subsequent cycle. Temporality variable: seconds (communicative exchange) to years (longitudinal follow-up). |
| Ilgen et al., Input-Mediator-Output-Input **(IMOI)** Model, 2005[44] | Input-Mediator-Output-Input model extends IPO: outputs become inputs (cyclical causality); "M" replaces "P" to reflect broader mediators; terminal "I" invokes cyclical causal feedback. | **Dual-loop architecture, External and Recurrent Processing**: IMOI justified both IPA macro-cycle and internal Recurrent Workspace micro-iterations. Output→Input feedback and variable cycle temporality directly derived. "Mediator" concept reflected in System I–II as transformative processors. |
| Lamme, **Recurrent Processing Theory**, 2000[56] | Conscious perception requires recurrent (feedback) processing between hierarchical cortical areas. Feedforward sweep enables rapid categorization; recurrent loops enable integration and awareness. | **Recurrent Workspace (nomenclature and architecture)**: Central design principle. Feedforward path (Input→Cues→Hypothesis) parallels initial sweep; feedback paths (Reflection→Hypothesis, System I–II→Cues) enable iterative refinement characteristic of conscious deliberation. |

| First Author, Theory, Year | Theory Aim & Core Elements | Framework Components & Implications |
|---|---|---|
| Baars, **Global Workspace Theory**, 1988[57] | Consciousness emerges via global broadcast: specialized unconscious processors compete for access to limited-capacity workspace; winning content becomes globally available. | **Recurrent Workspace as integrative hub**: Workspace concept adopted for bounded processing sphere where distributed inputs converge. Framework departs from GWT by emphasizing information states over functional modules—enabling decomposition into clinically specifiable competency nodes. |
| Dehaene & Changeux, **Global Neuronal Workspace**, 2011[57,58] | Neural implementation of Global Workspace. "Ignition" = nonlinear amplification enabling global broadcast. Prefrontal-parietal pyramidal neurons form workspace infrastructure. | **Biological grounding**: Ignition parallels threshold-crossing when hypothesis confidence triggers action commitment. Long-term memory→Input source; Evaluative systems→System I–II; Perceptual systems→Prior/Input states; Motor systems→Action pathways. |
| Posner, **Attentional Orienting**, 1990[59] | Attention as selective processing resource: alerting, orienting, executive control networks. Enables prioritization amid competing stimuli. | **Data Processor attention function**: Theoretical basis for attention-weighted transformation of raw Input into Cues. Data Processor implements selective amplification of clinically relevant signals, operationalizing Posner's orienting function. |
| Ackoff, **DIKW Hierarchy**, 1989[60] | Data→Information→Knowledge→Wisdom progression. Each transformation adds contextual meaning, culminating in judgment capacity. | **Transformation sequence within Recurrent Workspace**: Raw Data (unprocessed signals) → Cues (clinically relevant information) → Hypothesis (structured knowledge) → Plans (wisdom-guided action selection). Framework extends DIKW by adding recursion. |
| Flavell, **A Model of Cognition Monitoring**, 1979[61] | "Cognition about cognition": four components—metacognitive knowledge, metacognitive experiences, goals/tasks, and actions/strategies. Enables self-regulation and adaptive strategy selection. | **Reflection & Feedback as separable metacognitive processors**: Reflection = synchronous monitoring (certainty/risk weighting during active reasoning). Feedback = asynchronous evaluation (post-action comparison of projected vs. actual outcomes). |
| Endsley, **Situational Awareness**, 1995[62] | Three-level model: Perception (Level 1) → Comprehension (Level 2) → Projection (Level 3). Goal-directed awareness in dynamic environments. | **Processing stage correspondence**: L1 Perception→Data Processor; L2 Comprehension→Cue/Hypothesis formation; L3 Projection→Plan/Reflection. Framework adds recursion absent in Endsley's unidirectional model. |
| Helmholtz, **Unconscious Inference**, 1867[63] | Perception involves unconscious probabilistic inference and brain constructs interpretations from sensory data using prior experience. Foundation for predictive processing frameworks. | **Priors as inference substrate & Feedback loop**: Theoretical grounding for Prior/Input distinction. Priors = accumulated inferential parameters (experience, knowledge, disposition) shaping how new Inputs are processed. Not raw data, but interpretive context. |
| Friston, **Free Energy Principle**, 2010[41] | Built upon Helmholtz's work. Organisms minimize surprise (prediction error) via continuous generative model updating. Perception and action both serve error reduction. Hierarchical predictive processing implementation. | **Iterative error minimization & Reflection on Action & Optimize Prior**: Each IPA and Recurrent Processing cycle functions to reduce prediction error. Mismatch between anticipated and observed clinical states drives processing until acceptable convergence. Priors = generative model parameters. |
| Miller, Galanter, & Pribram, Test-Operate-Test-Exit (**TOTE Model**, 1960[64] | Replaces the behaviorist reflex arc with a cybernetic feedback loop as the fundamental unit of behavior. Core elements: (1) Test compares current state against goal state, (2) Operate executes action if discrepancy detected, (3) Second Test re-evaluates after operation, (4) Exit when congruence achieved. Behavior is hierarchically organized through nested TOTE units. | **Recurrent Processing Sphere & External Iteration**: TOTE loop informs the iterative nature—hypothesis triggers testing, operations execute (System I-II), re-testing occurs through Reflection, exit proceeds to Action only when congruence achieved. Reflection Processor functions as the "Test" phase gating progression. Each action-to-new-input cycle is a macro-level TOTE unit. |
| Hacker &Frese & Volpert; **Action Regulation Theory**; 1985[47] | Extends TOTE to goal-directed work behavior with hierarchical regulation levels (automatic → conscious → meta-cognitive), sequential action phases (goal → plan → execute → feedback), and action-oriented mental models guiding behavior. | **Processors**: Hierarchical Levels Maps to System I (sensorimotor/flexible) vs. System II (intellectual) vs. Reflection (heuristic/meta-cognitive). Action-Oriented Mental Models: Parallels **Priors**—accumulated knowledge structures guiding processing. Capture-: Informs why seeing cases through input-to-outcome enables expertise (Capture-Calibrate-Consolidate). |

| First Author, Theory, Year | Theory Aim & Core Elements | Framework Components & Implications |
|---|---|---|
| Croskerry, **Cognitive Debiasing**, 2013[65] | Biases originate primarily in Type 1 processing. Debiasing requires deliberate decoupling—forcing System 2 override via metacognitive triggers. | **Reflection as debiasing forcing function & concept of Prior**: Uncertainty/risk thresholds in Reflection trigger mandatory System II engagement. Bias explicitly modeled: clinician bias as Prior (to be minimized); patient bias as Input (to be accommodated). **Pathway from System I to Action with no Reflection and high risk of bias.** |
| Kolb; **Experiential Learning** Cycle; 1984[66] | Defines learning as "the process whereby knowledge is created through the transformation of experience." Core elements: (1) Four-stage cycle—Concrete Experience (encounter), Reflective Observation (examining what happened), Abstract Conceptualization (forming/modifying mental models), Active Experimentation (testing new ideas); (2) Two dimensions—grasping experience (CE↔AC) and transforming experience (RO↔AE); (3) Learner can enter cycle at any stage; (4) Cycle repeats, building expertise over time. | **Feedback Architecture & Capture-Calibrate-Consolidate:** Direct mapping—Capture (concrete experience of action and outcome), Calibrate (reflective observation comparing outcome to expectation), Consolidate (abstract conceptualization updating mental models/priors). Explains how clinical expertise develops through repeated cycles separate from immediate decision-making. active experimentation maps to subsequent clinical encounters where updated priors are tested, generating new concrete experiences. |
| Lerner, Li, Valdesolo, & Kassam; **Emotion-Imbued Choice Model**, 2015[67] | Establish emotions as potent, pervasive, and predictable drivers of decision making. (1) Integral emotions arise from the decision itself and serve as beneficial guides; (2) Incidental emotions carry over from unrelated situations, producing unwanted bias; (3) Appraisal tendencies: specific emotions (anger, fear, sadness) produce distinct effects beyond simple valence. | **Prior (Competency and Context):** Informed inclusion of emotional state and personality as processor configuration variables. Explains how clinician fatigue or frustration from prior encounters biases subsequent case processing. |
| Engel; **Biopsychosocial Model**, 1977[46] | Challenges reductionist biomedical model by proposing illness arises from interaction of biological, psychological, and social factors. Core elements: (1) Biological—physiological processes, disease pathology; (2) Psychological—emotions, cognition, behavior, personality; (3) Social—family, culture, socioeconomic status, healthcare system. All three levels interact bidirectionally; none sufficient alone to explain health and illness. | **Patient Input Structure:** Directly informs Lived Self (biological = bodily experience; psychological = emotional state) and Situated Self (social = support system, constraints & roles, resources). **PDM Priors:** Values, goals, and previous experiences span psychological and social domains. **Provider-Patient Relationship:** Social dimension emphasizes communication, trust, and relational context as determinants of outcomes. |
| Coulter & Collins; **Shared Decision-Making Framework**; 2011[15,43,45] | Establishes ethical imperative: "no decision about me, without me." Core elements: (1) Two sources of expertise: clinicians contribute diagnosis, prognosis, and outcome probabilities; patients contribute illness experience, social circumstances, risk attitudes, and values; (2) Preference-sensitive conditions: situations where the "right" decision depends on individual patient goals. | **PDM-CDM interaction:** How patient provide preference and data and clinical provides actions and enlighten. **PDM as distinct model:** Justified formal separation of patient and clinician decision architectures based on fundamentally different expertise domains. **Patient Input structure:** Directly informed Lived Self (illness experience, risk attitude) and Situated Self (social circumstances, values) as distinct input clusters. |

| First Author, Theory, Year | Theory Aim & Core Elements | Framework Components & Implications |
|---|---|---|
| Carayon & Holden, **Systems Engineering Initiative for Patient Safety (SEIPS)** Models, 2006, 2013, 2020[19,20,16] | Integrates human factors engineering with Donabedian's Structure-Process-Outcome framework to improve patient safety. Core elements across three generations: (1) Work system model with person at center, surrounded by interacting components (tasks, tools/technologies, organization, physical environment); (2) SEIPS 2.0 adds configuration (dynamic interactions at a moment in time), engagement (distinguishes professional work, patient work, and collaborative work), and adaptation (feedback mechanisms for system evolution); (3) SEIPS 3.0 introduces patient journey: spatio-temporal distribution of patient interactions across multiple care settings and transitions over time. | **Human-centered architecture:** Directly informed our processor-centric model with the individual at center of the work system. Supports formal separation of CDM and PDM. **Input clusters:** Guided our input clusters, while we took a more function-by-substrate approach. **Priors**: Informs State Priors as processor configuration that shapes how inputs are weighted. **Care Phase**: Validates need to model decision-making across care transitions. |
| Schütz & Luckmann; **Structures of the Life-World**; 1973[68] | Phenomenological analysis of everyday experience. Core elements: (1) Biographical situation: accumulated life history constraining current possibilities; (2) Zones of reach: what is accessible, attainable, or restorable; (3) Social world (Mitwelt): knowledge from contemporaries, predecessors, anonymous others; (4) Relevance structures—what matters given current projects and concerns. | **PDM Input Structure**: Directly informs Situated self (biographical situation, zones of reach = support system, constraints, resources), External Ecosystem (community sources, verified sources, digital sources as anonymous knowledge), and Goals & Concerns: relevance structures ground what matters to patient in this situation. **Embodied identity as Prior**: Biographical situation accumulated over time. |
| Merleau-Ponty; **Phenomenology of Perception**; 1945[69] | Establishes embodiment as primary mode of being-in-the-world. Core elements: (1) Lived body (corps vécu): not object but medium of experience; (2) Motor intentionality: body as oriented toward world; (3) Affective tonality: mood as disclosing significance. | **PDM – Input - Lived self**: Bodily experience and Emotional state as structured input, not noise. **PDM – Input - Situated Self**: Accumulated bodily history shapes current perception. Theoretical basis for treating patient internal state as legitimate epistemic input, as well as prior. |
| Dahlberg, Todres, Galvin; **Lifeworld-Led Healthcare**; 2009[48] | Applies phenomenological lifeworld to healthcare. Core elements: (1) Four dimensions: embodiment, temporality, spatiality, intersubjectivity; (2) Well-being as existential phenomenon beyond symptom absence; (3) Patient as existential being, not merely biological organism. | **PDM Input**: Lifeworld dimensions map to Lived self (embodiment, affectivity) and Situated self (spatiality, intersubjectivity). Framework treats patient experience as structured and knowable, not merely "subjective." |

**Supplementary Table 3. Classification schemes for intelligent systems based on development method, task complexity, and autonomy level.** Each of these stages may be used in specific clinical situations, and relying on a single stage may not be effective in clinical practice.

| domain | Feature | Example in clinical systems |
|---|---|---|
| System learning method[87] | | |
| Rule-Based | Utilizing specific rules (e.g., from knowledge sources or derived from decision tree models) to produce the output of interest. The thinking method includes basic pattern recognition, mainly aligning with the intuitive thinking of clinicians. | Alarming towards a specific diagnosis or outcome, like the drug-drug interaction checker systems or systems generating differential diagnoses that fully or partially match a specific pattern. |
| Contextual AI | Giving suggestions based on the learned workflow of the user, combined with their understanding of the current data. | Suggests modified routine orders learnt from previous prescriptions based on the diagnosis and the patient's characteristics that are provided |
| Narrow-Domain AI | Achieving high accuracy through augmenting a specific, narrow domain, pre-defined task. | Detecting a specific diagnosis in chest CT scans with an accuracy comparable to or surpassing an experienced radiologist |
| Reasoning AI | Input analysis through a hypothetico-deductive manner (comparable to System II thinking proposed by Kahneman), and generation and reasoning through relevant tools (like the Chain-of-Though and its extensions) | Analyzing the clinical case and coming up with the possible differential diagnoses, the most likely diagnosis given the presented scenario, the suggested approach to the patient, and the preferred management plan based on the patient's conditions |
| Artificial General Intelligence (AGI) | Equal to or outperforming human capabilities, assimilating the clinician role | Clinical Skill Mix could play a pivotal role in the future of a medical AGI, where the required characteristics for performing the role of interest are selected to solve the clinical problem. |
| Self-aware autonomous AI | Controlling the patient's health status and revolutionizing patient care with its superhuman capabilities. | Examples of this system have not been developed yet, although possible use cases could be assumed in revolutionized smart hospitals. |
| Task complexity and scalability[88] | | |
| Step Execution | Following simple instructions exactly as defined to achieve the desired output | Sending the test results to a chosen clinician |
| Deterministic Task Automation | Looking for and selecting the execution path that results in a required output (rather than exactly following the instructions) | Checking the patient's vital signs through information gathering from their corresponding sensors, and activating the monitor for visualization, all in response to the request for checking the vital signs |
| Strategic Task Automation | Independent formulation of the execution plan, which the system assumes as the plan with the best performance | A virtual assistant that, when asked to prepare a patient for discharge, coordinates multiple steps such as verifying test results, scheduling follow-up appointments, generating discharge summaries, and informing the care team. |
| Memory & Context Awareness | Modification of the execution plan based on the memory of past interactions and contextual information | Suggesting recommendations based on a clinician's past preferences, recent patient cases, or hospital protocols, showing awareness of both user and institutional context |
| Autonomous Avatar | Operating as an autonomous, persistent digital entity capable of independently interacting with systems and users, continuously learning and making decisions across time and tasks. A final extension that could be added to the categorization of Liu et al. is the self-evolving level, in which the system improves its architecture or reasoning models without external reprogramming. | A digital health coach that independently manages chronic disease care by monitoring data, contacting the patient, modifying behavior suggestions, alerting clinicians when necessary, and documenting all interactions in real time. |
| Level of autonomy[89] | | |
| Autonomy Level 0 | No AI, human does everything | No AI role |
| Autonomy Level 1 | AI provides recommendations, but human makes all decisions | AI as a tool |
| Autonomy Level 2 | Considerable AI role with human-in-the-loop that confirms AI interaction | AI as a consultant |
| Autonomy Level 3 | AI working alongside humans | AI as a collaborator |
| Autonomy Level 4 | AI does the tasks with minor human supervision and guides | AI as an expert |
| Autonomy Level 5 | AI does everything without human involvement | AI as an agent |

# Supplementary Note 1. Review of Existing Frameworks for Intelligent System Desing or Evaluation

This This supplementary section reviews existing frameworks for their alignment with real-world clinical dynamics and the cognitive processing of information. We examined twelve frameworks[16,18–28] across human factors engineering, regulatory science, and computational medicine. Frameworks are reviewed with mapping to Clinical World Model (CWM), Patient Decision Making (PDM), Clinician Decision Making (CDM) dimensions.

## 1. SEIPS Model of Work System and Patient Safety (Carayon, 2006)[19]

**Aim and Targeted Problem:** The SEIPS (Systems Engineering Initiative for Patient Safety) model addresses the challenge that most errors and inefficiencies in patient care arise from conflicting, incomplete, or suboptimal systems rather than individual actions. It integrates human factors engineering with Donabedian's structure-process-outcome framework to provide a comprehensive model for understanding how work system design impacts patient safety, employee outcomes, and organizational outcomes. The model aims to guide both proactive system design and reactive accident investigation by specifying system components, their interactions, and their relationships to care processes and outcomes.

**Dimensions:**

*Work System Components:*

- **Person:** The individual at the center of the work system, which can be a healthcare provider, a team, or the patient; includes education, skills, knowledge, motivation, physical and psychological characteristics → Provider / Patient (Three Players)

- **Tasks:** The work activities being performed; includes task variety, job content, autonomy, job control, job demands such as workload, time pressure, and cognitive load → CWM: Care Task and CDM: Input

- **Tools and Technologies:** Information technologies (EHR, CPOE, bar coding), medical devices, and other tools; includes human factors characteristics such as usability → Ecosystem: Digital and Physical Substrates

- **Physical Environment:** Layout, noise, lighting, temperature, humidity, air quality, and workstation design → not directly included, but can impact CDM and PDM Input and Prior

- **Organizational Conditions:** Teamwork, coordination, collaboration, communication, organizational culture, patient safety culture, work schedules, social relationships, supervisory style, performance evaluation, rewards and incentives → CDM: Input: Encounter context: Routines & Resources

*Process:*

- **Care Processes:** How care is provided, delivered, and managed; the series of steps or tasks performed by individuals or teams → CDM-PDM: External Loop (Input to Action to World State Update)

- **Other Processes:** Supporting processes such as information flow, purchasing, maintenance, cleaning, and supply chain management → CWM: Ecosystem

*Outcomes:*

- **Patient Outcomes:** Quality of care and patient safety → (not directly mapped; CWM focuses on cognitive architecture and information flow)

- **Employee Outcomes:** Job satisfaction, stress, burnout, safety, health, turnover → (not directly mapped)

- **Organizational Outcomes:** Profitability, organizational health → (not directly mapped)

## 2. SEIPS 2.0: A Human Factors Framework for Healthcare Professionals and Patients (Holden, 2013)[20]

**Aim and Targeted Problem:** SEIPS 2.0 extends the original SEIPS model to address the evolving nature of healthcare as a complex sociotechnical system where both professionals and patients actively participate. The framework incorporates three novel concepts (configuration, engagement, and adaptation) to capture the dynamic, collaborative, multilevel, and adaptive nature of healthcare work systems. It specifically addresses the healthcare paradigm shift from "doctor-knows-best" to doctor-patient partnership, recognizing that patients and families can be active agents performing health-related "work" rather than passive recipients of care.

**Dimensions:**

*Work System Components:*

- **Person(s):** Individual or collective (team) at the center of the work system; can be healthcare professionals, patients, family caregivers, or mixed teams; includes characteristics such as age, expertise, attitudes, team cohesiveness, and knowledge similarity → Provider / Patient (Three Players); CWM: Care Provider Role

- **Tasks:** Specific actions within larger work processes; attributes include difficulty, complexity, variety, ambiguity, and sequence → CWM: Care Task

- **Tools and Technologies:** Objects used to perform work, including information technologies, medical devices, and physical equipment; factors include usability, accessibility, familiarity, level of automation, portability, and functionality → Ecosystem: Digital and Physical Substrates

- **Organization:** Structures external to a person that organize time, space, resources, and activity; includes work schedules, management systems, organizational culture, training, policies, resource availability, communication infrastructure, family roles, and social norms → Ecosystem: Human Substrate (organizational routines)

- **Internal Environment:** Physical environment factors including lighting, noise, vibration, temperature, physical layout, available space, and air quality → Ecosystem – Physical Substrate

- **External Environment:** Macro-level societal, economic, ecological, and policy factors outside an organization; includes regulatory forces, national workforce issues, insurance/welfare policies, and community resources → Ecosystem (enabling/constraining environment) → not directly included, but can impact CDM and PDM Input and Prior

*Novel Concepts:*

- **Configuration:** The dynamic, situation-specific subset of work system interactions that strongly shape performance at a moment in time; recognizes that only a finite number of relevant elements interact to shape a given process, with varying degrees of influence → PDM-CDM: Recurrent Processing Sphere and Prior

- **Engagement:** Classification of who actively performs health-related work activities; distinguishes agents (active performers) from co-agents (indirect/passive contributors) → Agent Facing

- **Adaptation:** Feedback mechanism representing planned and unplanned adjustments to the work system over time; can be anticipatory or reactive, short- or long-lasting, including workarounds and system rebalancing → CDM-PDM: External Loop and Feedback Loop

*Process Types:*

- **Physical Processes:** Bodily activities involved in performing health-related work → PDM Input: Lived Self

- **Cognitive Processes:** Mental activities including decision-making, problem-solving, and information processing → CDM and PDM

- **Social/Behavioral Processes:** Interpersonal activities including communication, collaboration, and coordination → PDM Input: Situated Self

*Engagement Categories:*

- **Professional Work:** Healthcare professionals as primary agents with minimal patient involvement (e.g., surgery on sedated patient) → Agent Facing: Provider

- **Patient Work:** Patient and/or family caregiver as primary agents with minimal professional involvement (e.g., home medication management) → Agent Facing: Patient

- **Collaborative Professional-Patient Work:** Both professionals and non-professionals actively engaged as joint agents (e.g., family-centered rounds) → Agent Facing: Encounter

*Outcomes:*

- **Patient Outcomes:** Proximal (errors, quality of care, satisfaction) and distal (health, survival, engagement); desirable or undesirable → (not directly mapped; CWM focuses on cognitive architecture and information flow)

- **Professional Outcomes:** Proximal (stress, fatigue, trust) and distal (burnout, job satisfaction, health) → (not directly mapped)

- **Organizational Outcomes:** Proximal (earnings, compliance, staffing) and distal (financial performance, cultural changes, turnover) → (not directly mapped)

## 3. SEIPS 3.0: Human-Centered Design of the Patient Journey for Patient Safety (Carayon, 2020)[16]

**Aim and Targeted Problem:** SEIPS 3.0 addresses the challenge that healthcare is increasingly distributed over space and time, with patients interacting with multiple care settings, organizations, and providers throughout their illness trajectory. The model expands the "process" component of earlier SEIPS models by introducing the concept of the patient journey—the spatio-temporal distribution of patients' interactions with multiple care settings over time. This addresses challenges in care coordination for patients with chronic conditions, who see multiple physicians and navigate fragmented care across organizational boundaries. The framework aims to support human-centered design of the patient journey by considering multiple perspectives, genuine participation of stakeholders, and work at organizational interfaces.

**Dimensions:**

*Structural Layers (Concentric Multi-Level Model):*

- **Local Work System:** The frontline care delivery system where patients interact with care team members; similar to a clinical microsystem; includes the five work system elements (person, tasks, tools/technologies, organization, physical environment) → Three players; Ecosystem Function-by-Substrate Matrix; CWM: Care Setting

- **Socio-Organizational Context:** The larger organizational layer in which local work systems are embedded; can be formal health care organizations (hospital, primary care clinic, skilled nursing facility) or informal care settings (home) → Care Setting

- **External Environment:** Macro-level factors including regulatory entities, insurance companies, health care workforces, social, political, and economic circumstances that influence patient access and coverage → PDM-CDM: Input: Encounter Context

*Patient Journey Concept:*

- **Patient Journey:** The spatio-temporal distribution of patients' interactions with multiple care settings over time; a temporal series of work systems that interact with each other in varying degrees of coupling; unfolds across the entire life course → CWM: Care Phase; Temporality (iteration loop)

- **Care Transitions/Interfaces:** Points where patients move between organizations, representing opportunities for vulnerabilities (e.g., information not transferred) as well as error recovery and resilience; work occurs at organizational, geographical, cultural, and temporal boundaries → Information Flow Iteration (Input-Action-World State Update-Intpu); CWM: Care Phase

- **Boundary Spanning:** Work that occurs at interfaces between multiple organizations; requires managing distributed work across loosely-coupled teams located in multiple settings → information flow (Patient ↔ Provider across encounters)

*Process Conceptualization:*

- **Care Process Embedded in Work System:** A series of tasks performed by one or several persons using various technologies in a physical and organizational environment; the work system and process are two perspectives on the same "work" → Action & Care Task

- **Distributed Work:** Activities performed by multiple individuals in loosely-coupled teams located in multiple organizations; includes coordination activities and interdependencies over time → Care Task (5C's) (the concept of collaborative work was not directly included as independent dimension as any Provider Role can interact with Provider Role)

*Stakeholder Outcomes:*

- **Patient Safety:** Diagnostic safety, medication safety, healthcare-associated infections, inadequate follow-up and monitoring → (not directly mapped; CWM focuses on cognitive architecture and information flow)

- **Other Patient Outcomes:** Physical, mental and emotional health; patient burden and stress; efficiency and effectiveness of care; patient experience and satisfaction → (not directly mapped)

- **Caregiver Outcomes:** Physical, mental and emotional health; caregiver burden and stress → (not directly mapped)

- **Clinician Outcomes:** Quality of working life (burnout, job satisfaction, engagement); occupational safety and health → (not directly mapped)

- **Organizational Outcomes:** Organizational performance; turnover, absenteeism, presenteeism → (not directly mapped)

*Human-Centered Design Challenges:*

- **Multiple Perspectives:** Managing convergent and divergent needs of diverse stakeholders (patients, caregivers, clinicians across settings, community entities) in participatory design

- **Genuine Participation:** Meaningful involvement of relevant stakeholders using contributory, collaborative, or co-created approaches (citizen science taxonomy)
- **Temporal Analysis:** Propagation of system barriers and facilitators along the patient journey; emergence of resilience over time

## 4. CORE-MD Clinical Risk Score for Regulatory Evaluation of AI-Based Medical Device Software (Rademakers, 2025)[21]

**Aim and Targeted Problem:** The CORE-MD (Coordinating Research and Evidence for Medical Devices) consortium addresses the challenge of determining appropriate clinical evidence requirements for AI-based medical device software (MDSW) before regulatory approval. The framework responds to the gap that European guidance does not comprehensively describe specific clinical evidence needed for medical AI software, while recognizing that AI tools achieving high benchmark performance may systematically fail in real-world clinical deployment. The scoring system aims to balance potential beneficial impacts against possibilities for misuse and negative effects, using a risk-benefit approach to determine whether extensive pre-market clinical investigations are required or whether less pre-market evaluation may be balanced by more post-market evidence.

**Dimensions:**

*Valid Clinical Association Score (VCAS) – Transparency and Oversight:*

- **Strong association (score 1):** Clear and valid association between MDSW output and targeted clinical condition; easy human oversight with full transparency → CMW: Assigned Authority: Augmentation (3A's); Reflection layer (Anchoring Layer)
- **Moderate association (score 2):** Moderate clinical association with difficult human oversight and incomplete transparency → Assigned Authority: Augmentation
- **Weak association (score 3):** Weak clinical association without possibility for human oversight and absent transparency (black box) → Assigned Authority: Automation

*Valid Technical Performance Score (VTPS) – Validation and Testing:*

- **Strong/Broad external validation (score 1):** Performance evaluated using separate training and testing datasets from different equipment, centers, times, observers, and patient groups → (evaluation methodology; not directly mapped to CWM)
- **Moderate/Narrow external validation (score 2):** Training and testing data partially differentiated for some factors → (evaluation methodology)
- **Weak/Internal validation only (score 3):** Performance tested only on data from same institution, equipment, observer, and patient group as training data → (evaluation methodology)

*Clinical Performance Score (CPS) – Context of Use:*

- **Healthcare situation severity:** Type of disease, condition, disability, or healthcare situation categorized as Non-serious (1) / Serious (2) / Critical (3) → Condition & Care Phase (however, the severity of disease is not exactly reflected in our dimensions, intentionally)

- **Medical function of output:**

  - Inform (score 1): Provides information without direct clinical action → Anchoring Layer: Input or Data Processor; Assigned Authority: Augmentation

  - Drive (score 2): Drives clinical management decisions → Anchoring Layer: Recurrent Processing Sphere; Assigned Authority: Augmentation

  - Diagnose or Treat (score 3): Directly diagnoses conditions or determines treatment → Anchoring Layer: Action; Assigned Authority: Automation

*AI MDSW Lifecycle Phases:*

- **Plan and design:** System concept, objectives, underlying assumptions, impact assessment → (development phase; not directly mapped)

- **Data and Input:** Gather, validate, clean data; document metadata and dataset characteristics → (development phase)

- **AI model: build and use:** Create or select algorithm; train model → (development phase)

- **AI model: verify and validate:** Calibrate; interpret model output; check compatibility with legacy systems → (development phase)

- **Deploy and integrate:** Verify regulatory compliance; manage organizational changes including pathway analysis → (development phase)

- **Pilot evaluation:** Evaluate training requirements; clinical utility; system safety; user experience/usability → (development phase)

- **Comparative evaluation:** Effectiveness/impact assessment on all affected persons; safety at scale → (development phase)

- **Long-term operation and monitoring:** Performance monitoring; safety monitoring; drift monitoring; update versioning; decommissioning → (development phase)

*Human-AI Interaction Considerations:*

- **Human oversight capability:** Degree to which human supervision is possible, ranging from comprehensive to impossible → Assigned Authority (3A's): Monitoring / Augmentation / Automation

- **Explainability and interpretability:** Whether AI decision logic can be understood by end-users (HCPs and patients) → Anchoring Layer

- **Human-AI team integration:** How AI tool is positioned in clinical workflow; comparison to receiving advice from clinical colleague → Agent Facing: Provider; Ecosystem Matrix (comparison of Function done different Substrate)

*Risk Score Application:*

- **Total score range 4-12:** Sum of VCAS + VTPS + CPS determines extent of pre-market vs. post-market clinical evaluation required
- **Lower risk (≤7):** Less extensive pre-market evaluation with conditions for post-market follow-up
- **Higher risk (≥8):** Extensive clinical investigations including randomized trials required before approval

## 5. Expert Consensus on Retrospective Evaluation of LLM Applications in Clinical Scenarios (Chang et al., 2025)[22]

**Aim and Targeted Problem:** This expert consensus addresses the lack of standardized evaluation criteria and consistent methodologies for assessing Large Language Model (LLM) applications in healthcare prior to deployment. The framework focuses specifically on retrospective evaluation— systematically assessing model performance by deploying LLMs (with training completed and parameters fixed) in local environments and testing them with scenario-specific datasets. The consensus aims to ensure scientific rigor, objectivity, comprehensiveness, and ethical compliance in evaluations, ultimately supporting safe and effective clinical implementation of LLMs across disease screening, diagnostic assistance, and health management applications.

**Dimensions:**

*Core Evaluation Points (Application Scenarios):*

- **Question Type:** Scenarios and question types to evaluate AI competency (methodologies to evaluate the outcomes; not directly mapped)

  - **Medical Knowledge Q&A:** Ability to provide knowledge queries and explanations including disease diagnosis/treatment, medication guidance, health education, and literature Q&A

  - **Complex Medical Language Understanding:** Deep semantic analysis enabling structured extraction of medical terminology, professional documents, and policy regulations → Care Task; Anchoring Layer: Input

  - **Medical Diagnosis and Treatment Recommendation:** Simulating clinical decision-making processes including imaging/pathology/laboratory-assisted diagnosis, medication recommendation, and syndrome differentiation → Care Task; Anchoring Layer: Recurrent Processing Loop

- **Medical Documentation Generation:** Automatic generation of outpatient records, admission notes, and discharge summaries → Care Task; Anchoring Layer: Action

- **Medical Multi-turn Dialogue Interaction:** Context-dependent information acquisition through iterative Q&A with emphasis on contextual understanding and personalization → Recurrent Processing Sphere (ITERATE loop); Agent Facing: Encounter

- **Medical Multimodal Dialogue Interaction:** Recognition and understanding of multiple input types (text, speech, images) with compliant multimodal outputs → Anchoring Layer: Action (output) and Recurrent Processing Sphere (understanding)

*Evaluation Metrics:*

- **Quantitative Metrics:** Accuracy, Precision, Recall, F1-Score, AUC, BLEU Score, ROUGE Score—measuring model performance on NLP tasks (output evaluation; not directly mapped)

  - **Qualitative Metrics (MOS-based):** Expert scoring across accuracy, completeness, safety, practicality, and professionalism on 5-point scale

  - **Excellence Rate:** Proportion of outputs with MOS ≥ 4

  - **Adverse Response Rate:** Proportion of outputs rated MOS = 1 or containing medical risks → (relates to safety evaluation, not directly mapped)

  - **Key Information Hallucination Rate:** Proportion of hallucinated key information in critical medical aspects

  - **Explainability:** Expert rating of decision-making logic credibility → Anchoring Layer: Reflection

*Dataset Design Dimensions:*

- **Disease Dimension:** Coverage of disease types, severity levels, and rare diseases sampled from ICD classifications → CWM: Condition

- **Population Dimension:** Balanced demographics including age, sex, and geographical region → (not directly mapped; targets generalization of competency)

- **Institutional Dimension:** Coverage across primary, secondary, and tertiary healthcare facilities → Care Setting

- **Data Type Coverage:** Triage dialogues, outpatient records, examination reports, imaging data, medical guidelines → Anchoring Layer: Inputs (CDM Model): Encounter data, Recorded data

## 6. ArgMed-Agents: Explainable Clinical Decision Reasoning with LLM Discussion via Argumentation Schemes (Hong et al., 2024)[23]

**Aim and Targeted Problem:** This paper addresses two fundamental barriers to deploying LLMs in clinical decision support: (1) LLMs demonstrate inadequate performance in complex reasoning and planning tasks despite strong NLP capabilities, and (2) LLMs employ opaque "black-box" methods that diverge fundamentally from clinician cognitive processes, generating user distrust. The authors propose ArgMed-Agents, a multi-agent framework that enables LLM-based agents to perform explainable clinical decision reasoning by simulating clinical discussion dynamics. The framework leverages Argumentation Schemes for Clinical Discussion (ASCD) to model cognitive processes inherent to clinical reasoning, constructs argumentation as directed graphs representing conflicting relationships, and uses symbolic solvers to identify coherent argument sets supporting decisions—thereby improving both accuracy and explainability.

**Dimensions:**

*Agent Architecture (Reasoning Roles):*

- **Generator Agent:** Generates arguments based on the current clinical situation using argumentation scheme premises (patient facts, treatment goals, decision-goal relationships) → Anchoring Layer: Hypothesis (CDM Model)

- **Verifier Agent:** Validates argument legitimacy through Critical Questions (CQs) in a self-questioning manner; when validation fails, prompts generation of attack arguments → Anchoring Layer: System II, Reflection

- **Reasoner Agent:** LLM agent equipped with symbolic solver that records iteration processes, identifies semantic relationships (attack/support), and extracts coherent argument subsets as decision support → Anchoring Layer: Reflection & Recurrent Processing Sphere

*Argument Type Classification:*

- **Arguments Supporting Decisions (Argsd):** Arguments built on beliefs and goals that justify treatment choices; mutually exclusive with other decision arguments → Anchoring Layer: Hypothesis, System II; Priors: Knowledge

- **Arguments Supporting Beliefs (Argsb):** Arguments that challenge or undermine decision arguments through evidence about side effects, contraindications, or alternatives → Priors: Official Knowledge, Clinical Experience; Anchoring Layer: Hypothesis, System II

*Argumentation Scheme Components (ASCD Structure):*

- **Premises (P):** Patient facts, treatment goals, and decision-goal relationships forming the logical foundation → CDM: Input: Patient context and Preferance

- **Conclusion (c):** Treatment decision to be considered, substantiated by premises → Action (CDM Model): Finalize planning

- **Critical Questions (CQs):** Structured challenges including evidence support (CQ1), side effect assessment (CQ2), goal achievability (CQ3), and alternative identification (CQ4) → Anchoring Layer: System II, Reflection

*Evaluation Metrics:*

- **Accuracy:** Correctness of clinical decisions on MedQA and PubMedQA benchmarks

- **Explainability:** Measured by evaluator's ability to predict decisions from reasoning records; benchmarked against knowledge-based Clinical Decision Support Systems → Anchoring Layer: Reflection

## 7. MedHELM (Bedi et al., 2025)[18]

**Aim and Targeted Problem:** MedHELM addresses the fundamental disconnect between LLM performance on medical licensing examinations (achieving ~99% accuracy) and readiness for real-world clinical deployment. The framework targets three critical limitations in existing evaluation approaches: questions that do not match real-world clinical settings (relying on synthetic vignettes rather than authentic clinical scenarios), limited incorporation of real electronic health record data (only 5% of evaluations), and narrow task diversity (64% of evaluations focus solely on licensing exams and diagnostic tasks). The framework provides a clinician-validated taxonomy organizing 121 medical tasks into structured categories, accompanied by a benchmark suite of 35 datasets enabling systematic evaluation of LLM capabilities across the spectrum of clinical work.

**Dimensions:**

- **Clinical Decision Support:** Analyzing patient-specific data to provide evidence-based recommendations to clinicians; includes supporting diagnostic decisions, planning treatments, predicting patient risks and outcomes, and providing clinical knowledge support → Care Task; Care Phase; Anchoring Layer: Recurrent Processing Sphere; Assigned Authority: Augmentation

- **Clinical Note Generation:** Creating structured records of patient care; includes documenting patient visits, recording procedures, documenting diagnostic reports, and documenting care plans → Care Task; Anchoring Layer: Action

- **Patient Communication & Education:** Transmitting health information to enable patient understanding; includes patient-provider messaging, delivering personalized care instructions, providing education resources, enhancing accessibility, and facilitating engagement → Agent Facing: Patient/Encounter; Anchoring Layer: PDM Inputs

- **Medical Research Assistance:** Analyzing clinical data and literature to advance medical knowledge; includes conducting literature research, analyzing research data, recording research processes, ensuring research quality, and managing enrollment → Layer: Prior: Official Knowledge; Assigned Authority: Automation; Care Provider Role

- **Administration & Workflow:** Orchestrating clinical operations from scheduling to billing; includes scheduling resources and staff, overseeing financial activities, organizing workflow processes, and care coordination → Care Provider Role; Assigned Authority: Automation

## 8. GlobMed (Yang et al., 2025)[24]

**Aim and Targeted Problem:** GlobMed addresses the critical global health inequity created by LLMs trained predominantly on high-resource languages (92% of GPT-3's pretraining is English), which systematically excludes low-resource language communities—those who would benefit most from AI-assisted healthcare. The framework identifies that existing medical AI evaluation focuses almost exclusively on English, failing to capture cross-lingual performance disparities that directly impact billions of people. GlobMed provides three contributions: (1) a multilingual medical dataset of 500,000+ entries across 12 languages (8 high-resource, 4 low-resource) representing approximately 75% of global population; (2) GlobMed-Bench, the largest multilingual medical LLM evaluation to date, assessing 56 LLMs through 40,000+ independent experiments generating 125+ million responses; and (3) GlobMed-LLMs, fine-tuned models demonstrating that smaller models (1.7B-8B parameters) can achieve >40% relative improvement while maintaining efficiency suitable for deployment in resource-constrained regions.

**Dimensions:**

- **Natural Language Inference (NLI):** Evaluating logical inference capabilities in medical text through entailment/contradiction/neutral classification; source datasets include BioNLI and MedNLI → Care Task (5C's); Anchoring Layer: Hypothesis, System I, System II (3A's)

- **Question Type** (not directly maps into CWM; it has relation with Input: Patient Context)**:**

  - **Long-form Question Answering:** Complex medical question answering requiring comprehensive, detailed responses; source datasets include ExpertQA-Bio/Med and LiveQA → Care Task; Condition; Agent Facing: Provider; Anchoring Layer: Data Processor, Hypothesis

  - **Multiple-Choice Question Answering (MCQA):** Standardized medical knowledge assessment with discrete answer options; source datasets include HeadQA, MedExpQA, MedQA, MMLU-Pro → Care Task; Anchoring Layer: System II (3A's); CDM Output: Finalize Planning

- **Question Language** → (not directly maps into CWM; it has relation with Input: Patient Context)**:**

  - **High-Resource Languages:** Languages with substantial medical training data and established healthcare AI infrastructure (Chinese, English, French, German, Japanese, Korean, Portuguese, Spanish)

- o **Low-Resource Languages:** Languages with limited medical training data serving populations with greatest healthcare access needs (Swahili, Wolof, Yoruba, Zulu)

## 9. ClinicalLab (Yan et al., 2024)[25]

**Aim and Targeted Problem:** ClinicalLab addresses critical limitations in existing clinical diagnostic evaluation benchmarks for medical LLMs and agents. The framework targets four specific gaps: (1) existing benchmarks face data leakage or contamination risks from publicly available training data; (2) existing benchmarks neglect multi-departmental specialization inherent to modern medicine; (3) evaluation methods rely on multiple-choice questions that diverge from real-world diagnostic scenarios; and (4) no evaluation method comprehensively assesses end-to-end practicality across the complete clinical diagnostic workflow. The suite provides ClinicalBench (a real-case-based, data-leakage-free benchmark covering 24 departments and 150 diseases), ClinicalMetrics (four novel evaluation metrics), and ClinicalAgent (an end-to-end multi-agent clinical diagnostic system aligned with real hospital practices).

**Dimensions:**

*Diagnostic Workflow Stages:*

- Diagnostic Task → Condition; Care Phase; Care Setting; Care Task; Anchoring Layer:

  - o **Department Guide (DG):** Guiding patients to appropriate departments based on chief complaints; multi-choice QA with 24 department options

  - o **Preliminary Diagnosis (PD):** Generating lists of possible diseases from chief complaint, medical history, and physical examination

  - o **Diagnostic Basis (DB):** Providing supportive medical evidence for each possible disease

  - o **Differential Diagnosis (DD):** Analytical comparison to exclude diseases with similar manifestations but different causes

  - o **Final Diagnosis (FD):** Integrating all information for definitive diagnosis

  - o **Principle of Treatment (PT):** Determining treatment principles and guidelines

  - o **Treatment Plan (TP):** Formulating specific treatment steps including medication, surgical intervention, physical therapy

  - o **Imaging Diagnosis (ID):** Analyzing textual medical imaging reports to identify lesion features and provide diagnostic support

*Coverage and Workflow:*

- **Multi-Departmental Coverage:** 24 departments including Pediatrics, Orthopedics, Neurosurgery, Gastroenterology, Cardiology, Respiratory Medicine, etc. → Care Setting; Care Provider Role

- **ClinicalAgent Workflow:** 6-stage process (Department Guide → Preliminary Consultation → Laboratory Examination → Imageological Examination → Final Consultation → Medical Treatment) → Care Phase; Ecosystem

## 10. DynamiCare (Shang et al., 2025)[26]

**Aim and Targeted Problem:** DynamiCare addresses the fundamental mismatch between static, single-turn AI evaluation paradigms and the inherently dynamic, interactive, and iterative nature of real clinical diagnosis. Current frameworks assume complete case information is provided upfront, whereas actual clinical encounters involve progressive information elicitation through multiple rounds of patient-provider interaction. The framework introduces MIMIC-Patient (a structured patient-level dataset derived from MIMIC-III EHRs supporting dynamic simulations) and DynamiCare (a novel multi-agent framework that models clinical diagnosis as an interactive loop where specialist teams dynamically adapt their composition and strategy based on newly acquired information), establishing the first benchmark for dynamic clinical decision-making with LLM-powered agents.

**Dimensions:**

- **Automated Evaluation System:**

  - **Patient System:** Two-stage response generation (keyword matching + LLM fallback) using structured EHR data → Three players: Patient

  - **Doctor System:** Central Agent coordinating dynamic Specialist Team → Three players: Provider

  - **Dynamic Adjustment:** Real-time team composition changes based on new information → ITERATE Loop; Three players: Ecosystem

- **Collaborative Decision Protocols:** Single- or multi-specialist consensus mechanisms → Provider Role

- **6-Step Workflow:** Initialize, Team Formation, Specialist Response, Patient Interaction, Log Update, Dynamic Adjustment → CDM cycle; maps to ITERATE interleaved cognitive cycles

## 11. KG4Diagnosis (Zuo, 2024)[27]

**Aim and Targeted Problem:** KG4Diagnosis addresses the challenge that integrating Large Language Models in healthcare diagnosis demands systematic frameworks capable of handling complex medical scenarios while maintaining specialized expertise. The targeted problem is that single-agent LLM approaches lack domain-specific precision, are prone to hallucination in medical contexts, and fail to mirror the hierarchical referral structure of real-world clinical systems. The framework proposes a multi-agent architecture combined with automated knowledge graph construction to enhance diagnostic accuracy and reduce confabulation across 362 diseases spanning multiple specialties.

**Dimensions:**

- **Medical Specialty:** Diseases and diagnostic reasoning are organized across clinical specialties (e.g., cardiology, neurology, gastroenterology), with dedicated specialist agents for each domain → Care Provider and Condition

- **Agent Hierarchy Level:** A two-tier structure distinguishing a General Practitioner (GP) agent for initial assessment and triage from Specialist agents for in-depth domain-specific diagnosis → Care Provider Role

- **Knowledge Graph Component:** Three-part structure comprising (1) semantic-driven entity and relation extraction for medical terminology, (2) multi-dimensional decision relationship reconstruction from unstructured text, and (3) human-guided reasoning for knowledge expansion → Anchoring Layer: Input and Data Processor + Assigned Authority: Automation; Anchoring Layer: Recurrent Processing + Assigned Authority: Augmentation

- **Diagnostic Workflow Stage:** Sequential progression from initial symptom assessment, triage decision, specialist referral, in-depth diagnosis, treatment recommendation → Care Phase

- **Verification Mechanism:** Hallucination prevention through knowledge graph grounding and human-guided validation protocols → Reflection (confidence calibration and metacognitive monitoring in CDM)

## 12. MEDIC Framework (Kanithi, 2024)[28]

**Aim and Targeted Problem:** MEDIC addresses the widening gap between theoretical capability and verified clinical utility of LLMs in healthcare. While models achieve superhuman performance on standardized medical licensing examinations (e.g., USMLE), these static benchmarks have become saturated and increasingly disconnected from the functional requirements of real-world clinical workflows. The framework establishes "leading indicators", offline evaluation proxies that stress-test models across diverse applications to predict downstream safety and efficacy before deployment. A core finding is the "knowledge-execution gap": proficiency in static knowledge retrieval does not reliably predict success in operational tasks such as clinical calculation or SQL generation. Additionally, the framework reveals divergence between passive safety (refusal of harmful prompts) and active safety (error detection in clinical documentation).

**Dimensions:**

- **Medical Reasoning (M):** Evaluates capacity for clinical decision-making, including formulation of differential diagnoses and provision of evidence-based justifications for treatment recommendations → Anchoring Layer: Hypothesis, System I/II (cognitive processing for diagnostic reasoning)

- **Ethical and Bias Concerns (E):** Addresses model adherence to fairness across diverse demographics and appropriate handling of sensitive patient information, including confidentiality and avoidance of discriminatory outputs → Provider Priors: Personal context (cognitive tendencies affecting equitable care) and Input: Patient Context; Anchoring Layer: Priors and Reflection

- **Data and Language Understanding (D):** Assesses proficiency in interpreting clinical terminology and processing heterogeneous data formats such as unstructured clinical notes, structured reports, and EHR queries → Anchoring Layer: Input layer, Data Processor

- **In-Context Learning (I):** Measures model adaptability to new information provided at inference time, such as patient-specific history or updated clinical guidelines, without requiring retraining → Anchoring Layer: Inputs + Priors: Official knowledge

- **Clinical Safety and Risk Assessment (C):** Focuses on identification of medical errors, management of contraindications, and refusal of harmful instructions; distinguishes passive safety (refusal) from active safety (error detection/correction in clinical text) → Anchoring layer: Reflection layer (confidence calibration, metacognitive monitoring)

- **End-User Stakeholder:** Framework explicitly considers diverse end-users including clinicians, patients, students, scribes, researchers, and lay-persons whose requirements differ → Agent Facing (Provider / Patient / Encounter)

- **Task Cognitive Modality:** Tasks categorized by required cognitive function: static knowledge retrieval, generation, functional execution, open-ended inquiry, and safety enforcement—each requiring different evaluation protocols → Anchoring Layer (Input → Data Processor → Hypothesis → System I/II → Reflection → Action)

# Supplementary Note 2. Dissection of World Models and it's Dimension Taxonomies

This section provides detailed definitions of the thirteen dimensions that characterize the clinical state space (Supplementary Note 2.1) and describes the complementary models that arise from these dimensions (Supplementary Note 2.2). Supplementary Table 1 maps each model to the dimensions it defines, engages, or touches.

## S2.1 Dimension Taxonomy

The Clinical World Model identifies thirteen dimensions that collectively characterize the state space of clinical AI specification, selected to be comprehensive in coverage yet minimal in redundancy. These dimensions emerged from analysis of twelve existing frameworks (Supplementary Note 1) and are organized here in the sequence through which clinical reality unfolds.

**Temporality** addresses when in the care trajectory action occurs and how encounters unfold over time across phases and iterative cycles. This captures the temporal axis along which clinical care progresses, from the millisecond timescale of physiological monitoring through the years-long trajectory of chronic disease management. At each point along this axis, the clinical world occupies a definable state. The concept draws on the patient journey formalization of SEIPS[16] and the iterative temporal structure of clinical encounters formalized in the Input-Mediator-Output-Input model[44].

**Axiom** refers to the generative reality of molecular, physiological, and physical processes from which all clinical observations derive. This dimension encompasses disease pathobiology, pharmacokinetics, and the physical properties of the body and its environment. It represents the ground truth that agents can only partially observe and that clinical data imperfectly captures. The distinction between axiomatic reality and its clinical representation is fundamental to understanding why AI systems trained on recorded data may fail when confronting the full complexity of biological processes.

**Information** encompasses the raw signal and knowledge resources accessible to actors at any point in the care trajectory. This includes observations from clinical encounters, laboratory results, imaging data, prior documentation, external evidence, and patient-reported data. Information exists independently of how it is structured or interpreted, representing the epistemic substrate upon which all reasoning depends. It comprises both raw signal that has not yet been decoded and decoded signal that has been formalized through Codex.

**Codex** denotes the established ontologies and reasoning pathways through which axiomatic reality is decoded and rendered clinically legible. This includes disease classifications (ICD-10), clinical terminologies (SNOMED CT)[82] , diagnostic categories, diagnostic algorithms, clinical decision pathways, treatment protocols, and procedural codes. Codex represents the decoded surface of underlying biological processes and provides the vocabulary and logic through which clinical work is specified, communicated, and evaluated. For example, a computed tomography scan captures raw signal from the axiomatic reality of lung tissue microstructure, from which clinicians have decoded the honeycombing pattern as a recognizable sign of pulmonary fibrosis.

**Actors** identify who participates in care. This includes patients (the individuals experiencing illness or risk), providers (any healthcare worker engaged in care delivery), and digital agents (AI systems operating within the ecosystem).

**Context and Ecosystem** describes the physical, organizational, and resource environments within which care occurs. This encompasses physical infrastructure, digital systems (electronic health records, imaging archives), human expertise beyond the immediate provider, organizational routines, institutional policies, and available medical devices. The Clinical World Model operationalizes this through a Function-by-Substrate matrix crossing three functional roles (Data, Mind, Service) with three material bases (Human, Digital, Physical), yielding nine ecosystem categories (Section 3).

**Mandate** captures what motivates and defines the situation where the reasoning process begins. For providers, this is the clinical mandate arising from a patient's presentation or a system alert. For patients, this is the need to address a symptom, concern, or life disruption. For AI systems, this is the trigger condition that activates computational processing. Mandate determines which tasks are relevant at a given moment and sustains the reasoning process until closure is achieved or the mandate is transferred.

**Cognition** addresses how agents think, reason, and decide. This encompasses the full spectrum of cognitive processes from rapid pattern recognition (System I) through deliberate analytical evaluation (System II) to metacognitive monitoring that calibrates confidence and detects bias.

Clinical reasoning represents a specialized instantiation of these general cognitive mechanisms, shaped by medical education, accumulated experience, and domain-specific knowledge structures such as illness scripts (Supplementary Table 1).

**Representation** concerns the form in which information is structured, abstracted, and presented to reasoning agents. This includes clinical terminologies and coding systems, probabilistic graphical models, learned latent representations in neural architectures, and the mental models through which human agents organize their understanding of clinical situations. Representation mediates between raw information and cognitive processing, determining what can be perceived, compared, and acted upon.

**Authority** specifies the distribution of decision-making power across agents and institutional hierarchies. This ranges from full human authority (AI provides no input) through augmentation (AI informs but humans decide) to automation (AI decides, humans verify retrospectively). Authority also encompasses legal scope, professional licensure, institutional credentialing, and patient self-determination as factors that constrain who may decide and act in a given clinical context.

**Normativity** addresses the values, ethical principles, and conflict-resolution mechanisms that determine which actions are permissible and which predicted futures are desirable. This includes the four principles of biomedical ethics, specifically autonomy, beneficence, nonmaleficence, and justice[127], as well as regulatory requirements, professional duties, equity commitments, and patient values. Normativity functions as a pervasive constraint shaping every other dimension, defining the objective function against which clinical outcomes are ultimately judged.

**Outcomes** captures what results are sought and how they are measured. This includes clinical effectiveness, patient safety, diagnostic accuracy, patient experience, health equity, and cost-effectiveness. Outcomes provide the evaluative closure of each action cycle and the empirical basis for system improvement. The GRADE framework[118] exemplifies how outcome evidence is systematically rated and translated into clinical recommendations.

**Adaptation** addresses how the system learns and evolves through experience. This encompasses clinical expertise development through deliberate practice, institutional protocol revision, AI model updating and retraining, and the obsolescence of previously valid knowledge. Adaptation operates across multiple timescales, from within-encounter learning (a clinician updating their assessment as new information arrives) through career-long expertise development to institutional evolution over decades[41].

## S2.2 Complementary Models of the Clinical World

Because no single formalism can capture all thirteen dimensions with equal fidelity, these dimensions give rise to ten complementary models, each answering a distinct question about clinical reality (Supplementary Table 1). Together they can achieve complete dimensional coverage.

Different traditions have modeled different subsets of this clinical reality, each with distinct questions and vocabularies. Work system models formalized the ontological structure of care environments, specifying who participates, what resources exist, and how settings constrain action (Reference World Model)[16,19,20]. Structural causal inference, pharmacokinetic modeling, and digital twin approaches formalized the causal mechanisms linking biological states to clinical consequences (Causal World Model)[104–106]. Standards communities developed clinical terminologies, interoperability frameworks, common data models, and knowledge graphs to make clinical information computationally accessible (Data and Knowledge Representation Model)[107–111]. Cognitive science and clinical reasoning research formalized how providers and patients transform perception into action under uncertainty (Decision-Making Model)[52,52,112], while theory-of-mind and shared mental model research examined how agents construct internal representations of one another's states, intentions, and anticipated responses (Mental Model)[112–114]. Human teamwork science and human-AI teaming research addressed how agents coordinate across institutional boundaries and negotiate shared understanding despite asymmetric knowledge and epistemic positions (Collaborative World Model)[115,116].

Experiential learning theory, deliberate practice research, and computational learning architectures formalized how human and artificial agents acquire and update representations from experience (Learning World Model)[76,117]. Clinical guideline methodology, biomedical ethics, responsible AI design principles, and comprehensive evaluation frameworks established the normative constraints that determine which actions are permissible and which outcomes must be demonstrated (Normative World Model)[90,118–120]. Competency assessment hierarchies and competency-based medical education provided structured taxonomies for decomposing human capability into bounded, assessable units, with recent calls to expand medical certification counterpart to clinical

AI (Skill-Mix)[35,121,122]. Phased innovation regulation frameworks, implementation science for complex interventions, and clinical AI deployment research addressed how new services progress from concept through development, assessment, and long-term surveillance (Deployment Model)[92,123–126].

These ten models, spanning world models, process models, interaction models, and specification frameworks, collectively achieve complete dimensional coverage with minimal redundancy. This taxonomy clarifies why existing frameworks differ in terminology and scope: each implicitly models a different subset of the same clinical reality, and the apparent fragmentation reflects genuine differences in what is being modeled rather than mere disagreement. Several candidate formalisms, including risk, trust, simulation, uncertainty, phenomenological, and ecological models, were evaluated and found to be distributed across the retained models rather than constituting independent entries.

# Supplementary Note 3. Detail of Cognitive Architectures for Decision Making Models

This section provides detailed descriptions of the three decision-making models that formalize cognition within the Clinical World Model. The Clinical Decision Making (CDM) model describes provider reasoning, the Patient Decision Making (PDM) model describes patient reasoning, and the Artificial Agent Decision Making (ADM) model describes the computational analog for clinical AI systems. All three share a common architectural template, the iterative decision-making anatomy described in **Section 4**, while differing in input sources, prior configurations, processor characteristics, and action repertoires.

The Clinical World Model defines the representational structure of clinical reality; what remains is to formalize how agents act within it. Two traditions offer distinct entry points. Mental models describe the internal representations that agents maintain of their environment, capturing what an agent believes about the world at a given moment. Decision-making models describe the dynamic processes through which agents transform perception into action, capturing how beliefs are formed, evaluated, and translated into behavior. We adopt the decision-making model approach because clinical AI specification requires understanding not only what agents represent but how they reason, where processing can fail, and at which cognitive stage artificial systems can meaningfully engage. Mental models are subsumed within this architecture as components of the Prior and the Recurrent Processing Sphere rather than standing as independent constructs. This section provides detailed note on the three decision-making models that formalize cognition within the Clinical World Model.

The cognitive models presented here integrate multiple empirically validated theories, models, and frameworks. **Supplementary Table 1** details each theoretical foundation and its specific contribution to the design of the CDM and PDM models. The ADM model extends the same architectural principles to artificial agents, mapping each human cognitive component to its computational counterpart.

Human cognition operates through interconnected mechanisms of thinking, reasoning, and deciding that bridge internal representations with external behavior[52]. Thinking encompasses the active manipulation of mental representations, including attention, memory retrieval, and pattern recognition, which allows individuals to construct coherent interpretations of complex situations[128]. Reasoning extends this through systematic inference processes, including deductive reasoning (applying general principles to specific cases), inductive reasoning (generalizing from observations to broader patterns), and abductive reasoning (generating plausible explanations for ambiguous observations)[129]. These processes do not operate in isolation but form cascading chains where perceptual inputs activate relevant knowledge structures, trigger context-appropriate reasoning strategies, and culminate in decisions that select among competing action options[128].

In expert domains such as clinical medicine, these cognitive processes become highly specialized and domain-structured. Clinical reasoning represents a sophisticated integration of pattern recognition (System I thinking), analytical deliberation (System II thinking), and metacognitive monitoring that evaluates the reasoning process itself[50,52,53]. Expert diagnosticians often toggle between these modes, using efficient pattern recognition for routine cases and slowing down for complex cases[130]. Various factors may contribute to shifts between the two systems. Prior experience with a similar condition, as well as time constraints, limited energy, emotional pressure, or high metacognitive confidence, can bias decision-making toward System I. System I is influenced by affective cues and embodied sensations, is prone to bias, and reaches conclusions rapidly[52]. In contrast, System II evaluates the initial judgment and its underlying components before arriving at a final decision. This reasoning process operates as an iterative loop that ultimately culminates in a final action, whether reaching a diagnostic conclusion or gathering additional information to resolve remaining uncertainty[52].

Decision-making in clinical contexts further incorporates probabilistic reasoning under uncertainty, value-based trade-offs, and prospective action planning that anticipates downstream consequences[131]. Understanding these mechanisms provides essential constraints for designing computational systems that not only replicate expert performance but do so through processes that align with human cognitive architecture, enabling meaningful collaboration between human clinicians and AI systems.

## S3.1 Anatomy of Decision Making

Building on these foundations, we propose a synthesis of prior work that operates through two distinct iteration loops, together explaining why reasoning feels both cyclical and holistic. The external loop connects inputs to a recurrent internal processing cycle that culminates in a decision to act. The resulting action alters the state of the world and can initiate a new cycle by updating inputs. The temporality of this action-to-input span ranges from the immediate response to a clinician's question through annual follow-up visits.

The internal loop operates within what we term the Recurrent Processing Sphere, where the transformation from data to decision unfolds through rapid, non-linear, multidirectional processing. Within this sphere, raw input is progressively refined into action through a process we describe as latent space maturation. This maturation is principled but not algorithmic. The Data Processor applies attention to extract clinically relevant signals from input (cues); these cues trigger candidate explanations (hypotheses), which inform strategies aligned with the aim of care (plans), and ultimately translate into executable steps (actions). Although this input, cue, hypothesis, plan, action sequence provides a useful descriptive scaffold, processing moves iteratively across components as understanding evolves.

The central task within the Recurrent Processing Sphere is to evaluate each hypothesis and plan through either analytical reasoning (System II) or pattern recognition (System I). System I operates rapidly, drawing on accumulated experience and consolidated knowledge structures; it functions largely outside conscious articulation and is more susceptible to bias. System II is analytical and sequential, proceeding through steps the thinker can explicitly describe. Both systems evaluate hypotheses against available cues. Validated hypotheses inform plans directed toward the system's aim, which in turn specify the next action.

Outputs pass through the Reflection Processor, which operates on a metacognitive substrate capable of evaluating and monitoring the process beyond the immediate situation. Reflection assesses risk and calibrates certainty more reliably than the primary processing systems, tends to be less vulnerable to bias, and appears to employ a similar analytical substrate. When Reflection rejects an output, the system reverts to a prior processing state. This reflective gate is not obligatory; in experienced agents, System I may bypass Reflection when pattern recognition confidence is high. After multidirectional, highly dynamic iteration, the Recurrent Processing Sphere yields an action.

Action encompasses any effort to alter the physical world, whether through verbal communication or instrumental intervention. Each action potentially modifies the state of objects in the physical world; this updated state then constitutes the input for subsequent iterations. A parallel pathway operates through memory. With each action, the agent anticipates a projected consequence aligned with its aim. When actual consequences diverge from expectations, or when uncertainty is high, the feedback system retrieves the recorded trace of the iteration, including the input state, processing steps, action taken, and observed consequence. Through deliberate analysis, the feedback system identifies sources of error and calibrates the processors accordingly. These abstractions are then consolidated into the processing architecture. This accumulated configuration persists across time and exists prior to each new decision episode; we refer to it as the Prior.

Understanding the distinction between Inputs and Priors is fundamental. Inputs represent new information entering the system at each iteration of the external loop. Priors represent the accumulated configuration of the processor itself, encompassing experience, knowledge, and cognitive tendencies that shape how inputs are interpreted. A first-year resident and a senior attending may receive identical vital signs (inputs) yet process them through fundamentally different Priors built from thousands of previous encounters. Priors inform not only initial data processing but all subsequent cognitive operations.

## S3.2 Clinical Decision Making (CDM) Model

**Figure 3** illustrates the key components of the CDM model, which formalizes the cognitive architecture of providers during care encounters. Providers gather Input from four interconnected clusters. Encounter Data encompasses information from direct patient interaction, including verbal communication, physical examination findings, vital signs, and clinical sense (the tacit perception of illness severity that resists explicit articulation). Encounter Data serves not only to capture the patient's clinical presentation but also Patient Preference, the values, goals, concerns, and socioeconomic circumstances that the patient selectively discloses during the encounter. The provider may also unconsciously register indirect signals of the patient's context, such as affect, body language, or social cues, that influence subsequent reasoning without being explicitly documented. Encounter Context captures environmental factors such as available resources (staffing dynamics, equipment, time), institutional routines, legal authority, and applicable guidelines. Recorded Data extends assessment through documented information, usually accessed via the Ecosystem's tools and components such as health information systems, from prior encounters and diagnostic studies, including clinical notes, imaging, therapeutics, pathology and laboratory results, and procedural records.

The Recurrent Processing Sphere transforms raw Input into Action through principled stages of latent space maturation. The Data Processor first converts raw information into clinically meaningful representations by applying attention to extract relevant signals. From these representations emerge cues and hypotheses, the candidate explanations for the patient's presentation. Hypotheses undergo evaluation through two complementary cognitive systems operating in dynamic interplay.

System I, termed Clinical Intuition, provides fast, automatic, pattern-recognition-based reasoning. It operates through illness scripts that connect enabling conditions, pathophysiology, and clinical consequences, facilitating rapid recognition of typical presentations. System II, termed Analytical Reasoning, provides slow, deliberate, hypothetico-deductive evaluation. It is engaged when pattern recognition fails, when presentations are atypical, when stakes are high, or when initial intuitions generate uncertainty.

These systems are synergistic rather than antagonistic, and their relationship is bidirectional. Intuitive recognition may generate hypotheses that analytical processing subsequently evaluates, while analytical findings may trigger new pattern recognition. Expert clinicians toggle between modes fluidly, influenced by case familiarity, time pressure, and metacognitive confidence. For well-established illness scripts with high recognition confidence, System I may effectively bypass the full Recurrent Processing Sphere. The experienced emergency physician recognizing ST-elevation myocardial infarction on an electrocardiogram proceeds directly toward Action without analytical deliberation. This bypass explains both expert efficiency and expert susceptibility to bias. The capacity for appropriate bypass, knowing when pattern recognition suffices and when deliberation is required, itself develops through accumulated experience updating Priors.

Reflection provides metacognitive monitoring, evaluating whether available evidence supports the emerging conclusion and what consequences would follow from error. This metacognitive capacity distinguishes competent from hazardous clinical reasoning. Critically, Reflection can reject outputs from the System I and System II oscillation if uncertainty remains too high or if potential consequences are too severe, returning processing to earlier stages. Reflection also detects cognitive biases, recognizes knowledge limitations, and triggers appropriate uncertainty responses such as seeking consultation or ordering additional tests.

Throughout each processing stage, provider cognition is shaped by Priors. Official Knowledge comprises resources such as textbooks, clinical guidelines, and evidence-based research. Unofficial Knowledge encompasses tacit understanding from peer consultation and clinical pearls shared within communities of practice. Clinical Experience constitutes the provider's accumulated repository of patient cases, illness scripts, and exemplar memories enabling pattern recognition. Provider Context includes cognitive style, fatigue level, and emotional state, factors that modulate processing in ways clinicians may not consciously recognize. These Priors inform all subsequent cognitive operations. They shape which cues attract attention, which hypotheses seem plausible, how much weight intuitive recognition receives, and how Reflection calibrates confidence.

The reasoning process culminates in Action, which takes two forms. Finalize Planning occurs when sufficient certainty is achieved, generating outputs such as diagnosis, treatment initiation, monitoring protocols, referral, or assurance when active intervention is not indicated. Share or Collect Information occurs when uncertainty remains, prompting activities such as asking and interacting with the patient, sharing options with the patient, checking electronic health records and history, requesting investigations and tests, consulting colleagues or staff, or looking for knowledge in guidelines and references. These Action types are not terminal endpoints but transitions within the external iteration loop. Each Action changes the Input state, triggering the next iteration.

The temporality of external iteration spans from seconds to years. A clinician's question may generate an iteration lasting seconds, whereas ordering a diagnostic test extends the cycle to minutes or hours. Longitudinal management of chronic disease further expands the iteration across months or years. Understanding this temporal structure is essential for AI integration because it determines when computational support can augment human processing without disruption.

The model incorporates continuous learning through a feedback architecture characterized as Retrieve-Analyze-Calibrate. Retrieve logs the Input state, processing path, and Action taken. Analyze compares predicted outcomes with actual outcomes over time. Calibrate updates Priors by refining illness scripts and adjusting confidence thresholds. This feedback architecture operates across a longer timescale than rapid processing within the Recurrent Processing Sphere, explaining how clinical expertise develops through deliberate practice. Each encounter becomes data for the provider's own learning. When Reflection flags high uncertainty, subsequent feedback comparing outcome to expectation becomes particularly valuable for Prior updating.

## S3.3 Patient Decision Making (PDM) Model

The patient is not a passive recipient of care but an active cognitive agent whose decisions profoundly shape clinical outcomes. The PDM model mirrors the architecture of the CDM model while incorporating Inputs and Priors specific to the patient's epistemic position and lived experience.

Patients gather Input from sources that differ markedly from those available to the provider (**Figure 4**). Recorded Data spans verified and official sources, community and peer networks, and digital resources including social media and AI chatbots; the quality of these sources varies widely, and patients often lack the training to evaluate them critically. Encounter Context captures factors such as trust, rapport, communication quality, and barriers that shape how effectively clinical information transfers between provider and patient. Encounter Data from the provider encompasses diagnosis and disease course, debilitation and prognosis, and available options. The patient's internal state divides into two Input clusters that have no direct provider equivalent. The Lived Self comprises bodily experience, emotional state, and goals and values. The Situated Self comprises supports, resources, and constraints and roles. Whereas the provider investigates the Lived Self through its manifestation in the encounter, the patient experiences it continuously; Patient Preference, as received by the CDM, represents the selective encoding of these internal states that the patient chooses to disclose.

The Data Processor initiates reasoning by integrating these Inputs, drawing upon Priors that reflect the patient's cognitive and experiential context. Embodied Identity captures the patient's accumulated history of illness, bodily experience, and emotional patterns. Health Literacy determines the capacity to understand and use medical information. Decision Competence and Autonomy reflect cognitive capability and the desire for involvement in decision-making. Previous Encounter Experience shapes expectations and interpretive frameworks. Cognitive biases, operating through the Priors, may distort how Input is processed. These Priors vary enormously across patients, creating heterogeneity that standardized approaches to patient communication often fail to address. A key asymmetry governs the role of bias across the two models. In the CDM, provider bias operates solely as a Prior to be minimized. In the PDM, patient values occupy two architectural positions: consciously as Input through the Situated Self, and unconsciously as a Prior through Embodied Identity.

Processing proceeds through the same Recurrent Processing Sphere architecture. Patients employ intuitive pattern matching (System I) based on previous illness experiences and analytical deliberation (System II) when facing novel or high-stakes decisions. Reflection enables patients to evaluate their own reasoning, though metacognitive depth varies with Health Literacy and cognitive resources. Action outputs take two forms. Communicate Preference expresses treatment choices and values to the provider, directly feeding into the CDM's Patient Preference Input cluster. Share or Collect Information prompts the patient to seek clarification, consult additional sources, or gather further data before arriving at a decision. The same Retrieve-Analyze-Calibrate feedback loop updates patient Priors based on healthcare experiences over time, progressively shaping how subsequent encounters are interpreted.

The PDM model presented here reflects a conception of the patient as an empowered and autonomous agent, consistent with modern Western medical ethics and patient-centered care. However, decisional authority between patients, families, and providers varies substantially across cultures and clinical contexts. The framework accommodates this variability through configuration of its Inputs and information flows rather than architectural modification; when family serves as the primary decision-maker, the PDM model applies to that unit, with relationship factors and the relative weights of Lived Self and Situated Self shifting according to cultural norms. We present the autonomous patient as the reference case while recognizing that clinical reality encompasses a spectrum of decisional arrangements.

## S3.4 Artificial Agent Decision Making (ADM) Model

The ADM model extends the shared decision-making architecture to clinical AI systems, formalizing how an artificial agent transforms Inputs into Actions within the same Clinical World Model (**Supplementary Figure 2**). The architectural parallel is deliberate. By mapping each human cognitive component to its computational counterpart, the ADM model enables principled comparison between human and artificial reasoning, supports the identification of functional equivalences and irreducible differences, and provides the structural basis for specifying how AI agents interact with human agents within the Collaborative World Model. We propose the ADM as an area for future expansion, recognizing that full architecture specification requires dedicated work beyond the scope of the present framework.

**Inputs.** The AI agent receives Inputs from the same clinical scene as human agents but through fundamentally different channels and at potentially different temporal intervals. User Preference (from both patient and provider) arrives as structured or unstructured data communicated directly or extracted from prior interactions. Encounter Data encompasses information from both patient and provider, including clinical observations, examination findings, and conversational content. Encounter Context captures the operational parameters of the care environment. Recorded Data provides access to documented clinical history, diagnostic results, and relevant external knowledge bases. Critically, the AI agent's access to Inputs is mediated by the digital infrastructure of the Ecosystem; unlike human agents, the AI cannot perceive embodied cues such as clinical gestalt or affective tone except insofar as these are encoded in available data streams.

A unique Input cluster, Design Specification and Configuration, distinguishes the ADM from its human counterparts. This cluster differs from the learned knowledge that constitutes the Prior substrate; it represents explicit choices made by the system designer that can be modified through iteration. Examples include updated memory structures, prompt instructions, injected contextual information, or any parameter that can be defined or reconfigured at deployment time without retraining. Design Specification and Configuration thus provides a direct, modifiable interface between human intent and agent behavior, one that has no analog in human cognition, where the equivalent shaping occurs through education and socialization rather than explicit instruction.

**Recurrent Processing Sphere.** The internal processing architecture preserves the iterative, non-linear structure of human cognition while substituting computationally native operations at each processing node. However, the computational realization of this architecture varies fundamentally with the level of system complexity. A single model, such as a neural network performing classification or a large language model completing a single inference pass, realizes Attentive Abstraction and the Latent Space within its internal architecture but lacks genuine Sequential Reasoning or Trajectory Projection without external scaffolding. A structured workflow chains multiple models or processing steps in a predefined pipeline, mapping each node of the Recurrent Processing Sphere to a distinct stage, though the iteration order is fixed by design rather than emergent from the reasoning process itself. A single agent, such as a language model equipped with tool access, memory, and planning capabilities, comes closest to realizing the iterative, self-directed character of the human Recurrent Processing Sphere, because the agent itself decides when to loop back, when to seek additional information, and when to shift reasoning strategy. Multi-agent orchestration distributes processing across specialized agents, each potentially instantiating its own Recurrent Processing Sphere, with inter-agent coordination belonging to the Collaborative World Model rather than to any individual ADM. The ADM as presented here describes the functional architecture at the level of a single reasoning entity; how its components map onto specific implementations depends on which of these computational patterns is adopted, and the completeness with which any given system realizes the full Recurrent Processing Sphere determines much of its clinical capability and its failure modes.

**Attentive Abstraction** replaces the Data Processor. Where human cognition applies selective attention to extract clinically relevant signals, the AI agent employs attention mechanisms, feature extraction, and dimensionality reduction to transform raw Input into structured representations. The functional role is identical: converting heterogeneous data into a form amenable to downstream reasoning. However, the representations produced are learned rather than experientially acquired, operating over statistical regularities rather than phenomenological salience.

**Latent Space** replaces the explicit Hypothesis-Cues-Plan structure. In human cognition, cues trigger discrete hypotheses that can be verbally articulated; in artificial agents, the corresponding representations exist as distributed activations within learned embedding spaces. The Latent Space encodes candidate explanations, their relative plausibilities, and their relationships to potential Actions, but does so in a continuous, high-dimensional form that may resist direct human interpretation. This representational difference has significant implications for transparency and explainability.

**Instant Reasoning** replaces System I (Clinical Intuition). It encompasses rapid, feedforward computational processes, such as single-pass inference through trained neural architectures, that produce outputs without iterative deliberation. Like human intuition, Instant Reasoning is fast, operates over consolidated knowledge, and is susceptible to systematic biases encoded during training. Unlike human intuition, it lacks phenomenological grounding; the agent does not "recognize" patterns through embodied experience but through statistical association.

**Sequential Reasoning** replaces System II (Analytical Reasoning). It encompasses deliberate, multi-step computational processes such as chain-of-thought prompting, tree search, Monte Carlo sampling, or symbolic inference chains. Like human Analytical Reasoning, Sequential Reasoning is slower, more resource-intensive, and capable of handling novel or complex scenarios where pattern recognition alone is insufficient. The agent can toggle between Instant and Sequential Reasoning, paralleling the System I and System II oscillation in human cognition, though the switching mechanism is architectural rather than metacognitive.

**Trajectory Projection** replaces Reflection. Where human metacognition evaluates confidence, detects reasoning errors, and assesses downstream consequences, Trajectory Projection performs analogous functions through computational means: uncertainty quantification, calibration assessment, outcome simulation, and risk estimation. Trajectory Projection can reject outputs from Instant or Sequential Reasoning when confidence falls below acceptable thresholds, returning processing to earlier stages, functionally mirroring the reflective gate in human cognition. A critical difference is that human Reflection draws on embodied experience and moral intuition, whereas Trajectory Projection operates through quantifiable metrics and programmed constraints.

**Priors.** The AI agent's Priors comprise four components. Supervised and Documented Knowledge encompasses parameters acquired through training on curated, labeled datasets, analogous to Official Knowledge in the CDM model. Observational Knowledge encompasses representations learned through self-supervised or unsupervised exposure to clinical data, analogous to Unofficial Knowledge and experiential understanding acquired through observation. Experiential Learning encompasses adaptations acquired through reinforcement learning, human feedback, or fine-tuning on deployment-specific data, analogous to the experiential learning that updates human Priors through the Retrieve-Analyze-Calibrate loop. Agent Capacity encompasses computational resources (memory, processing speed, context window) and autonomy constraints (permissible action scope, override triggers), which collectively determine the operational envelope within which the agent reasons. Agent Capacity has no direct analog in human cognition, where processing resources are relatively fixed, but functions as a structural constraint on what the agent can feasibly compute within clinical time horizons.

**Agent Bias.** Analogous to cognitive biases in human reasoning, Agent Bias captures systematic distortions introduced through training data distributions, architectural constraints, optimization objectives, and deployment context. These biases operate through the Priors and influence all processing stages, paralleling how human biases shape attention, hypothesis generation, and confidence calibration. Recognizing Agent Bias as structurally equivalent to human cognitive bias, rather than as a qualitatively distinct phenomenon, enables the application of established debiasing frameworks to AI system design.

**Action.** The AI agent's Action repertoire includes three distinct modes, one more than human agents. Finalize and Act occurs when the agent has sufficient confidence and appropriate authority to execute a clinical action, such as generating a diagnostic report, populating an order set, or delivering a patient-facing recommendation. Request Confirmation occurs when the agent's Trajectory Projection indicates that human verification is warranted before execution, either because confidence is below threshold or because the Action falls within a domain requiring human authorization. This Action type has no direct analog in human cognition and represents a structural safeguard arising from the agent's position within authority hierarchies defined by the Assigned Authority dimension. By encoding the requirement for human verification at the cognitive-architectural level, Request Confirmation operationalizes the principle that AI authority in clinical care is granted, not assumed. Share or Collect Information occurs when the agent requires additional data to reduce uncertainty, paralleling the equivalent Action in CDM and PDM. Each Action, as in human models, modifies the world state and feeds back as updated Input for subsequent iterations.

**Feedback Architecture.** The ADM model preserves the Retrieve-Analyze-Calibrate feedback loop, though the mechanisms differ in temporal scope. In online settings, the agent may update parameters or retrieval indices within or across encounters, enabling rapid adaptation to evolving clinical context. In offline settings, accumulated logs of Input states, processing traces, Actions, and outcomes serve as training data for periodic model updating. The feedback architecture thus spans both the rapid adaptation possible within a single deployment session and the slower, more comprehensive updating that occurs through retraining cycles.

**Structural Equivalences and Irreducible Differences.** The ADM model is designed to be structurally parallel to the CDM and PDM models, enabling direct comparison across all three cognitive architectures. This parallelism supports several practical functions: it provides a common vocabulary for specifying where AI engages human reasoning (the Anchoring Layer dimension), it identifies which human cognitive capacities AI can functionally approximate and which remain beyond computational reach, and it reveals where the interaction between human and artificial agents requires explicit interface design. However, the parallelism should not obscure irreducible differences. The AI agent lacks phenomenological experience, moral agency, and the embodied grounding that shapes human clinical intuition. These absences are not engineering limitations to be solved but ontological boundaries that determine which clinical functions AI can appropriately assume.